\documentclass[11pt,twoside]{article}

\usepackage{geometry}
\usepackage[linesnumbered, algoruled, boxed, lined]{algorithm2e}
\usepackage{amsmath,amssymb,fullpage,graphicx,mathtools,amsthm,enumitem,subfig,dsfont,xspace}
\usepackage{tablefootnote}
\usepackage{hyperref,cleveref}
\usepackage[numbers,sort&compress]{natbib}
\usepackage{booktabs}
\usepackage{caption}
\usepackage{subcaption}
\usepackage{tikz}

\let\citet\cite
\let\citep\cite
\newcommand{\figureheight}{0.21\linewidth}

\newtheorem{theorem}{Theorem}
\newtheorem{proposition}{Proposition}

\newtheorem{fact}{Fact}
\newtheorem{observation}{Observation}

\newcommand{\p}[1]{\left(#1\right)}
\newcommand{\sqb}[1]{\left[#1\right]}
\newcommand{\cb}[1]{\left\{#1\right\}}
\newcommand{\EE}[2][]{\mathbb{E}_{#1}\left[#2\right]}
\newcommand{\PP}[2][]{\mathbb{P}_{#1}\left(#2\right)}
\newcommand{\var}{\text{var}}

\let\Expect\E

\newcommand{\Prob}{\mathbb{P}}

\newcommand{\sample}{\sim}
\newcommand{\Beta}{\textrm{Beta}}

\newcommand{\given}{\mid}

\newcommand{\defn}{:=}
\newcommand{\setcomplement}[1]{\overline{#1}}

\DeclareMathOperator*{\argmax}{argmax}

\DeclareMathOperator*{\sign}{sign}

\let\Bernoulli\bernoulli

\DeclarePairedDelimiter\abs{\lvert}{\rvert}

\DeclarePairedDelimiter\ceil{\lceil}{\rceil}
\DeclarePairedDelimiter\floor{\lfloor}{\rfloor}

\newcommand{\indicator}{\mathds{1}}

\newcommand{\intersect}{\cap}
\newcommand{\union}{\cup}

\newcommand{\stepone}{\text{(i)}\xspace}
\newcommand{\steptwo}{\text{(ii)}\xspace}
\newcommand{\stepthree}{\text{(iii)}\xspace}

\newcommand{\const}{c}
\newcommand{\Const}{C}

\newcommand{\binomial}{\text{Binom}}

\newcommand{\egreedy}{$\epsilon$-greedy\xspace}
\newcommand{\datafeedback}{data sharing\xspace}
\newcommand{\rs}{recommendation system\xspace}
\newcommand{\rss}{recommendation systems\xspace}
\newcommand{\Rss}{Recommendation systems\xspace}
\newcommand{\ra}{recommendation algorithm\xspace}
\newcommand{\ras}{recommendation algorithms\xspace}

\newcommand{\sutva}{stable unit treatment value assumption\xspace}

\newcommand{\dm}{DM\xspace}
\newcommand{\GTE}{\text{GTE}}
\newcommand{\estGTE}{\widehat{\GTE}}

\newcommand{\ts}{Thompson sampling\xspace}

\newenvironment{quotex}%
  {\list{}{\leftmargin=0.3in\rightmargin=0.3in}\item[]}%
  {\endlist}

\newcommand{\timestep}{t}
\newcommand{\idxtimestep}{i}
\newcommand{\horizon}{T}
\let\numusers\horizon
\newcommand{\meanarm}{\mu}
\newcommand{\meanbest}{\meanarm^*}
\let\meantrue\meanarm

\newcommand{\numarms}{K}
\newcommand{\idxarm}{k}
\newcommand{\idxarmbest}{\idxarm^*}
\newcommand{\history}{\mathcal{H}}
\newcommand{\conf}{c}

\newcommand{\regret}{R}
\newcommand{\reward}{Y}
\newcommand{\varnumpulls}{N}

\newcommand{\armselect}{I}

\newcommand{\dist}{D}

\newcommand{\algo}{\mathcal{A}}
\newcommand{\algoucb}{\texttt{UCB}}
\newcommand{\algoucbone}{\algoucb_{\paramexplore_1}}
\newcommand{\algoucbtwo}{\algoucb_{\paramexplore_2}}

\newcommand{\algogreedy}{\texttt{grd}}
\newcommand{\algoegreedy}{\texttt{$\epsilon$-grd}}
\newcommand{\algoexp}{\texttt{EXP3}}
\newcommand{\algoalt}{\algo'}
\newcommand{\algotot}{\texttt{tot}}
\newcommand{\idxsample}{s}
\newcommand{\numsamples}{s}

\newcommand{\paramexplore}{\alpha}
\newcommand{\probexplore}{\epsilon}

\newcommand{\gap}{\Delta}
\newcommand{\gapmin}{\gap_{\textrm{min}}}
\newcommand{\term}{M}

\newcommand{\priorsize}{m}
\newcommand{\paramprior}{\gamma}

\newcommand{\cumprob}{\tau}
\newcommand{\event}{\mathcal{E}}

\newcommand{\assignment}{W}

\newcommand{\probassign}{p}
\newcommand{\paramrampup}{\beta}

\usepackage[dvipsnames]{xcolor}

\begin{document}

\begin{center}
{\bf{\LARGE{Choosing the Better Bandit Algorithm under Data Sharing: When Do A/B Experiments Work?}}}
\vspace*{.2in}

\renewcommand*{\thefootnote}{\fnsymbol{footnote}}
{\large{
\begin{tabular}{ccc}
Shuangning Li$^{1}$, Chonghuan Wang$^{2}$, Jingyan Wang$^{3}$\footnotemark
\end{tabular}
}}
\vspace*{.2in}
\footnotetext[1]{Corresponding author: \texttt{jingyanw@ttic.edu}}

\begin{tabular}{c}
$^{1}$University of Chicago\\
$^{2}$Naveen Jindal School of Management, University of Texas at Dallas\\
$^{3}$Toyota Technological Institute at Chicago
\end{tabular}
\renewcommand*{\thefootnote}{\arabic{footnote}}
\setcounter{footnote}{0}

\vspace*{.2in}

July 2025; \qquad Revised: February 2026

\vspace*{.2in}

\begin{abstract}
We study A/B experiments that are designed to compare the performance of two recommendation algorithms. Prior work has observed that the stable unit treatment value assumption (SUTVA) often does not hold in large-scale recommendation systems, and hence the estimate for the global treatment effect (GTE) is biased. Specifically, units under the treatment and control algorithms contribute to a shared pool of data that subsequently train both algorithms, resulting in interference between the two groups. In this paper, we investigate when such interference may affect our decision making on which algorithm is better. We formalize this insight under a multi-armed bandit framework and theoretically characterize when the sign of the difference-in-means estimator of the GTE under data sharing aligns with or contradicts the sign of the true GTE. Our analysis identifies the level of exploration versus exploitation as a key determinant of how data sharing impacts decision making, and we propose a detection procedure based on ramp-up experiments to signal incorrect algorithm comparison in practice.


\end{abstract}
\end{center}


\section{Introduction}

\Rss are widely deployed across online platforms. Users receive numerous recommendations every day, including news and creators' content on social media, products in online marketplaces, services in freelancing labor markets, ads on websites, and so on. During the development of such \rss, a crucial task that companies face all the time is to compare the performance of different recommendation algorithms, and make business decisions on which one to eventually deploy in production.

A common approach to comparing the performance of two recommendation algorithms is through randomized controlled trials, also known as A/B experiments. In a typical user-randomized A/B experiment, each user is assigned to a treatment group (running one recommendation algorithm) or a control group (running the other recommendation algorithm), uniformly at random. The metric to measure the performance of the two algorithms can be, for example, user engagement, click-through rates, purchase revenues, etc. Our goal is to estimate the global treatment effect (GTE), the difference between the treatment group and the control group in terms of this performance metric. More precisely, the GTE is defined as the difference in this performance metric between deploying the treatment algorithm to all users versus deploying the control algorithm to all users. The GTE captures the change if the company switches from the control algorithm to the treatment algorithm in production; it informs the company what algorithms to develop and deploy. In A/B experiments, a simple estimator for the GTE, termed the difference-in-means (\dm) estimator, is computed by taking the difference between the empirical mean of the performance metric in the treatment group and that in the control group. In causal inference, it is known that the \dm estimator is consistent for the true GTE under the \sutva (SUTVA), which requires that the outcome for any user is independent from the treatment assignment of any other user.

Interference refers to the phenomenon where users in the treatment and control groups influence each other, rendering the \dm estimator biased for the GTE.
On a production-scale \rs, the following field experiment has been conducted to demonstrate the effect of such interference~\cite{brennan2025symbiosis}. In this experiment, the treatment to users is to recommend more content that is recently published. Accordingly, these users consume more recently-published content. However, the more interesting observation is that users in the control group, with the \ra serving them unchanged, also start to consume more recently-published content.

\begin{figure}
    \centering
    \subfloat[]{\includegraphics[width=0.45\linewidth]{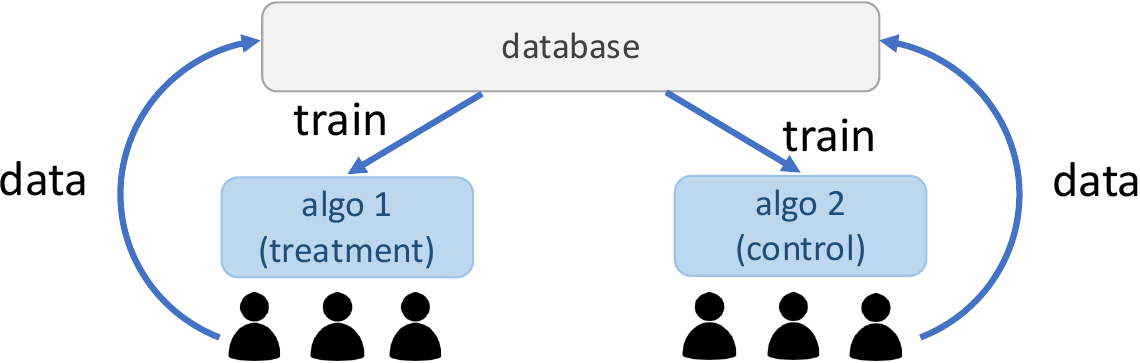}}\hspace{.5cm}
    \unskip\ \vrule
    \hspace{.5cm}\subfloat[]{\includegraphics[width=0.45\linewidth]{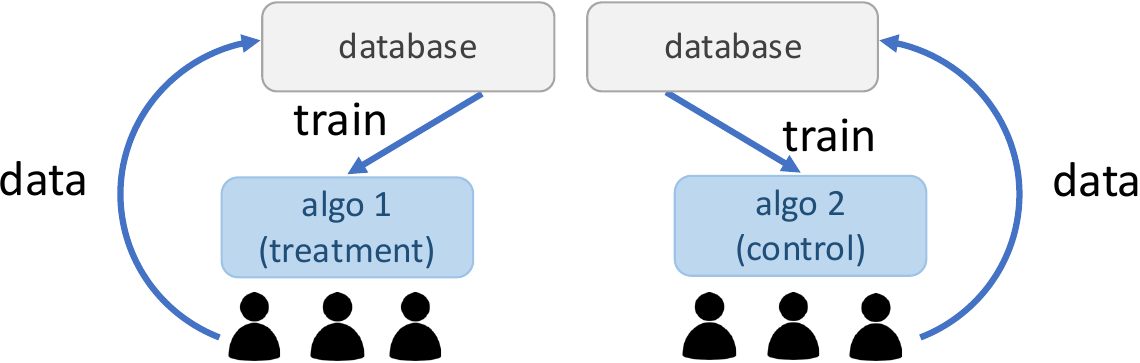}}
    \caption{Shared feedback loops in data create interference between the treatment and control groups (left panel), compared to an idealized environment where the two groups operate in isolation (right panel), which is often infeasible in practice.\label{fig:rct}}
\end{figure}

This phenomenon of interference is attributed to shared data feedback loops~\cite{brennan2025symbiosis}, illustrated in Figure~\ref{fig:rct} (left). As users in the treatment group engage more with recently-published content, they produce training data associated with the recently-published content. Such data is shared with and used to train the control algorithm, creating a mechanism for the treatment group to influence the control group. This experiment setup thus violates SUTVA and yields biased GTE estimates. This bias arising from such shared feedback loops is termed ``symbiosis bias'', describing how the two algorithms depend on each other in a symbiotic fashion through sharing data.\footnote{Interactions among users across treatment and control groups may also contribute to such phenomenon. For example, users in the control group engage with social media posts that their friends in the treatment group have liked or commented on, or users in the control group make purchases because they hear about these products from their friends in the treatment group.
To minimize such network effects, the experiment randomizes treatment assignment on a country level. That is, each country, along with all users in the country, is assigned to either the treatment or control groups uniformly at random. See~\cite[Section 5]{brennan2025symbiosis} for additional details on the precise experiment setup.}

Before we proceed to analyze the impact of symbiosis bias, let us consider two natural ideas to mitigate this bias and the additional challenges they introduce. The first mitigation strategy is to isolate the treatment and control groups so that they do not share any data, as shown in Figure~\ref{fig:rct} (right). This idealized design eliminates the shared feedback loops and hence the symbiosis bias, but has a number of drawbacks. First, it reduces the efficiency of learning because each algorithm now only has access to a smaller dataset for training~\cite{si2024weighted}, which is especially a problem if a small fraction of users is assigned to the treatment group. In business applications, it is desirable to reach decisions quickly, and using shared data to speed up learning is especially attractive in scenarios where the data from the treatment and control groups have similar distributions and the impact of symbiosis bias is minimal. Second, in large-scale \rss, maintaining two separate datasets increases the engineering efforts. Using social media as an example, the content may be ranked by a scoring function that predicts the number of likes for the content. If this scoring function is trained from the number of likes on historical data, now the number of likes by users in the treatment and control groups needs to be separately counted. The scoring function may also use neural networks to extract features from images and texts, and these neural networks are fine-tuned on user data. Now two separate neural networks need to be trained and stored. Such extra computation and storage needs are not scalable as companies often run many A/B experiments in parallel. 

A second mitigation strategy is to deploy the treatment or control algorithm to all users, but at different times, a design known as ``switchback experiments''. However, this experiment design introduces other types of time-dependent biases~\cite{hu2022switchback,bojinov2023design}. For example, in online marketplaces, consider a scenario where the control algorithm is deployed first and recommends a product to the user. After some time, the treatment algorithm is deployed and the user makes the purchase. It is unclear if the purchase decision is made due to the treatment algorithm or the control algorithm. Moreover, user behavior is inherently non-stationary. For instance, purchases of umbrellas increase during rainy seasons, and consumer demand in general fluctuates in response to broader economic conditions. It is thus non-trivial to design GTE estimators that carefully account for such non-stationary behavior in switchback experiments. 

Given the challenges of mitigating symbiosis bias through experimental design, in this paper, we focus on A/B experiments under data sharing and characterize the impact of symbiosis bias on the \dm estimator for the GTE. We highlight that the expected GTE estimates are often used for downstream decision-making. That is, companies are interested in knowing which one of the two recommendation algorithms is better, so that they can deploy the better algorithm in practice. With this objective in mind, precise distinctions such as whether the improvement is 20\% or 25\% are often less critical.
Our central research question is therefore:
\begin{quotex}
    \emph{Despite the presence of symbiosis bias in the \textbf{magnitude} of the \dm estimates for the GTE, how does symbiosis bias impact the \textbf{sign} of these expected GTE estimates?}
\end{quotex}

\subsection{Overview of results}
We answer the above question through the lens of the exploration–exploitation trade-off. Our key finding is that when comparing two recommendation algorithms, the impact of symbiosis bias depends on their level of exploration: If the worse algorithm underperforms because it over-explores (without data sharing), then symbiosis bias does not alter the sign of the expected GTE estimates under data sharing (a regime that we refer to as ``sign preservation''). In contrast, if the worse algorithm underperforms because it overexploits (without data sharing), then symbiosis bias reverses the sign of the expected GTE estimates (a regime that refer to as ``sign violation'').
We formally establish this result in terms of bandit algorithms, which play a central role in \rss. Under a multi-armed bandit setup, our metric of interest for measuring the GTE is the expected cumulative regret, where a higher regret indicates worse performance.

We use the greedy algorithm, which always selects the arm with the highest empirical mean, as our example of an over-exploiting bandit algorithm. While the greedy algorithm is highly suboptimal and incurs a linear regret, we show that its regret is reduced to a constant under \datafeedback, when it runs jointly with a general class of algorithms (Theorem~\ref{thm:greedy+general}), including the classic \egreedy algorithm (Theorem~\ref{thm:greedy+e-greedy}) and the upper confidence bound (UCB) algorithm (Theorem~\ref{thm:greedy+UCB}). This result suggests that the sign of the expected GTE estimate under \datafeedback yields a wrong comparison of the two algorithms. Moreover, we show that the magnitude of GTE estimate under data sharing is $\Theta(\log\horizon)$ in expectation, whereas the true GTE is $\Theta(\horizon)$, suggesting that the \dm estimator significantly under-estimates the magnitude of the GTE.

We then consider \egreedy and UCB algorithms as our examples of over-exploring bandit algorithms. We parameterize these algorithms  by a parameter $\paramexplore\in [0, 1]$, where a higher value of $\paramexplore$ corresponds to more exploration. The algorithm incurs a regret of $\Theta(\log\horizon)$ when $\paramexplore=0$ and a regret of $\Theta(\horizon^\paramexplore)$ when $0<\paramexplore\le 1$. These parameterized classes hence cover algorithms with different performances from being optimal $\Theta(\log\horizon)$ to the worst possible $\Theta(\horizon)$ regret due to over-exploration. We show that \datafeedback does not affect either the sign or the order of magnitude of the expected GTE estimates, for both the \egreedy and UCB algorithms (Theorem \ref{thm:sign-preserve-combined}).

Beyond the greedy, \egreedy, and UCB algorithms, we also study EXP3 and Thompson Sampling in Appendix~\ref{app:bandit_algorithms} and \Cref{sec:sim_ts} respectively, where we establish similar results, either theoretically or empirically.

\begin{figure}[tb]
    \centering
    \includegraphics[height=0.275\linewidth]{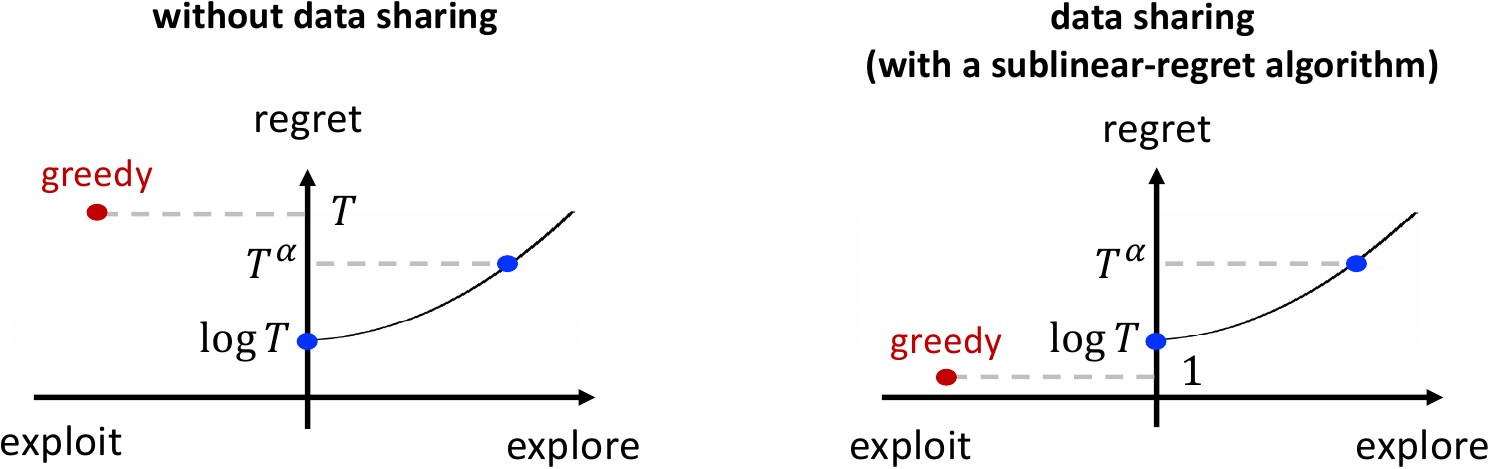}
    \caption{Our theoretical results contrast an algorithm's performance when running individually (left) versus under \datafeedback (right).\label{fig:conjecture}}
\end{figure}

These results together substantiate our theoretical insights on exploration versus exploitation, illustrated in Figure~\ref{fig:conjecture}. When an algorithm runs individually (left panel), too much exploration (e.g., sampling arms uniformly at random) or too much exploitation (e.g., the greedy algorithm) is suboptimal. The optimal $O(\log\horizon)$ regret is achieved by a careful balance between exploration and exploitation.
In contrast, when two algorithms run jointly under \datafeedback (right panel), exploration becomes the ``tragedy of the commons''. When an algorithm explores, the data from exploration is shared between two algorithms, but the cost to exploration (i.e., a high instantaneous regret by pulling a potentially suboptimal arm) only incurs to the algorithm that explores. Hence, the algorithm is incentivized to free ride the exploration done by the other algorithm rather than explore on its own.

Our results allow companies interested in selecting the better algorithm in A/B experiments to face a simpler problem: identifying whether their experiment falls under the regime of sign preservation or sign violation, rather than estimating the precise magnitude of symbiosis bias (as in, for example, \cite{si2024weighted}). In \Cref{sec:guidelines}, we address the practical question of how to diagnose these regimes in real A/B experiments. Specifically, we study ramp-up experiments in which the treatment allocation probability varies over a sequence of experiments. We find that when increasing traffic to the seemingly better algorithm leads to a deterioration in overall performance, this pattern is indicative of a sign violation regime. We formalize this intuition by proposing a simple diagnostic based on the monotonicity of average outcomes and provide theoretical justification for why this criterion distinguishes sign violation from sign preservation. This result connects our theoretical insights to standard experimental practice and offers a lightweight tool for practitioners to assess whether comparisons under data sharing can be trusted.

At a high level, we study the comparison of recommendation algorithms by combining ideas from causal inference and bandit algorithms. On the one hand, our motivation to estimate the GTE in A/B experiments brings new questions under a bandit formulation. On the other hand, our focus on comparing adaptive learning algorithms rather than fixed policies draws attention to dynamic settings in causal inference. We envisage that our results provide new perspectives to both communities.

\subsection{Related work}
Our work combines ideas from causal inference and bandit algorithms in the application of \rss. We review related work in these domains.

\subsubsection{Causal inference}
\paragraph{Interference due to shared data feedback loops.}
Shared data feedback loops can lead to interference when comparing recommendation algorithms in A/B experiments~\cite{si2024weighted,brennan2025symbiosis}. Specifically, training two recommendation algorithms on each other's data in an online fashion introduces bias in estimating the GTE; this bias is termed ``symbiosis bias''~\cite{brennan2025symbiosis}.  The existence of symbiosis bias has been demonstrated through a field experiment on a production-scale \rs~\cite{brennan2025symbiosis}. Different mitigation strategies have been considered. Brennan et al.~\cite{brennan2025symbiosis} propose and compare different treatment assignments including user-side randomization (on an individual level or a cluster level) and two-sided randomization (of both users and items being recommended).
Guo et al.~\citet{guo2023evaluating} propose a two-phase experimental procedure where data is only shared in the first phase.
Si~\cite{si2024weighted} proposes a new training procedure where each data sample is weighted by its probability of being generated from the treatment or the control \ra, and each \ra is trained using a different loss that weighs data samples by these probabilities.
While we also study interference arising from shared data feedback loops, our work takes a different perspective: instead of aiming to correct for interference, we seek to characterize how it shapes decision-making.

\paragraph{Decision making in A/B experiments.} Our focus on the sign of the GTE for decision making shares similar spirit to Johari et al.~\cite{johari2025matter}, which studies interference due to constrained inventory in online marketplaces. Despite the fact that the difference-in-means (\dm) estimator is biased for the GTE, Johari et al.~\citet{johari2025matter} identify scenarios under which the estimated GTE and the true GTE have the same or opposite signs, and show that the biased \dm estimator surprisingly leads to a better hypothesis test in certain cases. Johari et al.~\citet{johari2023price} study the impact of interference on decision-making when conducting pricing experimentation. Ni~\citet{ni2025decision} provides the guidance on whether to roll out or shelve the change from a robust optimization perspective. 

\paragraph{Evaluation bias in recommendation systems.}
In our work, evaluation bias arises in recommendation systems due to shared data feedback loops, though it can also stem from other sources. Jadidinejad et al. \cite{jadidinejad2020closed} examine the effect of training and evaluating a recommender system using data collected from an existing system, a phenomenon termed ``closed feedback loops'', and show that increased exploration by the existing system reduces bias. Evaluation bias also emerges in experiments comparing ranking algorithms, such as those used in recommendation systems~\citep{ha-thuc2020ranking, nandy2021ranking}. For instance, in an A/B experiment where only 1\% of new content is assigned to a treatment that boosts its ranking, the position of this content differs from a setting where all new content receives the treatment. This mismatch introduces bias in estimating the GTE. Prior work addresses this by constructing a ``counterfactual" ranking in which the 1\% treated content is ranked as if all new content had received the treatment~\citep{ha-thuc2020ranking, nandy2021ranking, wang2023producer}. Zhu et al.~\cite{zhu2024feedback} study this counterfactual ranking in more complicated settings involving feedback loops induced by pacing algorithms. For example, when sellers in a marketplace have fixed budgets for advertising, the pacing algorithm will reduce the ranking of sellers who have already been recommended heavily. Goli et al.~\cite{goli2024ranking} propose an alternative debiasing approach that leverages historical data from previous A/B experiments.

\paragraph{Interference in platform experiments.}
Our work broadly fits within the literature on addressing interference in online platforms and marketplaces, where interference can arise from various sources beyond data feedback loops. The literature discussed here is not limited to recommendation systems.
Interference can result from competition in two-sided marketplaces \cite{li2022interference}, which can be handled through experimental design strategies such as two-sided randomization \citep{bajari2021multiple, johari2022experimental, ye2023cold}, bipartite experiments \citep{pouget2019variance, harshaw2023design}, clustered experiments \citep{eckles2017design, holtz2020reducing}, budget-split designs~\citep{liu2021trustworthy}, or ranking designs~\citep{habibi2025prioritized}. Interference may also stem from system dynamics, when treatments influence system states that subsequently affect outcomes in future periods, leading to both cross-unit and temporal interference. Switchback experiments are commonly used in this context \citep{glynn2020adaptive, hu2022switchback, bojinov2023design, xiong2024data}; when the system follows a Markovian structure, estimators such as the difference-in-Q estimator, the WDE estimator, and the DML estimator are applicable \citep{li2023experimenting, farias2022markovian, hays2025double}. Another source of interference is social networks, where outcomes may be affected through peer interactions. In such cases, methods include clustered randomized designs \cite{eckles2017design, ugander2023randomized}, two-stage experiments \cite{hudgens2008toward}, and estimators that explicitly model the network structure \cite{aronow2017estimating, cortez2023exploiting, leung2020treatment, li2022interference}; see~\citep{si2024weighted} for a comprehensive overview.

\paragraph{Ramp-up experiments under interference} Ramp-up experiments are widely used in industry to gradually increase exposure to new features or algorithms in order to manage risk and monitor performance, and are a standard component of large-scale experimentation platforms \citep{xiong2024optimal, kohavi2009controlled, kohavi2020trustworthy, xu2015infrastructure}. Beyond their operational role, recent work has shown that ramp-up experiments can be informative in settings with interference. Ramp-up experiments are used to detect the existence of interference~\citet{han2023detecting}, to aid identification and model selection~\citet{boyarsky2023modeling}, and to estimate the global treatment effect under network interference without requiring full knowledge of the interference graph~\citet{yu2022estimating}. Our work develops a new use of ramp-up experiments in the context of interference induced by shared data feedback loops when comparing learning algorithms. Specifically, we use changes in overall performance during ramp-up to diagnose whether an A/B comparison operates in the regime of sign violation or sign preservation, providing guidance on when the direction of an experimental conclusion can be trusted.
\subsubsection{Bandit problems}

We consider a problem formulation where each \rs is modeled by a bandit algorithm, and the two bandit algorithms share data with each other. Our formulation is related to the following lines of work in the bandit literature.

\paragraph{Bandits with auxiliary data.}
Shivaswamy and Joachims~\citet{shivaswamy2012history} consider the stochastic multi-armed bandit problem where the algorithm has access to a set of historical samples from each arm, and shows that $\Omega(\log\horizon)$ historical samples enable the algorithm to achieve a constant regret. 
Variants of this problem include allowing the algorithm to receive additional samples in each iteration by paying a certain cost (where the cost is allowed to be zero)~\citet{yun2018additional}, or allowing the algorithm  to ``abstain'' in an iteration, so that the sample is observed by the algorithm but not counted towards its regret~\citet{yang2024abstention}. In our formulation, the two algorithms receive ``auxiliary'' data from each other. Contrary to prior work, the treatment algorithm is not free to choose directly what arms are pulled in the auxiliary data which is determined by the control algorithm, and vice versa, though the treatment algorithm indirectly influences the auxiliary data supplied by the control algorithm through data sharing. Such dependencies between the two algorithms do not exist when the historical data is offline.

\paragraph{Multi-agent and multi-player bandits.}
In the multi-agent setup, each agent runs a bandit algorithm, and the goal is to minimize the sum of their regrets in a cooperative manner by sharing data with each other. Variants of the problem include assuming that agents have homogeneous~\cite{sankararaman2019social,he2022federated,barnea2024cooperative} or heterogeneous~~\cite{wang2021multitask,zhu2023distributed} rewards for the arms, or considering communication graph structures that describe how data is shared among agents~\cite{sankararaman2019social,barnea2024cooperative}.
The multi-player bandit problem, inspired by the application of cognitive radio networks, assumes that the reward for an arm becomes lower under a ``collision'' where multiple algorithms select this arm at the same time; see~\cite{boursier2024multi-player} for a survey. In multi-agent and multi-player bandit problems, the goal is to design algorithms that perform well in cooperative settings. In contrast, our goal is to evaluate an algorithm's performance when it runs in isolation, but we only observe the algorithm when it runs jointly with another algorithm.

\paragraph{Equilibria for game-theoretic agents.}
Prior work also considers algorithms that compete with each other in a game-theoretic fashion. 
On a simple two-armed bandit problem with a safe arm and a risky arm, Bolton and Harris~\citet{bolton1999experimentation} analyze the equilibria of multiple algorithms under data sharing, and identify similar effects to ours where agents are incentivized to free ride on other agents' exploration; see~\cite{horner2017experimentation} for a survey on strategic experimentation by agents that share information. Aridor et al.~\cite{aridor2025competing} consider running two bandit algorithms that compete for data. At each timestep, the two algorithms each recommend an arm. Each user chooses the arm that yields a higher reward for the user, and only the algorithm recommending this arm receives the user data. The paper reveals that such competition disincentivizes exploration and leads to a lower social welfare at the equilibrium. Jagadeesan et al.~\citet{jagadeesan2023competition} combine these two ideas of data sharing and user self-selection, and analyze their effects on the social welfare at the equilibrium.
Due to our A/B experiment setup, we do not allow users to choose their treatment assignment (i.e., which algorithm they use), and the algorithms of interest are not strategic in nature.

\paragraph{Corralling and online model selection.}
Corralling~\cite{agarwal2017corralling} or online model selection~\cite{foster2019selection,chatterji2020simultaneously} concerns designing a meta-algorithm that has access to a collection of base algorithms (or algorithm classes) and performs as well as the best base algorithm. At a high level, the meta-algorithm maintains some online estimates of the regret of each base algorithm in order to choose the best base algorithm to call, which is similar in spirit to our goal of comparing the performance of algorithms. In our problem, we do not have a meta-algorithm which decides which base algorithm a data sample is provided to, and we only consider the identification of the best base algorithm but not regret minimization for the meta-algorithm. However, it is interesting future work to design adaptive experiments where a meta-algorithm makes treatment assignments to users under data sharing.

\subsubsection{Intersection between causal inference and bandit algorithms}
Our work lies at the intersection of causal inference, experimentation, and bandit problems. A growing body of prior literature~\cite{zhang2020inference, hadad2021confidence, zhan2021off, han2022online,liang2023experimental, simchi2023multi} uses bandit algorithms for adaptive data collection to address causal questions that involve evaluating the performance of a policy or an arm. For evaluating policies, one objective is to perform off-policy evaluation, that is, to evaluate the policy's performance on the true distributions, only given biased data that is collected adaptively. For evaluating arms, the objective is to estimate the mean of each arm or to construct confidence intervals for these means. Prior work examines the bias in the sample mean of each arm due to adaptive sampling~\cite{nie2018why,shin2019or,shin2020conditional,shin2021bias}. This line of work uses bandit algorithms as a tool to evaluate other objects, whereas our goal is to evaluate bandit algorithms themselves.
In other words, we focus on comparing adaptive algorithms rather than fixed policies or arms.

\section{Problem formulation}
We formulate our A/B experiments under \datafeedback through a bandit framework.
Following the bandit literature, we use the standard notion of regret as our performance metric. The GTE is hence the difference between the regrets of the two algorithms.

\subsection{Stochastic multi-armed bandit}
We start by introducing the classic stochastic multi-armed bandit problem. Consider $\numarms \ge 2$ arms, where each arm $\idxarm\in [\numarms]$ is associated with an unknown distribution $\dist_\idxarm$. Every time arm $\idxarm$ is pulled, a stochastic reward is sampled from distribution $\dist_\idxarm$. We denote by $\meantrue_\idxarm \defn \Expect_{\reward \sample \dist_\idxarm}[\reward]$ the expected reward of arm $\idxarm$. For ease of exposition, we assume that each distribution $\dist_\idxarm$ is supported on $[0, 1]$ and there is a unique best arm $\idxarmbest\defn \argmax_{\idxarm\in [\numarms]} \meantrue_\idxarm$ with expected reward $\meanbest \defn \meanarm_{\idxarmbest}$. 
For any suboptimal arm $\idxarm\ne \idxarmbest$, we define the gap $\gap_\idxarm \defn \meanbest - \meanarm_\idxarm > 0$ as the difference between the expected reward of the best arm and that of arm $\idxarm$. We define the minimum gap $\gapmin \defn \min_{\idxarm\ne \idxarmbest} \gap_\idxarm > 0$.

A bandit algorithm aims to maximize the cumulative reward over a horizon of $\horizon\ge 1$ timesteps. At each timestep $\timestep\in [\horizon]$, the bandit algorithm selects an arm $\armselect_\timestep\in [\numarms]$ and receives a stochastic reward $\reward_\timestep \sample \dist_{\armselect_\timestep}$ from this arm. The arm selection is based on the history of all arm pulls and reward samples in the previous $(\timestep-1)$ timesteps. 
Formally, we denote the history at timestep $\timestep$ by $\history_{\timestep} = (\history_{1, \timestep}, \ldots, \history_{\numarms, \timestep})$, where $\history_{\idxarm, \timestep}$ denotes the set of reward samples collected from arm $\idxarm\in [\numarms]$ in the previous $(\timestep-1)$ timesteps. That is, we define $\history_{\idxarm, \timestep} \defn \{\reward_\idxtimestep: \armselect_\idxtimestep = \idxarm, \idxtimestep \in [\timestep - 1]\}$. In words, we consider a class of bandit algorithms that select arms based on the set of reward samples for each arm so far, but are oblivious to the order of the past arm pulls (or reward samples). Many classic bandit algorithms belong to this general class, including the \egreedy algorithm, the Upper Confidence Bound (UCB) algorithm, the EXP3 algorithm, and Thompson sampling. For any algorithm $\algo$ in this class, we write the arm selected by this algorithm at timestep $\timestep$ by $\armselect_\timestep = \algo(\history_{\timestep})$.

The performance of an algorithm $\algo$ is measured by its regret, namely, the difference between the highest possible cumulative reward in expectation and the actual cumulative reward obtained:
\begin{align*}
    \regret_\horizon(\algo) \defn \horizon \meanbest - \sum_{\timestep \in [\horizon]} \reward_{\armselect_\timestep}.
\end{align*}
We are often interested in the expected regret
\begin{align*}
    \Expect[\regret_\horizon(\algo)] = \horizon \meanbest - \Expect\big[\sum_{\timestep \in [\horizon]} \reward_{\armselect_\timestep}\big],
\end{align*}
where the expectation is taken over the stochastic rewards and potential randomness of the algorithm.

\subsection{Bandit algorithms under \datafeedback}

To set up our bandit formulation under \datafeedback, let us use a running example where an online platform runs a recommendation algorithm to match cleaning companies with users in need. Users arrive at the platform sequentially. When a user arrives, the recommendation algorithm recommends one of the $\numarms$ cleaning companies to the user, and for simplicity we assume that the user always hires the cleaning company that is recommended. 
If a user hires company $\idxarm\in [\numarms]$, then the user gives this company a rating $\reward\sample \dist_\idxarm$ to indicate their satisfaction, where the distribution $\dist_\idxarm$ represents the quality of company $\idxarm$. The recommendation algorithm, without extra knowledge about the companies, makes recommendations solely based on the ratings from previous users.

We formulate an A/B experiment comparing two recommendation algorithms as a bandit problem with $2\numusers$ users. We assume that two users arrive at the platform at each timestep $\timestep\in [\horizon]$. These two users are assigned to the treatment or the control algorithm one each uniformly at random.
To model data sharing between the treatment and control algorithms, we assume that the two algorithms make recommendations based on past ratings from users in both groups. Formally, consider two bandit algorithms $\algo_1$ and $\algo_2$ that share history $\history_{\timestep}$ over the entire course of $\horizon$ timesteps. At each timestep $\timestep\in [\horizon]$, each of the two algorithms independently chooses the arm $\algo_1(\history_\timestep)$ and $\algo_2(\history_\timestep)$ to pull based on history $\history_{\timestep}$, observes a reward sample from the chosen arm, and adds this reward sample to the shared history $\history_\timestep$. Thus, two reward samples are added to the shared history at each timestep $\timestep$, and the total number of samples included in history $\history_{\timestep}$ is $2(\timestep-1)$. This procedure of running two algorithms under \datafeedback is formalized in Algorithm~\ref{algo:joint}. 

\begin{algorithm}[tb]
    \DontPrintSemicolon
    
    Initialize $\history_1 = (\history_{1, 1}, \ldots, \history_{\numarms, 1})$ with $\history_{1, 1} = \ldots = \history_{\numarms, 1} = \emptyset$\;
    \For{\timestep=1, \ldots, \horizon}{
        Algorithm $\algo_1$ selects arm $\armselect_1 = \algo_1(\history_{\timestep})$ and observes reward $\reward_{1, \timestep}\sample \dist_{\armselect_1}$\;
        Algorithm $\algo_2$ selects arm $\armselect_2 = \algo_2(\history_{\timestep})$ and observes reward $\reward_{2, \timestep} \sample \dist_{\armselect_2}$\;
        Add the two reward samples to the history:
        \begin{align*}
            & \history_{\armselect_1, \timestep} \leftarrow \history_{\armselect_1, \timestep} \union \{\reward_{1, \timestep}\}\\
            & \history_{\armselect_2, \timestep} \leftarrow \history_{\armselect_2, \timestep} \union \{\reward_{2, \timestep}\}\\
            & \history_{\timestep+1}\leftarrow \history_{\timestep}
        \end{align*}
    }
    \caption{Running two algorithms $\algo_1$ and $\algo_2$ under \datafeedback.
    \label{algo:joint}
}

\end{algorithm}

\subsection{Comparing two algorithms by their regrets}

We consider estimating the GTE for a pair of \ras in A/B experiments by comparing the regret for two bandit algorithms. 
The platform is interested in designing an algorithm $\algo$ that maximizes the expected overall satisfaction $\Expect[\sum_{\timestep=1}^{2\numusers} \reward_\timestep]$ if algorithm $\algo$ is deployed to all the $2\horizon$ users, or equivalently, minimizes the expected regret $\Expect[\regret_{2\numusers}(\algo)]$. The GTE concerns the regret difference between the two algorithms:
\begin{align*}
    \GTE = \Expect[\regret_{2\numusers}(\algo_1)] - \Expect[\regret_{2\numusers}(\algo_2)].
\end{align*}
In  A/B experiments, we measure the regret incurred by each of the two algorithms under data sharing, defined by
\begin{align*}
    & \regret_\horizon(\algo_1\given (\algo_1, \algo_2) ) \defn \horizon \meanbest - \sum_{\timestep\in [\horizon]} \reward_{1, \timestep}\\
    & \regret_\horizon(\algo_2 \given (\algo_1, \algo_2)) \defn \horizon\meanbest - \sum_{\timestep\in [\horizon]} \reward_{2, \timestep}.
\end{align*}
where $\reward_{1, \timestep}$ and $\reward_{2, \timestep}$ are rewards collected by the two algorithms respectively according to Algorithm~\ref{algo:joint}. We use the notation $\regret_\horizon(\algo_1\given (\algo_1, \algo_2))$ and $\regret_\horizon(\algo_2\given (\algo_1, \algo_2))$ to emphasize that when running algorithms $1$ and $2$ under data sharing, the regret of each algorithm depends on both algorithms, because the arms selected by algorithm $1$ depend on history $\history_{\timestep}$, which in turn depend on the arms selected by algorithm $2$ in the previous timesteps, and vice versa. In general, the expected regret incurred by running an algorithm individually does not necessarily equal to the regret when running this algorithm jointly with another algorithm. That is, 
\begin{subequations}\label{eq:gte_not_equal}
\begin{align}
    \Expect[\regret_\horizon(\algo_1\given (\algo_1, \algo_2))] & \ne \Expect[\regret_\horizon(\algo_1)]\\
    \Expect[\regret_\horizon(\algo_2\given (\algo_1, \algo_2))] & \ne \Expect[\regret_\horizon(\algo_2)].
\end{align}
\end{subequations}
Moreover, our theoretical results to follow suggest that these regrets with or without another algorithm running jointly can even have different rates.
The \dm estimator, denoted by $\estGTE$, computes the difference between the empirical overall satisfaction of users under the treatment and control algorithms:
\begin{align*}
    \estGTE = \regret_\numusers(\algo_1\given (\algo_1, \algo_2)) - \regret_\numusers(\algo_2\given (\algo_1, \algo_2)).
\end{align*}
We consider the GTE estimate in expectation:
\begin{align*}
    \Expect[\estGTE] = \Expect[\regret_\numusers(\algo_1\given (\algo_1, \algo_2))] - \Expect[\regret_\numusers(\algo_2\given (\algo_1, \algo_2))].
\end{align*}

For decision-making purposes, the platform is primarily interested in determining whether algorithm 1 or algorithm 2 performs better. Although $\Expect[\estGTE] \ne \GTE$ in general due to~\eqref{eq:gte_not_equal}, we focus on characterizing the conditions under which $\Expect[\estGTE]$ and $\GTE$ have the same sign (termed ``sign preservation'') or opposite signs (termed ``sign violation''). Under sign preservation, the sign of $\Expect[\estGTE]$ correctly indicates which algorithm is better. Under sign violation, however, the sign of $\Expect[\estGTE]$ leads to an incorrect comparison between the two algorithms.

\section{Two regimes: sign violation and sign preservation}\label{sec:regimes}
We now present our theoretical results that characterize various cases for sign violation and sign preservation. 
We use the big-O notation for the regret rate of algorithms in terms of the horizon $\horizon$. The notation $f(\horizon) = O(g(\horizon))$ (resp. $f(\horizon) = \Omega(g(\horizon))$) means that there exists a positive constant $\const > 0$ such that $f(\horizon) \le \const \cdot g(\horizon)$ (resp. $f(\horizon) \ge \const \cdot g(\horizon)$) for all $\horizon\ge 1$. We use the notation $f(\horizon) = \Theta(g(\horizon))$ when both $f(\horizon) = O(g(\horizon))$ and $f(\horizon) = \Omega(g(\horizon))$ hold. We also use the little-o notation where $f(\horizon) = o(g(\horizon))$ means that for every $\const > 0$, there exists a constant $\horizon_0\ge 1$ such that $f(\horizon) \le \const \cdot g(\horizon)$ for all $\horizon \ge \horizon_0$. 
In all our results, we treat the problem parameters $\numarms$ (the number of arms) and $\{\gap_\idxarm\}_{\idxarm\in [\numarms]}$ (the arm gaps), and the algorithm parameters $\paramexplore$ (the exploration level associated with the UCB and \egreedy algorithms) and $C$ (the multiplicative factor for the exploration probability associated with the \egreedy algorithm) all as constants. That is, the constant $\const$ in the big-O and little-o notation is allowed to depend on all these parameters, as long as it is independent from $\horizon$.
The proofs of all theoretical results are included in \Cref{app:proofs}.

\subsection{Preliminaries on classic bandit algorithms}\label{sec:classic_algorithms}
We consider the following classes of classic bandit algorithms. We denote by $\abs*{\history_{\idxarm, \timestep}}$ the number of samples for arm $\idxarm$ up to timestep $\timestep$, and denote by $\abs*{\history_{\timestep}} \defn \sum_{\idxarm\in [\numarms]} \abs*{\history_{\idxarm, \timestep}} = 2(t-1)$ the total number of samples so far across all arms.
\begin{enumerate}
    \item The \textbf{greedy} algorithm (denoted by $\algogreedy$), selects the arm with the highest empirical mean\footnote{Ties are broken arbitrarily for all algorithms.}:
    \begin{align*}
        \algogreedy(\history_{\timestep}) = \argmax_{\idxarm\in [\numarms]} \frac{1}{\abs*{\history_{\idxarm, \timestep}}} \sum_{y\in \history_{\idxarm, \timestep}} y.
    \end{align*}

    \item The \textbf{\egreedy} algorithm (denoted by $\algoegreedy$) maintains a sequence of probabilities $\{\probexplore_\timestep\}_{\timestep\in [\horizon]}$ with each $\probexplore_\timestep\in [0, 1]$. At each timestep $\timestep\in [\horizon]$, the \egreedy algorithm selects one of the $\numarms$ arms uniformly at random with probability $\probexplore_\timestep$, and otherwise follows the greedy algorithm with probability $(1-\probexplore_\timestep)$.
    We define a class of \egreedy algorithms (denoted by $\algoegreedy_{\alpha,\Const}$) parameterized by a parameter $\paramexplore\in [0, 1]$ and a constant $\Const > 0$, where the exploration probability is defined by $\probexplore_\timestep= \min \left\{1, \frac{\Const}{\abs*{\history_{\timestep}}^{1-\paramexplore}}\right\}$.

    \item The \textbf{Upper Confidence Bound} algorithm (denoted by $\algoucb_\paramexplore$) maintains a confidence bound for the mean of each arm parameterized by a parameter $\paramexplore\in [0, 1]$, and selects the arm with the highest empirical mean under optimism:
    \begin{subequations}\label{eq:ucb_def_conf_interval}
    \begin{align}
        \algoucb_\paramexplore(\history_{\timestep}) = \argmax_{\idxarm\in [\numarms]} \frac{1}{\abs*{\history_{\idxarm, \timestep}}} \sum_{y\in \history_{\idxarm, \timestep}} y + \sqrt{\frac{2(\abs*{\history_{\timestep}}^\paramexplore - 1)}{\paramexplore\abs*{\history_{\idxarm, \timestep}}}}.
    \end{align}
    When $\paramexplore= 0$, we use the fact that $\lim_{\paramexplore\rightarrow 0} \frac{\kappa^\paramexplore-1}{\paramexplore} = \log \kappa$ for any real-valued $\kappa > 0$, so that
    \begin{align}
        \algoucb_{0}(\history_{\timestep}) = \argmax_{\idxarm\in [\numarms]} \frac{1}{\abs*{\history_{\idxarm, \timestep}}} \sum_{y\in \history_{\idxarm, \timestep}} y + \sqrt{\frac{2\log\abs*{\history_{\timestep}}}{\abs*{\history_{\idxarm, \timestep}}}}.
    \end{align}
    \end{subequations}
\end{enumerate}
A few remarks are in order. First, for ease of exposition, we focus on algorithms that depend only on history $\history_\timestep$ and are oblivious to the horizon $\horizon$. Such algorithms that do not require prior knowledge of the horizon are referred to as ``anytime'' algorithms.
Second, we define the empirical mean of an empty set to be infinity. For the UCB algorithm, we set the confidence interval length of arm $\idxarm$ to be infinity if $\history_{\idxarm, \timestep}$ is empty. For the \egreedy algorithm, if some $\history_{k,t}$ is empty, the algorithm is required to forgo exploration and instead pull an arm with no history. Under this setup, for all the three algorithms considered above, each arm is guaranteed to be pulled at least once within the first $\numarms$ timesteps, regardless of whether the algorithms share data or run individually.

For both the \egreedy algorithm and the UCB algorithm, a higher value of $\paramexplore$ corresponds to more exploration, as shown by the higher exploration probability in the \egreedy algorithm, and a larger confidence bound in the UCB algorithm. Setting $\paramexplore=0$ yields the most common \egreedy algorithm with $\probexplore = \Theta(\frac{1}{\timestep})$ and UCB algorithm with a confidence interval of length $\Theta(\sqrt{\log\timestep / \abs{\history_{\idxarm, \timestep}}})$, both of which incur an optimal $O(\log\horizon)$ regret. Hence, $\paramexplore=0$ corresponds to the optimal level of exploration.
Different values for the parameter $\paramexplore$ capture algorithms with different regret performances, formalized in the following result.

\begin{proposition}\label{prop:run_individually}
Consider running an algorithm individually without data sharing.
For any $0 \le \paramexplore \le 1$, the expected regret of the UCB algorithm $\algoucb_\paramexplore$ satisfies
\begin{align*}
    \Expect[\regret_T(\algoucb_\paramexplore)] =\begin{cases}
        \Theta(\log \horizon) &\text{if } \paramexplore = 0,\\
        \Theta(\horizon^\paramexplore) &\text{if } 0 < \paramexplore \le 1.
    \end{cases}
\end{align*}
The expected regret of the \egreedy algorithm $\algoegreedy_{\paramexplore, \Const}$ with $C\ge\max\left\{120K,\frac{16K}{\gapmin^2}\right\}$ satisfies
\begin{align*}
    \Expect[\regret_T(\algoegreedy_{\paramexplore,\Const})] =\begin{cases}
         \Theta(\log \horizon) &\text{if } \paramexplore = 0,\\
         \Theta(\horizon^\paramexplore) &\text{if } 0 < \paramexplore \le 1.
    \end{cases}
\end{align*}
\end{proposition}
The proof of this proposition is outlined in Appendix~\ref{sec:proof_prop_run_individually} for completeness.
For the \egreedy algorithm, the upper bound of $O(\log\horizon)$ when $\paramexplore= 0$ is proved in~\cite[Theorem 3]{auer2002finite}. The proof of the upper bound $O(\horizon^\paramexplore)$ when $0 < \paramexplore \le 1$ follows the same steps.
For the upper bound of the UCB algorithm, our proof is similar to~\cite[Appendix C]{guo2022demonstrator} which shows $O(\horizon^\paramexplore)$ regret for the non-anytime UCB algorithm using the confidence interval $\sqrt{\frac{2(\horizon^\paramexplore-1)}{\paramexplore\abs*{\history_{\idxarm, \timestep}}}}$ instead of $\sqrt{\frac{2(\abs{\history_\timestep}^\paramexplore-1)}{\paramexplore\abs*{\history_{\idxarm, \timestep}}}}$. 

\subsection{A general observation}

We make the following observation that connects the exploration-exploitation tradeoff of the two algorithms and the regimes of sign preservation and sign violation.

\begin{observation}
When comparing two bandit algorithms, whether $\Expect[\estGTE]$ and $\GTE$ have the same sign is driven by the exploration-exploitation tradeoff of the two algorithms.
\begin{itemize}
\item If the worse algorithm (when run individually without data sharing) underperforms because it over-explores, then sign preservation occurs under data sharing. That is, $\Expect[\estGTE]$ and $\GTE$ have the same sign.  
\item If the worse algorithm  (when run individually without data sharing) underperforms because it over-exploits, then sign violation occurs under data sharing. That is, $\Expect[\estGTE]$ and $\GTE$ have opposite signs.
\end{itemize}
\end{observation}

In what follows, we formally establish this result by analyzing the greedy algorithm as an example of over-exploitation, and the parameterized classes of the greedy and UCB algorithms (with $0<\alpha \le 1$) as examples of over-exploration.

\subsection{Sign violation}\label{sec:results_sign_violation}

We now present cases where interference from data sharing causes $\GTE$ and $\Expect[\estGTE]$ to have opposite signs. We consider the greedy algorithm that is known to incur a linear regret when running individually. Our first result shows that when the greedy algorithm runs jointly with the \egreedy algorithm, its regret reduces to $O(1)$, whereas the regret of the \egreedy algorithm remains the same rate.

\begin{theorem} \label{thm:greedy+e-greedy}
When we jointly run the greedy algorithm $\algogreedy$ and the \egreedy algorithm $\algoegreedy_{\alpha, C}$ with $C\ge\max \left\{80K,\frac{32K}{\gapmin^2}\right\}$, the expected regrets of the two algorithms under data sharing satisfy
\begin{subequations}
\begin{align}
     & \EE{R_T(\algogreedy\mid(\algogreedy,\algoegreedy_{\alpha, \Const}))} = O(1),\label{eq:thm_violate_grd_egreedy_ub}\\[2ex]
     & \EE{R_T(\algoegreedy_{\alpha, \Const} \given (\algogreedy,\algoegreedy_{\alpha, \Const}))} = \begin{cases}
        \Omega( \log T) & \text{if } \alpha = 0,\\
        \Omega(T^\alpha) & \text{if } 0 < \alpha \le 1.\label{eq:thm_violate_egreedy_greedy_lb}
\end{cases}
\end{align}
\end{subequations}
\end{theorem}

The proof of this theorem is provided in \Cref{sec:proof_thm_greedy+e-greedy}.
For the greedy algorithm, its regret decreases from being linear to a constant when running jointly with the \egreedy algorithm. For the \egreedy algorithm, combining Theorem~\ref{thm:greedy+e-greedy} and Proposition~\ref{prop:run_individually}, we find that its regret remains $\Omega(T^\alpha)$ (or $\Omega(\log T)$ when $\alpha=0$), with or without data sharing. Thus, the expected GTE estimate $\Expect[\estGTE]$ under \datafeedback is $-\Omega(T^\alpha)$ (or $-\Omega(\log T)$ when $\paramexplore=0$), while the true GTE is $\Theta(T)$, giving a sign reversal.

Intuitively, the linear regret of the greedy algorithm stems from its lack of exploration. When paired with the \egreedy algorithm, it benefits from \datafeedback and effectively receives exploration samples at no cost. In this sense, data sharing provides a ``free ride'' for the greedy algorithm.
Following this intuition of a free ride for the greedy algorithm, we demonstrate a similar result when pairing the greedy algorithm with the UCB algorithm.

\begin{theorem}\label{thm:greedy+UCB}
When we jointly run the greedy algorithm $\algogreedy$ and the UCB algorithm $\algoucb_\alpha$, the expected regrets of the two algorithms under data sharing satisfy 
\begin{subequations}
\begin{align}
    & \EE{R_T(\algogreedy \mid (\algogreedy, \algoucb_\alpha))} 
    = O(1), \label{eq:thm_violate_ucb_greedy_ub}\\[2ex]
    &\EE{R_T(\algoucb_\alpha \mid (\algogreedy, \algoucb_\alpha))} 
    = \begin{cases} \Omega(\log T) & \text{if } \alpha = 0,\\
    \Omega(T^\alpha) & \text{if } 0 < \alpha \le 1.
    \end{cases}\label{eq:thm_violate_ucb_greedy_lb}
\end{align}
\end{subequations}
\end{theorem}
The proof of this theorem is provided in \Cref{sec:proof_thm_greedy+UCB}. Similar to Theorem~\ref{thm:greedy+e-greedy}, we find that the linear regret of the greedy algorithm decreases to a constant when running jointly with the UCB algorithm, while the regret of the UCB algorithm remains the same order, with or without data sharing. The expected GTE estimate under \datafeedback is again $-\Omega(T^\alpha)$ (or $-\Omega(\log T)$ when $\paramexplore=0$) whereas the true GTE is $\Theta(T)$, indicating opposite signs between the expected GTE estimate and the true GTE.

Generalizing these two examples, we now define a general class of anytime algorithms. When the greedy algorithm runs jointly with any algorithm in this class under data sharing, the greedy algorithm incurs only a constant regret.
For any two algorithms $\algo$ and $\algo'$, we denote by $N_{k,\timestep}(\algo\given (\algo, \algo'))$ the number of pulls of arm $k$ by algorithm $\algo$ before timestep $\timestep$, when it runs jointly with algorithm $\algo'$.
\begin{theorem}\label{thm:greedy+general}
Let $\const > 0$ be a constant that depends only on $\numarms, \{\gap_\idxarm\}_{\idxarm\in [\numarms]}, \paramexplore$ and $C$. Consider jointly running the greedy algorithm with any anytime algorithm $\algo$. Suppose that for any $\horizon\ge 1$, the algorithm $\algo$ satisfies
\begin{align}
    \Prob\left(\varnumpulls_{\idxarmbest, \horizon}(\algo\given (\algo, \algogreedy)) \le \frac{4}{\gapmin^2}\log\horizon\right) < \frac{\const}{\horizon^2}.\label{eq:assume_high_probability_bound}
\end{align}
Then under data sharing, the regret incurred by the greedy algorithm satisfies
\begin{align*}
    \Expect\left[\regret_\horizon(\algogreedy\given (\algo, \algogreedy))\right] <\const',
\end{align*}
where $\const' > 0$ is a constant that depends only on the constant $\const$ and the arm gaps $\{\gap_\idxarm\}_{\idxarm\in [\numarms]}$.
\end{theorem}
The proof of this theorem is provided in Appendix~\ref{sec:proof_thm_greedy+general}. The high-probability bound~\eqref{eq:assume_high_probability_bound} is a mild condition, as it concerns only the number of times that the \textit{optimal} arm is pulled. For any algorithm with a sublinear regret, the optimal arm is expected to be pulled $\horizon-o(\horizon)$ times. The bound~\eqref{eq:assume_high_probability_bound} merely requires the optimal arm to be pulled $\Omega(\log\horizon)$ times with high probability.
Theorems~\ref{thm:greedy+e-greedy} and \ref{thm:greedy+UCB} are established by proving condition~\eqref{eq:assume_high_probability_bound} for the \egreedy and UCB algorithms, respectively. 
In Appendix~\ref{app:bandit_algorithms}, we consider a variant setting with one-way data sharing, where one algorithm runs individually using its own data, whereas the other algorithm has access to the data by both algorithms. Under this one-way data sharing, we provide additional results for the EXP3 algorithm.

We emphasize that the $O(1)$ regret for the greedy algorithm is non-trivial to show, as we cannot rely on standard regret analysis in the literature. In the literature, the desired regret bound is often of order $\log\horizon$. It then suffices to consider the case when a suboptimal arm has already been pulled $\Omega(\log\horizon)$ times---by which point the empirical mean of this arm nicely concentrates around its true mean---and bound the number of additional pulls. In contrast, to show a constant regret, we need to bound the number of pulls on each suboptimal arm from the beginning, before concentration happens to the empirical means of the arms. A natural alterative approach is to consider the case when a suboptimal is pulled by \emph{the other algorithm} an order of $\log\horizon$ times, and bound the additional number of pulls by the greedy algorithm. When running a single algorithm, prior work has shown that if there are $\Omega(\log \horizon)$ samples of auxiliary historical data for a suboptimal arm $\idxarm$, then this arm will only be pulled a constant number of additional times, resulting in a constant regret~\cite{shivaswamy2012history}. A similar argument may show that the greedy algorithm pulls the suboptimal arm a constant number of times if the other algorithm has already pulled it $\Omega(\log \horizon)$ times. However, if the other algorithm has $O(\log \horizon)$ regret (say the UCB algorithm), it will take $\Omega(\horizon)$ timesteps before it pulls any suboptimal arm $\Omega(\log \horizon)$ times. By that point, the greedy algorithm may have already incurred a linear regret! Therefore, our proof requires new ways to carefully track the regret as a function of the number of existing arm pulls. 

Our proof strategy is to decompose a pull on any suboptimal arm $\idxarm$ by considering two causes: the empirical mean of arm $\idxarm$ does not concentrate, or the empirical mean of the optimal arm $\idxarmbest$ does not concentrate. In the latter case, condition~\eqref{eq:assume_high_probability_bound} ensures that algorithm $\algo$ supplies $\Omega(\log\horizon)$ samples on the optimal arm $\idxarmbest$, so that the probability that the empirical mean of the optimal arm $\idxarmbest$ does not concentrate is low. In the former case, one concern is that the greedy algorithm is ``stuck'' and keeps pulling arm $\idxarm$ for a long time. However, each pull on arm $\idxarm$ contributes towards the concentration of its empirical mean: considering the event where ``the greedy algorithm pulls arm $\idxarm$ because the empirical mean of arm $\idxarm$ does not concentrate'' allows us to take a union bound over the number of samples to control the probability of non-concentration of this arm, as opposed to over the timesteps (where the number of samples may stay the same over many timesteps).

\subsection{Sign preservation}\label{sec:results_sign_preservation}
The previous results illustrate scenarios in which data sharing flips the sign of the expected GTE estimate. A natural follow-up question is whether there exist settings where data sharing does not affect the resulting comparison. We show that if both algorithms belong to the parameterized class of \egreedy or UCB algorithms, then the sign of the expected GTE estimate is preserved.

\begin{theorem}\label{thm:sign-preserve-combined}
When we jointly run two algorithms 
$\algo_{\alpha_1} \in \{ \algoegreedy_{\alpha_1, C}, \, \algoucb_{\alpha_1}\}$, $\algo_{\alpha_2} \in \{ \algoegreedy_{\alpha_2, C}, \, \algoucb_{\alpha_2}\}$, 
with $0 \le \alpha_1 < \alpha_2 \le 1$ and $C \ge \max\{120K,\, 32K/\gapmin^2\}$,  
the expected regrets of the two algorithms under data sharing satisfy
\begin{alignat*}{3}
&\EE{R_T(\algo_{\alpha_1} \given (\algo_{\alpha_1}, \algo_{\alpha_2}))}  =
    \begin{cases}
        O(\log T), & \text{if } \alpha_1 = 0,\\
        O(T^{\alpha_1}), & \text{if } 0 < \alpha_1 \le 1,
    \end{cases}\\[2ex]
&\EE{R_T(\algo_{\alpha_2} \given (\algo_{\alpha_1}, \algo_{\alpha_2}))} 
    = \Theta(T^{\alpha_2}).
\end{alignat*}
\end{theorem}
The proof of this theorem is provided in \Cref{sec:proof_thm_sign_preservev_combined}. Looking at Theorem~\ref{thm:sign-preserve-combined} together with Proposition~\ref{prop:run_individually}, we find that under data sharing, the regret of $\algo_{\alpha_1}$ remains $O(T^{\alpha_1})$ (or $O(\log T)$ when $\alpha_1=0$), while the regret of $\algo_{\alpha_2}$ remains $\Theta(T^{\alpha_2})$. As a result, the sign of the expected GTE estimate is preserved when comparing two such algorithms: the one with the larger value of $\alpha$ incurs the higher regret, both with and without data sharing. Intuitively, $\algo_{\alpha_2}$ performs worse because it over-explores: it explores more than $\algo_{\alpha_1}$ and, in particular, more than the optimal rate, regardless of data sharing. This excess exploration accounts for the majority of its regret. For \egreedy, the probability of selecting each arm under $\algo_{\alpha_2}$ is always bounded below by its exploration rate $\epsilon_t$, so $\algoegreedy_{\alpha_2, C}$ continues to over-explore under data sharing. For UCB, $\algoucb_{\alpha_2}$ requires approximately $\Theta(T^{\alpha_2})$ samples of a suboptimal arm to achieve a sufficiently small confidence bound, whereas the other algorithm $\algo_{\alpha_1}$ contributes only $O(T^{\alpha_1})$ such samples; thus, it also over-explores under data sharing.

The result shows that if both algorithms sufficiently explore to achieve sublinear regrets when running individually, the sign of the expected GTE estimate is preserved under \datafeedback as their performance is determined by the level of exploration, which is inherent to the algorithms and thus unaffected by sharing data.

\section{Detecting sign violation with ramp-up experiments}\label{sec:guidelines}

In practice, we often do not know a priori whether an algorithm over-explores or over-exploits before running an experiment (which is why we need experiments to start with!). That is, we cannot determine whether an experiment lies in the regime of sign preservation or sign violation, and thus do not know how to interpret the comparison results from the experiment. In this section, we provide a procedure to determine whether an experiment lies in the sign violation or sign preservation regime.

\subsection{Detection procedure}
We propose to detect potential sign violation using ramp-up A/B experiments. We first extend the previous data sharing scheme with a fixed 50–50 split in Algorithm~\ref{algo:joint} to a general Bernoulli assignment in Algorithm~\ref{algo:joint_bernoulli}, where each user is independently assigned to treatment with probability $\probassign$. A ramp-up experiment then consists of a sequence of tests with treatment probabilities $0 < \probassign_1 < \probassign_2 < \cdots$, where Algorithm~\ref{algo:joint_bernoulli} is first run with treatment probability $\probassign_1$, and the treatment probability is subsequently increased to $\probassign_2$, and then $\probassign_3$, and so on. In online experimentation, A/B experiments are often run in this ramp-up fashion, starting with a low treatment probability, typically for a new algorithm that has not been deployed before, in order to monitor performance and detect unintended side effects at low exposure levels before broader deployment \citep{kohavi2020trustworthy}.

\begin{algorithm}[tb]
    \DontPrintSemicolon
    
    Initialize $\history_1 = (\history_{1, 1}, \ldots, \history_{\numarms, 1})$ with $\history_{1, 1} = \ldots = \history_{\numarms, 1} = \emptyset$\;
    \For{\timestep=1, \ldots, 2\horizon}{
        Sample $\assignment_\timestep\sample \Bernoulli(p)$\;
        \uIf{$\assignment_\timestep = 0$}{
        Algorithm $\algo_1$ selects arm $\armselect = \algo_1(\history_{\timestep})$ and observes reward $\reward_{\timestep}\sample \dist_{\armselect}$\;}
        \Else{Algorithm $\algo_2$ selects arm $\armselect = \algo_2(\history_{\timestep})$ and observes reward $\reward_{\timestep} \sample \dist_{\armselect}$\;}
        Add the reward sample to the history:
        \begin{align*}
            & \history_{\armselect, \timestep} \leftarrow \history_{\armselect, \timestep} \union \{\reward_{\timestep}\}\\
            & \history_{\timestep+1}\leftarrow \history_{\timestep}
        \end{align*}
    }
    \caption{Running two algorithms $\algo_1$ and $\algo_2$ under \datafeedback, with a Bernoulli assignment (probability $(1-\probassign)$ for $\algo_1$ and probability $\probassign$ for $\algo_2$).
    \label{algo:joint_bernoulli}
}
\end{algorithm}

We now propose a simple warning system that alerts experimenters when they may be operating in a sign violation regime. The key idea is to monitor how the overall average outcome evolves as the treatment probability increases during a ramp-up experiment. Concretely, suppose that at a baseline allocation, Algorithm~2 appears to outperform Algorithm~1. As the share of traffic assigned to Algorithm~2 increases, one would typically expect the overall average outcome to improve. A decline in the overall average outcome as the traffic share of Algorithm~2 increases suggests a potential sign violation.

Formally, consider a Bernoulli assignment with treatment probability~$p$. At each timestep $t$, let
$W_t \sim \Bernoulli(p)$
denote the treatment assignment, where $W_t=1$ corresponds to Algorithm~2 and $W_t=0$ to Algorithm~1. Define the inverse probability weighted (IPW) estimators of the mean outcome for each algorithm as
\[ \widehat{Y}^p(\algo_2) = \frac{\sum_{\timestep\in [2\horizon]}\reward_\timestep W_t}{ 2\horizon p }, \qquad
\widehat{Y}^p(\algo_1) = \frac{\sum_{\timestep\in [2\horizon]}\reward_\timestep (1-W_t)}{2\horizon (1-p)}, \]
and the overall average outcome as
\[ \overline{Y}^p = \frac{1}{2T}\sum_{\timestep\in [2\horizon]}\reward_\timestep. \]
The detection rule is as follows:
\begin{quotex}
\emph{In a ramp-up experiment, if increasing the traffic assigned to the empirically better algorithm results in a decline in the overall average outcome, we flag this behavior as a potential sign violation and recommend further investigation.}
\end{quotex}
Mathematically, we flag a potential sign violation if there exist treatment probabilities $p < p'$ such that
\begin{align}
    \sign\p{\widehat{Y}^{0.5}(\algo_2) - \widehat{Y}^{0.5}(\algo_1)} \neq \sign\p{\overline{Y}^{p'} - \overline{Y}^{p}},\label{eq:detection_rule}
\end{align}
The intuition behind this diagnostic is as follows. When increasing the probability on the better algorithm worsens the average outcome, a likely cause is that the better algorithm free rides the other algorithm, and once the traffic of the other algorithm decreases, the level of exploration is no longer sufficient for the (seemingly) better algorithm to perform well. For example, consider the UCB algorithm with the greedy algorithm. Under data sharing, the greedy algorithm appears to perform better than the UCB algorithm, yielding a higher average outcome. However, when the greedy algorithm is assigned a large share of traffic, the UCB algorithm now only takes a very small share of traffic and no longer provides sufficient exploration that is needed by the greedy algorithm, resulting in a high regret and, consequently, a low average outcome.

In contrast, consider a second example comparing two UCB algorithms,  $\algoucb_0$ and $\algoucb_{0.5}$. $\algoucb_0$ already explores sufficiently on its own for any treatment probability. Increasing the share of $\algoucb_0$ therefore reduces unnecessary exploration by $\algoucb_{0.5}$, which reduces the total regret and improves the average outcome. The sign terms on the left- and right-hand sides of the detection criterion~\eqref{eq:detection_rule} coincide, indicating sign preservation.

\subsection{Theoretical justifications of the detection procedure}

We now provide theoretical justifications for the proposed detection procedure. In Algorithm~\ref{algo:joint_bernoulli} under a Bernoulli assignment with probability $\probassign$, we define the regrets of the two algorithms:
\begin{align*}
    \regret^\probassign(\algo_1 \given (\algo_1, \algo_2)) &= \sum_{\timestep\in [2\horizon]: \assignment_\timestep = 0} (\meanbest-\reward_\timestep), \\
    \regret^\probassign(\algo_2 \given (\algo_1, \algo_2)) &= \sum_{\timestep\in [2\horizon]: \assignment_\timestep = 1} (\meanbest-\reward_\timestep),
\end{align*}
with $\assignment_\timestep \sim \Bernoulli(\probassign)$. We further define the total regret:
\begin{align*}
    \regret^\probassign(\algotot\given (\algo_1, \algo_2)) &=  \regret^\probassign(\algo_1 \given (\algo_1, \algo_2)) +  \regret^\probassign(\algo_2 \given (\algo_1, \algo_2))\\
    &= 2\horizon\meanbest - \sum_{\timestep\in [2\horizon]}\reward_\timestep.
\end{align*}
We note that the detection procedure~\eqref{eq:detection_rule} is equivalent to the following criterion. We flag a potential sign violation if there exist $\probassign < \probassign'$ such that
\begin{align}
    \sign\big(\regret^{0.5}(\algo_2 \given (\algo_1, \algo_2)) - \regret^{0.5}(\algo_1 \given (\algo_1, \algo_2)\big) 
    \ne \sign\big(\regret^{\probassign'} (\algotot\given (\algo_1, \algo_2)) - \regret^{\probassign}(\algotot\given (\algo_1, \algo_2)\big).\label{eq:detection_rule_bandit}
\end{align}
In what follows, we build upon our results from \Cref{sec:regimes}, and use the \egreedy and UCB algorithms as examples to show that (1) a sign violation caused by running the greedy algorithm jointly with the \egreedy or UCB algorithm can be correctly detected by~\eqref{eq:detection_rule_bandit}; (2) under a sign preservation by running a pair of \egreedy or UCB algorithms, no warning is issued according to~\eqref{eq:detection_rule_bandit}.

First, we consider the regime of sign violation. According to~\eqref{eq:detection_rule_bandit}, any non-monotonic behavior in the total regret as the treatment probability increases triggers a warning. We now show that when the greedy algorithm runs jointly with the \egreedy or UCB algorithm, this warning is indeed triggered. Suppose Algorithm~1 is either the \egreedy or UCB algorithm, while Algorithm~2 is the greedy algorithm. The results in Theorems~\ref{thm:greedy+e-greedy} and~\ref{thm:greedy+UCB} can be readily extended to a Bernoulli assignment with probability $0.5$ to show that Algorithm 2 performs better under data sharing. Now think about a ramp-up procedure with $\probassign$ increasing from $0$ to $1$. When $\probassign = 0$, the system only runs Algorithm 1, incurring a low regret. When $\probassign = 1$, the system reduces purely to the greedy algorithm, which incurs a linear regret. Therefore, the total regret must increase at some point. This corresponds precisely to the situation in which increasing the treatment probability of the seemingly better performing algorithm leads to an increase in the total regret, thereby triggering a warning for a potential sign violation.

Now we move on to the regime of sign preservation. We show that when running a pair of \egreedy or UCB algorithms with parameters $0 \le \alpha_1 < \alpha_2 \le 1$, the regret is monotonically increasing in the probability of the worse algorithm.
Specifically, we consider treatment probabilities of the form
\[
\probassign = T^{\beta - 1},
\]
for varying values of $\beta\in (0, 1)$. The following two theorems provide the regret rates in terms of $\beta$ for the \egreedy and UCB algorithms, respectively.
We start with the following results for \egreedy algorithms.

\begin{theorem} \label{thm:e-greedy_ramp_up}
Consider running two \egreedy algorithms, $\algoegreedy_{\alpha_1, C}$ and $\algoegreedy_{\alpha_2, C}$ with $0 \le \alpha_1 < \alpha_2 \le 1$ and 
$C \ge \max\{120K,\; 32K / \gapmin^2\}$.
Suppose the traffic allocation percentage to 
$\algoegreedy_{\alpha_2, C}$ is 
$p = T^{\beta-1}$ for some $\beta \in (0,1)$.
Then the expected total regret under data sharing satisfies  
\begin{align}\label{eq:regret_rampup_pair_egreedy}
      \EE{R^\probassign(\algotot \mid(\algoegreedy_{\alpha_1, C}, \algoegreedy_{\alpha_2, C})} 
 = \begin{cases}
        \Theta( \log T) & \text{if } \alpha_1 = 0 \text{ and } \alpha_2+\beta \le 1,\\
        \Theta(T^{\max\cb{\alpha_1,  \alpha_2 + \beta-1}}) & \text{otherwise.}
\end{cases}
\end{align}
\end{theorem}
The proof of this theorem is provided in \Cref{sec:proof_thm_egreedy_ramp_up}.  Now we provide the results for UCB algorithms.

\usetikzlibrary{arrows.meta}
\begin{figure}[t]
\subfloat[\egreedy]{
\centering
\begin{tikzpicture}[scale=0.65, x=8cm,y=5cm,>=Stealth,thick]

\draw[->] (-0.08,0) -- (1.08,0) node[below right] {$\beta$};
\draw[->] (0,0) -- (0,1.08) node[above] {Regret};

\def\xk{0.6} 
\def\yhi{0.72}
\def\ylo{0.32}

\draw (0,0) -- (0,-0.02) node[below] {$0$};
\draw (\xk,0) -- (\xk,-0.02) node[below] {$1 - \alpha_2 + \alpha_1$};
\draw (1,0) -- (1,-0.02) node[below] {$1$};

\draw (0,\yhi) -- (-0.02,\yhi) node[left] {$T^{\alpha_2}$};
\draw (0,\ylo) -- (-0.02,\ylo) node[left] {$T^{\alpha_1}$};

\draw[dashed, line width=1pt] (0,\yhi) -- (1,\yhi);

\draw[line width=1.2pt] (0.0,\ylo) -- (\xk,\ylo);

\def\kappa{3.0}

\draw[line width=1.2pt]
  plot[domain=\xk:1, samples=120, variable=\x]
  ({\x},
   {\ylo + (\yhi-\ylo) * (exp(\kappa*(\x-\xk)) - 1) / (exp(\kappa*(1-\xk)) - 1)}
  );

\node[below,align=center] at (0,-0.13) {More traffic to\\ $\algoegreedy_{\alpha_1, C}$};
\node[below,align=center] at (1,-0.13) {More traffic to\\ $\algoegreedy_{\alpha_2, C}$};

\end{tikzpicture}
}
\subfloat[UCB]{
\centering
\begin{tikzpicture}[scale=0.65, x=8cm,y=5cm,>=Stealth,thick]

\def\xa{0.3}   
\def\xb{0.7}   
\def\ylo{0.32}  
\def\yhi{0.72}  

\def\kappa{3.0}

\draw[->] (-0.08,0) -- (1.08,0) node[below right] {$\beta$};
\draw[->] (0,0) -- (0,1.08) node[above] {Regret};

\draw (0,0) -- (0,-0.02) node[below] {$0$};
\draw (\xa,0) -- (\xa,-0.02) node[below] {$\alpha_1$};
\draw (\xb,0) -- (\xb,-0.02) node[below] {$\alpha_2$};
\draw (1,0) -- (1,-0.02) node[below] {$1$};

\draw (0,\ylo) -- (-0.02,\ylo) node[left] {$T^{\alpha_1}$};
\draw (0,\yhi) -- (-0.02,\yhi) node[left] {$T^{\alpha_2}$};

\draw[dashed] (0,\yhi) -- (1,\yhi);


\draw[line width=1.2pt] (0,\ylo) -- (\xa,\ylo);

\draw[line width=1.2pt]
  plot[domain=\xa:\xb, samples=140, variable=\x]
  ({\x},
   {\ylo + (\yhi-\ylo)
    * (exp(\kappa*(\x-\xa)) - 1)
    / (exp(\kappa*(\xb-\xa)) - 1)}
  );

\draw[line width=1.2pt] (\xb,\yhi) -- (1,\yhi);

\node[below,align=center] at (0,-0.13) {More traffic to\\ $\algoucb_{\alpha_1}$};
\node[below,align=center] at (1,-0.13) {More traffic to\\ $\algoucb_{\alpha_2}$};

\end{tikzpicture}
}
\caption{Total regret as a function of the traffic allocation parameter $\beta$.\label{fig:ramp_up_rates}}
\end{figure}

\begin{theorem} \label{thm:ucb_ramp_up}
Consider running two UCB algorithms, $\algoucb_{\alpha_1}$ and $\algoucb_{\alpha_2}$, with parameters $0 \le \alpha_1 < \alpha_2 \le 1$.
Suppose the traffic allocation percentage to 
$\algoucb_{\alpha_2}$ is 
$p = T^{\beta - 1}$ for some $\beta \in (0,1)$. Then the expected total regret under data sharing satisfies 
\begin{align}\label{eq:regret_rampup_pair_ucb}
      \EE{R^\probassign(\algotot \mid(\algoucb_{\alpha_1}, \algoucb_{\alpha_2}, p)} 
 = \Theta(T^{\max\cb{\alpha_1,  \min\{\alpha_2, \beta\}}}).
\end{align}
\end{theorem}
The proof of this theorem is provided in \Cref{sec:proof_thm_ucb_ramp_up}.
Taken together, these two theorems show that the total regret is monotonically increasing as the probability of the worse algorithm increases, as illustrated in Figure~\ref{fig:ramp_up_rates}. Specifically, the algorithm with exploration parameter $\paramexplore_1$ outperforms the algorithm with parameter $\paramexplore_2$. With increasing treatment probability  (i.e., increasing $\paramrampup$) for the worse algorithm (i.e., $\algoegreedy_{\paramexplore_2, C}$ or $\algoucb_{\paramexplore_2, C}$), the total regret increases.

As an aside, while the \egreedy and UCB algorithms behave similarly in our previous results under a fixed 50-50 split, their regret rates become different in a ramp-up experiment as seen from~\eqref{eq:regret_rampup_pair_egreedy} and~\eqref{eq:regret_rampup_pair_ucb}, due to the following reason. The exploration level of the \egreedy algorithm is determined by its parameter $\paramexplore$, whereas the exploration level of the UCB algorithm can also be affected by data sharing: If the other algorithm exploits and pulls the optimal arm excessively, the confidence interval of the optimal arm becomes very small, pushing the UCB algorithm to explore even more. Thus, the pair of UCB algorithms reach a high regret of order $\horizon^{\paramexplore_2}$ faster, before $\algoucb_{\paramexplore_2}$ is allocated the full traffic.

\section{Simulation}\label{sec:sim}

We now present simulation studies that validate our theoretical results and explore additional settings, including using the probability of correct comparison as the performance metric and examining the impact of a misspecified prior.
We consider $\numarms = 2$ arms for simplicity, where rewards are drawn from Bernoulli distributions with means $\meantrue_1= 0.8$ and $\meantrue_2=0.2$. The horizon of the problem is $\horizon = 100$ unless otherwise specified. For the \egreedy algorithm $\algoegreedy_{\paramexplore, C}$, we set its exploration constant to be $C = 1$ for simplicity. In all simulations, each data point is based on 10,000 runs. Error bars (barely visible) indicate the standard error of the mean. 
All code for simulation is provided at \url{https://github.com/jingyanw/bandit-data-sharing}.

\subsection{Sign violation and preservation}
\begin{figure}[tb]
    \centering
    \subfloat[Running greedy with \egreedy]{
        \includegraphics[width=0.475\linewidth]{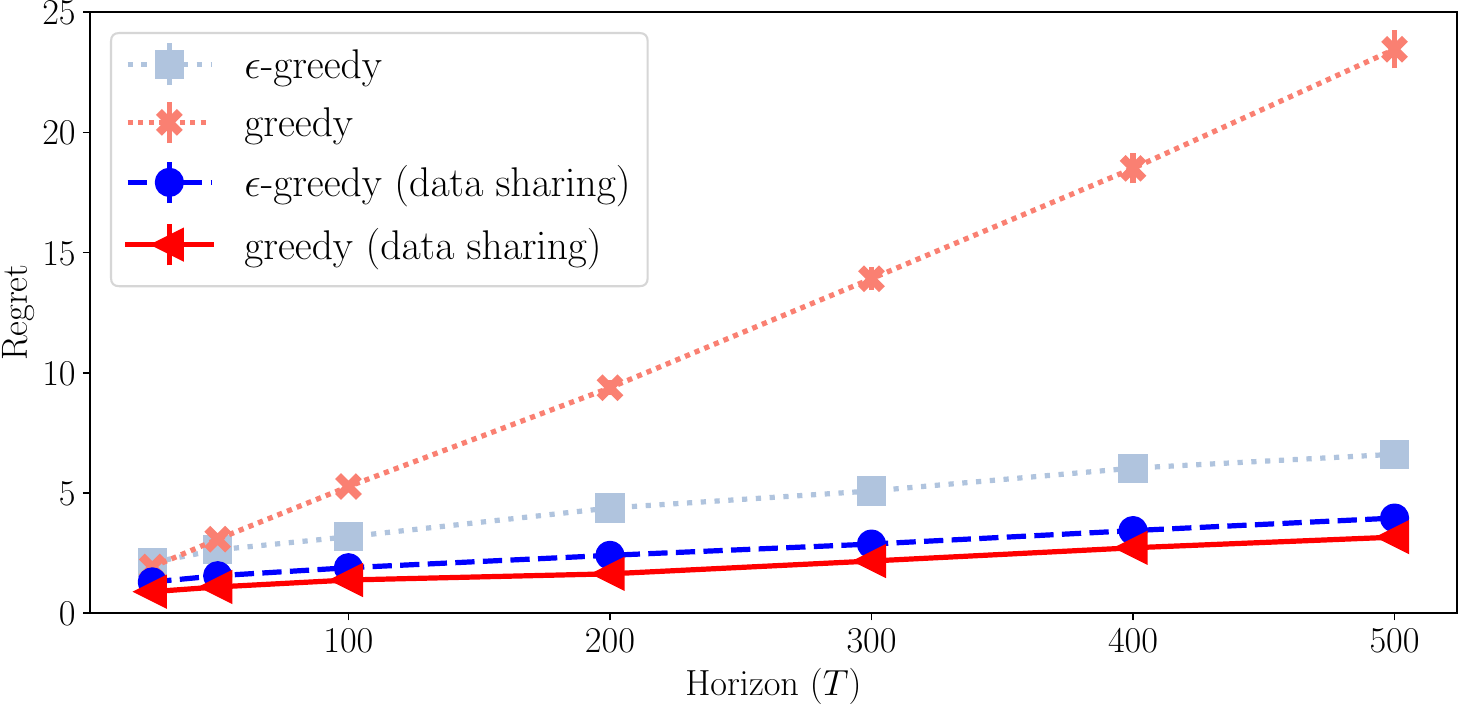}\label{float:sign_violation_pair_egreedy}
    }
    \hspace{1mm}
    \subfloat[Running greedy with UCB]{
        \includegraphics[width=0.475\linewidth]{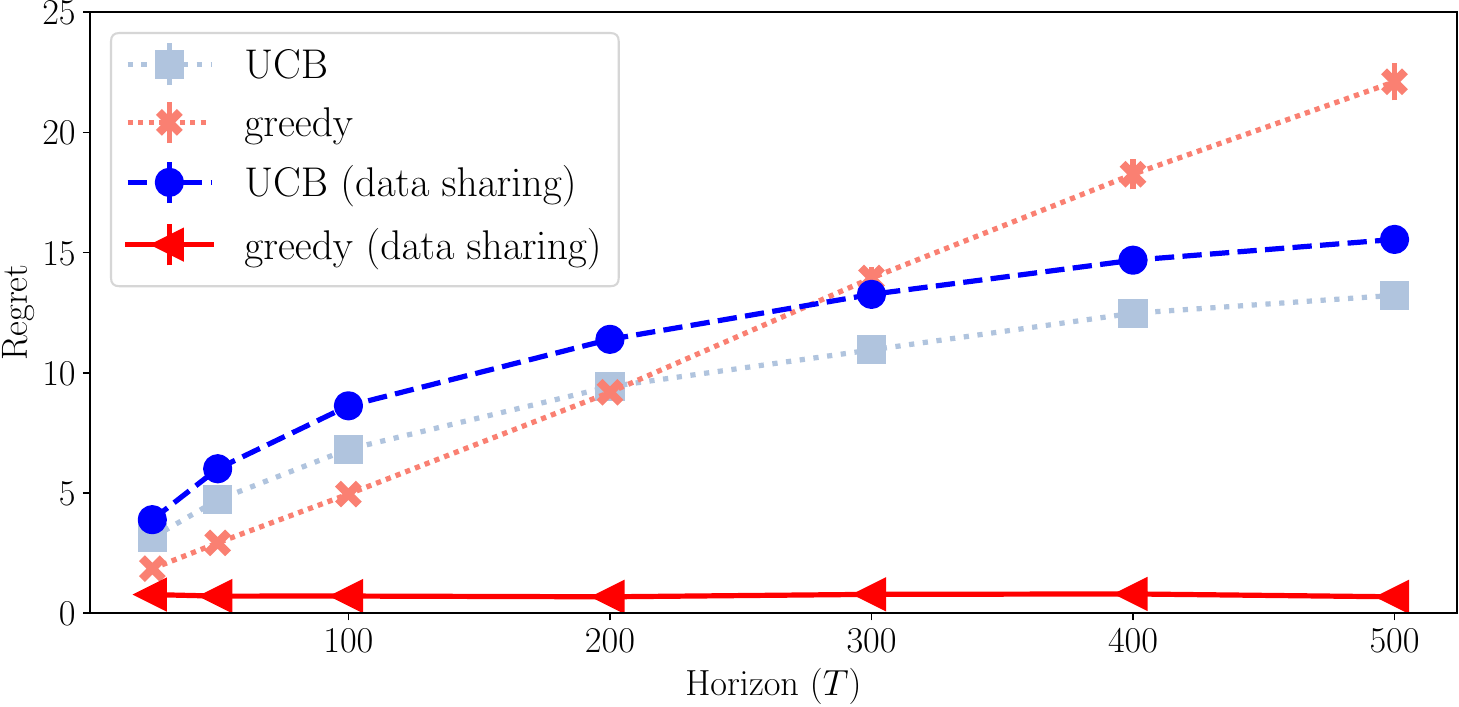}\label{float:sign_violation_pair_ucb}
    }
    \caption{Sign violation. The greedy algorithm runs jointly with the \egreedy or UCB algorithm.\label{fig:sign_violation_pair}}
\end{figure}

\paragraph{Sign violation.} Following the setting considered in Section~\ref{sec:results_sign_violation}, we run the greedy algorithm jointly with the \egreedy algorithm (with $\paramexplore=0$) or the UCB algorithm (with $\paramexplore=0$) under \datafeedback, shown in Figure~\ref{fig:sign_violation_pair}. The results are consistent with Theorems~\ref{thm:greedy+e-greedy} and~\ref{thm:greedy+UCB}. In both cases, while the greedy algorithm incurs a linear regret when running individually, its regret decreases significantly and outperforms the other algorithm when running jointly. We observe that the regret of the \egreedy and UCB algorithms also decrease notably under \datafeedback, suggesting that they also benefit from the shared data, though not as much as the greedy algorithm does. In addition, when we increase the exploration level by the \egreedy algorithm by increasing the value of $C$ to be greater than $1$, the regret of the greedy algorithm further decreases (not shown).

\begin{figure}
\vspace{5mm}
\centering
\subfloat[\egreedy]{
    \includegraphics[height=\figureheight, trim={0 0 0 0},clip]{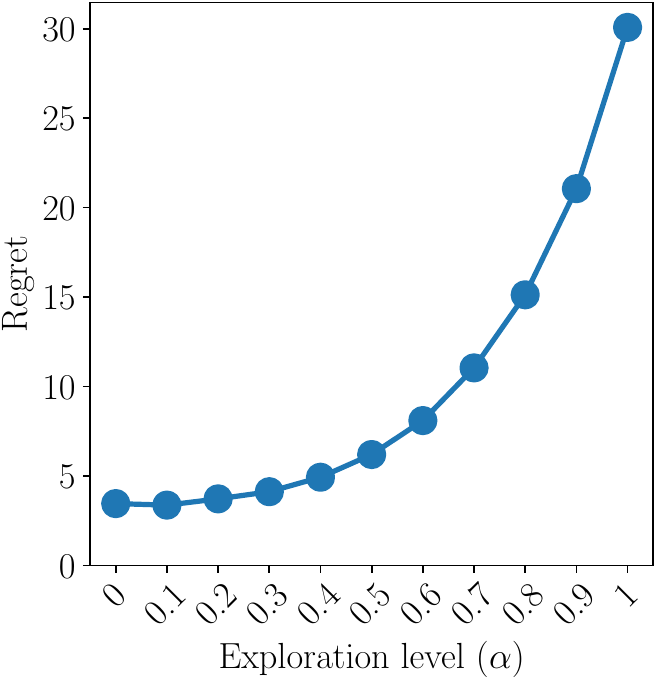}
    \includegraphics[height=\figureheight]{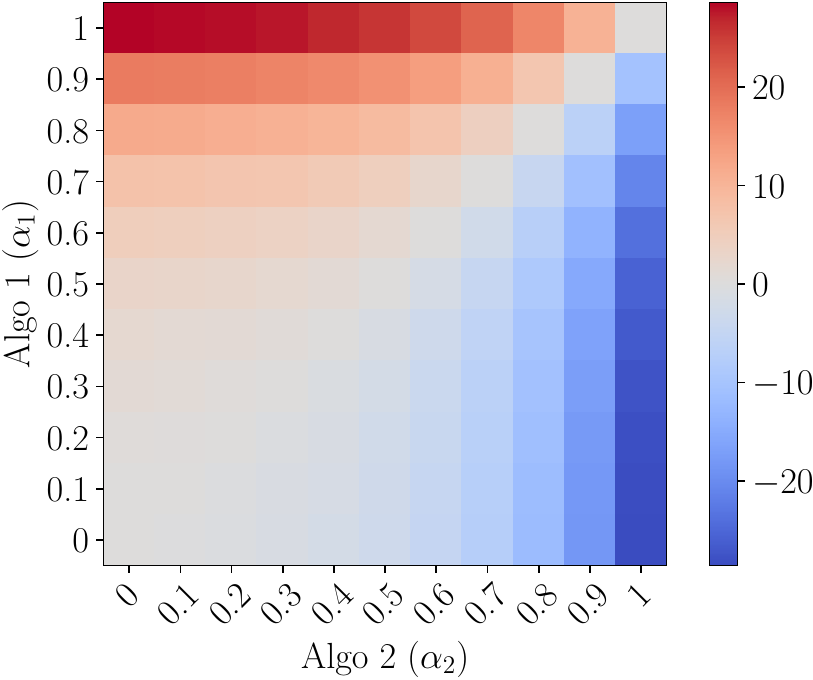}
    \begin{picture}(0,0)
        \put(-199,122){\makebox(0,0)[lt]{\fontsize{8}{10}\selectfont Individual run}}
        \put(-116,122){\makebox(0,0)[lt]{\fontsize{8}{10}\selectfont Joint run with data sharing}}
        \put(-180,110){\makebox(0,0)[lt]{\fontsize{8}{10}\selectfont $\Expect[R]$}}
        \put(-104,110){\makebox(0,0)[lt]{\fontsize{8}{10}\selectfont $\Expect[R(\mathcal{A}_1)] - \Expect[R(\mathcal{A}_2)]$}}
    \end{picture}
}
\subfloat[UCB]{
    \includegraphics[height=\figureheight, trim={0 0 0 0},clip]{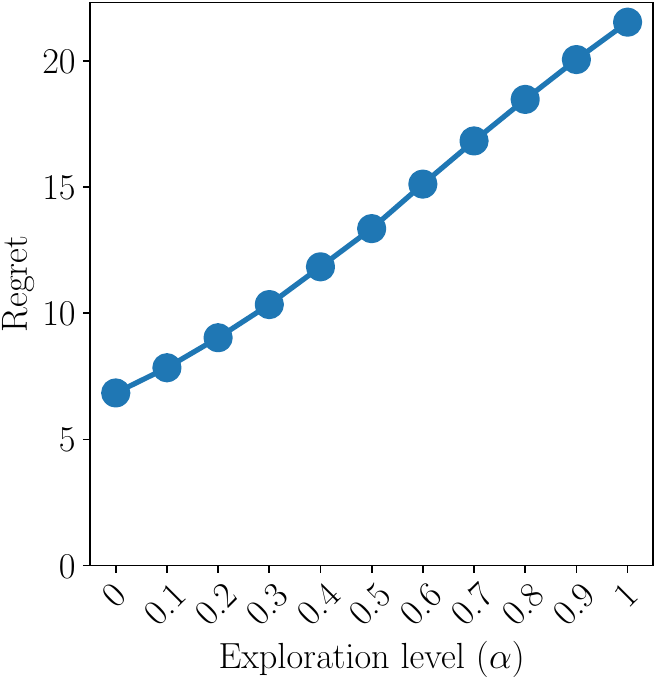}
    \includegraphics[height=\figureheight]{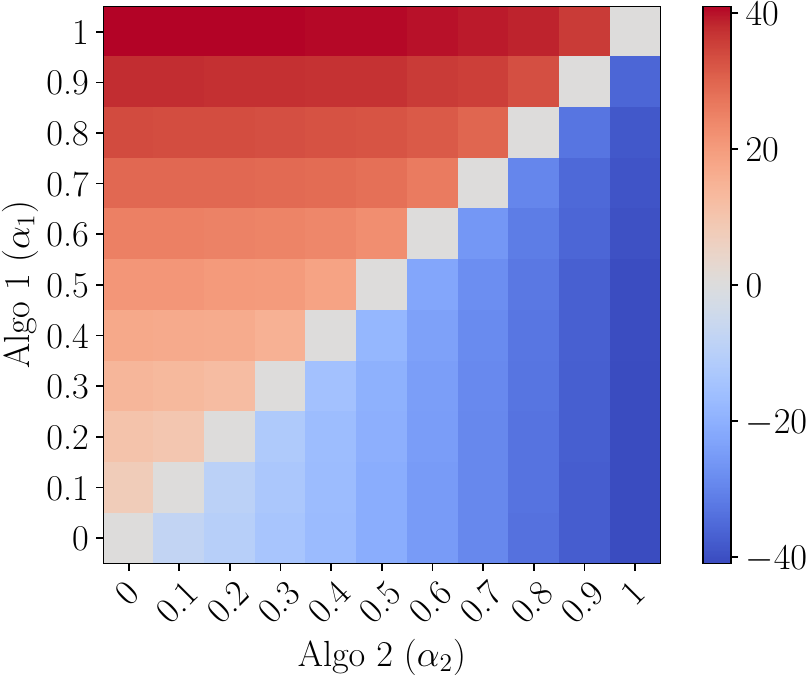}
    \begin{picture}(0,0)
        \put(-199,122){\makebox(0,0)[lt]{\fontsize{8}{10}\selectfont Individual run}}
        \put(-116,122){\makebox(0,0)[lt]{\fontsize{8}{10}\selectfont Joint run with data sharing}}
        \put(-180,110){\makebox(0,0)[lt]{\fontsize{8}{10}\selectfont $\Expect[R]$}}
        \put(-104,110){\makebox(0,0)[lt]{\fontsize{8}{10}\selectfont $\Expect[R(\mathcal{A}_1)] - \Expect[R(\mathcal{A}_2)]$}}
    \end{picture}
}
    \caption{Sign preservation. Red color means that algorithm 1 is worse than algorithm 2; blue color means that algorithm 1 is better than algorithm 2. (Best viewed in color.)}
    \label{fig:sign_preservation}
\end{figure}

\paragraph{Sign preservation.}
Figure~\ref{fig:sign_preservation} considers the classes of \egreedy or UCB algorithms. The individual run shows the regret of a single algorithm as a function of the exploration parameter $\paramexplore$, and the joint run shows the difference between the regrets of two algorithms within the same class under \datafeedback. When an algorithm runs individually, as expected from Proposition~\ref{prop:run_individually}, the regret increases in $\paramexplore$ as the algorithm starts to over-explore. Then within each class, we run a pair of algorithms with parameters $\paramexplore_1$ and $\paramexplore_2$ jointly. When algorithm 1 and algorithm $2$ run jointly, we measure the difference between the regrets of the two algorithms $\Expect[\regret(\algo_{\paramexplore_1} \given (\algo_{\paramexplore_1},\algo_{\paramexplore_2})]- \Expect[\regret(\algo_{\paramexplore_2} \given (\algo_{\paramexplore_1},\algo_{\paramexplore_2})]$. We observe that algorithm 1 incurs a higher regret (red color) when $\paramexplore_1 > \paramexplore_2$ (left upper triangle), and a lower regret (blue color) when $\paramexplore_1 < \paramexplore_2$ (right bottom triangle). Hence, the sign of the comparison between the two algorithms is preserved under \datafeedback.

\subsection{Probability of correct comparison}\label{sec:sim_prob}

\begin{figure}[tb]
    \vspace{15mm}
    \centering
    \subfloat[Comparing two \egreedy algorithms]{
        \includegraphics[height=0.2\linewidth, trim={0 0 50 0},clip]{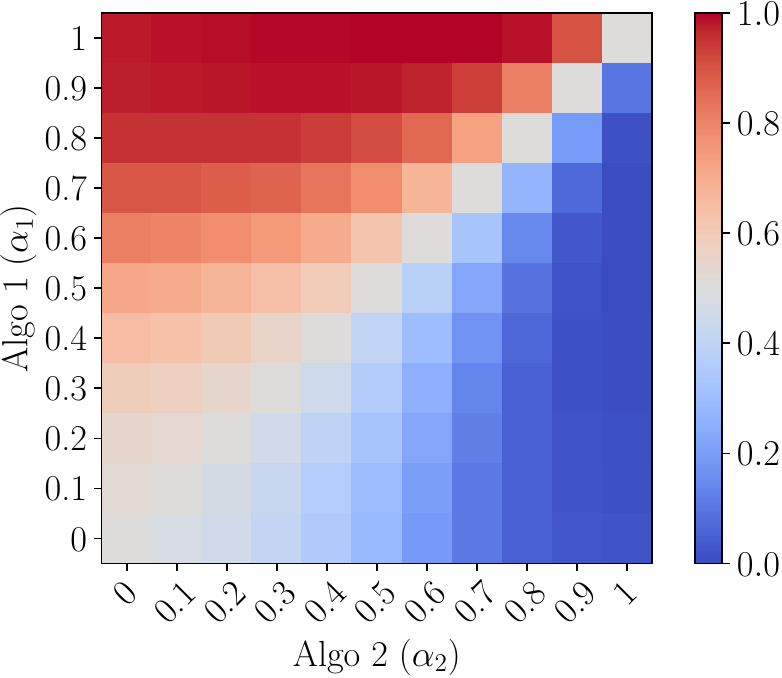}
        \includegraphics[height=0.2\linewidth]{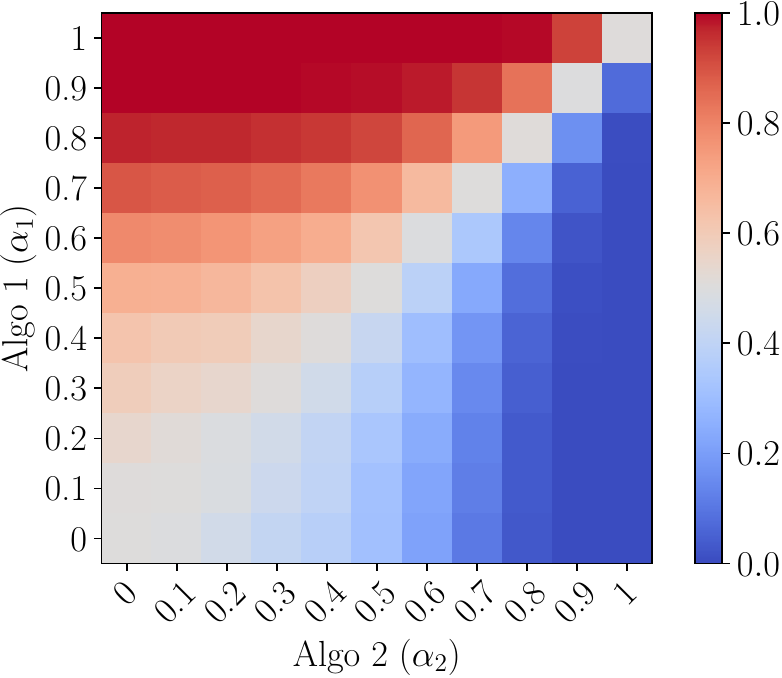}
        \begin{picture}(0,0)
            \put(-188,122){\makebox(0,0)[lt]{\fontsize{8}{10}\selectfont Individual run}}
            \put(-112,122){\makebox(0,0)[lt]{\fontsize{8}{10}\selectfont Joint run with data sharing}}
            \put(-205,110){\makebox(0,0)[lt]{\fontsize{8}{10}\selectfont $\Prob(\regret(\algo_1) > \regret(\algo_2))$}}
            \put(-125,110){\makebox(0,0)[lt]{\fontsize{8}{10}\selectfont $\Prob\left(\regret(\algo_1 \given (\algo_1, \algo_2)) > \regret(\algo_2 \given (\algo_1, \algo_2))\right)$}}
        \end{picture}
    }
    \hspace{5mm}
    \subfloat[Comparing two UCB algorithms]{\label{float:stochastic_ucb}
        \includegraphics[height=0.2\linewidth, trim={0 0 50 0},clip]{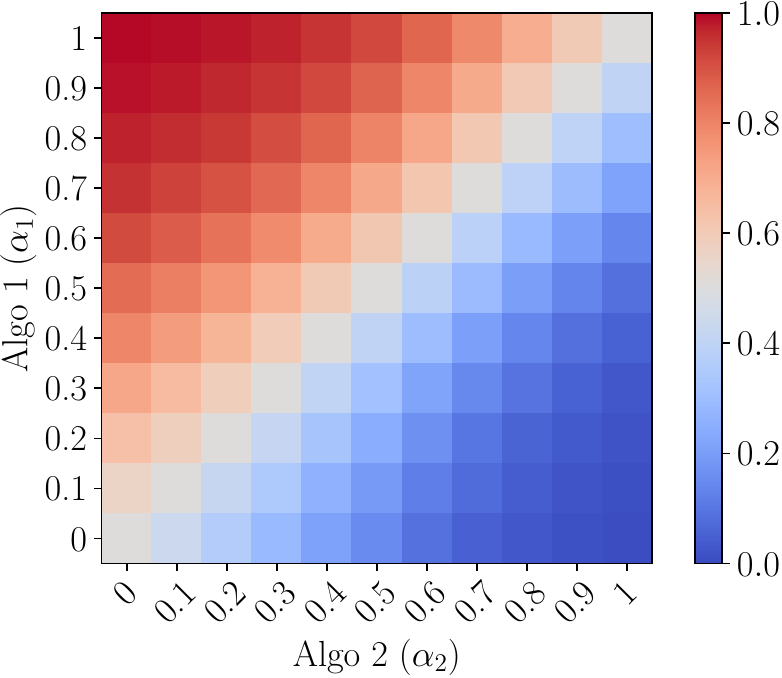}
        \includegraphics[height=\figureheight]{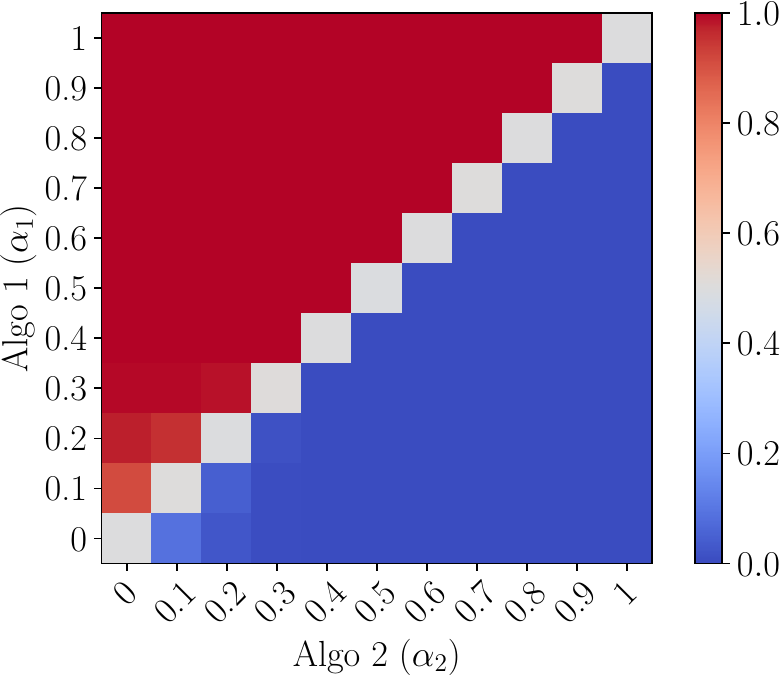}
        \begin{picture}(0,0)
            \put(-195,122){\makebox(0,0)[lt]{\fontsize{8}{10}\selectfont Individual run}}
            \put(-114,122){\makebox(0,0)[lt]{\fontsize{8}{10}\selectfont Joint run with data sharing}}
            \put(-210,110){\makebox(0,0)[lt]{\fontsize{8}{10}\selectfont $\Prob(\regret(\algo_1) > \regret(\algo_2))$}}
            \put(-130,110){\makebox(0,0)[lt]{\fontsize{8}{10}\selectfont $\Prob\left(\regret(\algo_1 \given (\algo_1, \algo_2)) > \regret(\algo_2 \given (\algo_1, \algo_2))\right)$}}
        \end{picture}
    }
    \caption{Comparison of stochastic regret. Red color means that algorithm 1 performs worse than algorithm 2 with probability greater than $0.5$; blue color means algorithm 1 performs better than algorithm 2 with probability greater than $0.5$. (Best viewed in color.)}
    \label{fig:probability}
\end{figure}

In practice, the comparison between two algorithms is made by running an A/B experiment once or a few times. So far we have examined the \emph{expected} regret $\Expect[\regret(\algo_1 \given (\algo_1, \algo_2))] - \Expect[\regret(\algo_2 \given (\algo_1, \algo_2))]$. Now we also examine the probability of correctness in comparing the \emph{stochastic} regrets, that is, $\PP{\regret(\algo_1 \given (\algo_1, \algo_2)) > \regret(\algo_2 \given (\algo_1, \algo_2))}$, where the regrets of the two algorithms are computed from the same run of Algorithm~\ref{algo:joint}. In Figure~\ref{fig:probability}, we compute the probability $\Prob(\regret(\algo_1) > \regret(\algo_2))$ when running each algorithm individually and the probability $\Prob(\regret(\algo_1 \given (\algo_1, \algo_2)) > \regret(\algo_2 \given (\algo_1, \algo_2)))$ when running them jointly. For both cases, we observe that the probability of the better algorithm incurring a lower regret is greater than $0.5$, as indicated by upper left triangle being red and the bottom right triangle being blue. Interestingly, we empirically observe that the comparison has a higher accuracy when two UCB algorithms run jointly than individually. We posit that the higher accuracy comes from sharing the same randomness in reward samples for the two algorithms. Algorithms in general tend to achieve a lower regret on ``easy'' instances where the reward samples of the best arm are high (so that the best arm is easily identified), and higher regret on ``hard'' instances where the reward samples of the best arm are low. When both algorithms run individually, a comparison may be incorrect if the worse algorithm incurs a low regret on an ``easy'' instance, and the better algorithm incurs a high regret on a ``hard'' instance. However, when running two algorithms jointly on the same instance, their relative ordering of performance is retained.

\subsection{Misspecified prior in \ts}\label{sec:sim_ts}
We have so far characterized the performance of algorithms through the level of exploration versus exploitation. Are there other dimensions that capture an algorithm's performance? We now examine algorithms' performance in terms of the accuracy of their priors.

We consider \ts with a prior size of $\priorsize=5$. \ts algorithm is described in Algorithm~\ref{algo:ts} in Appendix~\ref{appendix:TS}. For a parameter $\paramprior\in [0, 1]$, we set the prior of arm $1$ (better arm) as $\Beta(\priorsize (1-\paramprior), \priorsize \paramprior)$ and the prior of arm $2$ (worse arm) as $\Beta(\priorsize \paramprior, \priorsize(1-\paramprior))$. Informally, a prior of $\Beta(\priorsize \paramprior, \priorsize(1-\paramprior))$ can be thought of as having $\priorsize$ Bernoulli observations with $\paramprior$ fraction of value $1$ and $(1-\paramprior)$ fraction of value $0$. Hence, a small value of $\paramprior$ means that the algorithm has a stronger belief that arm $1$ is indeed the better arm. Each algorithm holds its own prior and the prior is not shared across the two algorithms.

\begin{figure}[tb]
\vspace{5mm}
\centering
\subfloat[Expected regret]{
    \includegraphics[height=0.21\linewidth, trim={0 0 0 0},clip]{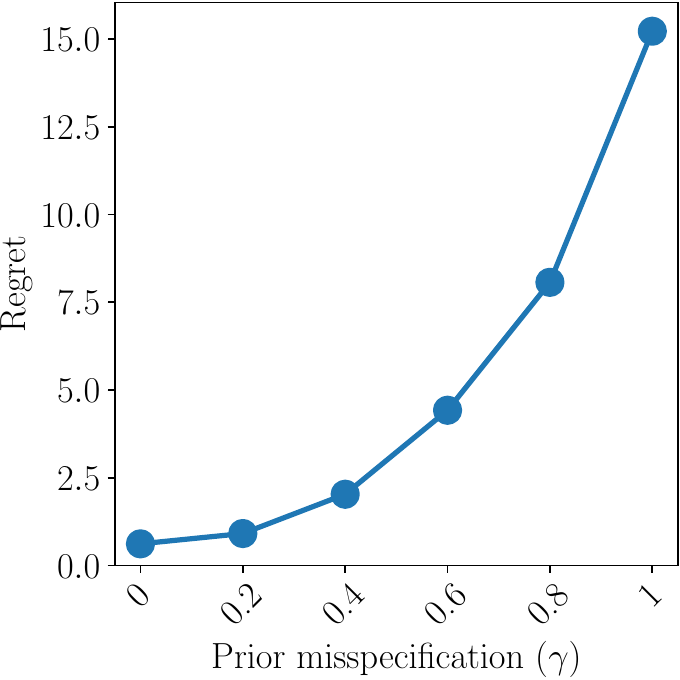}
    \includegraphics[height=0.21\linewidth]{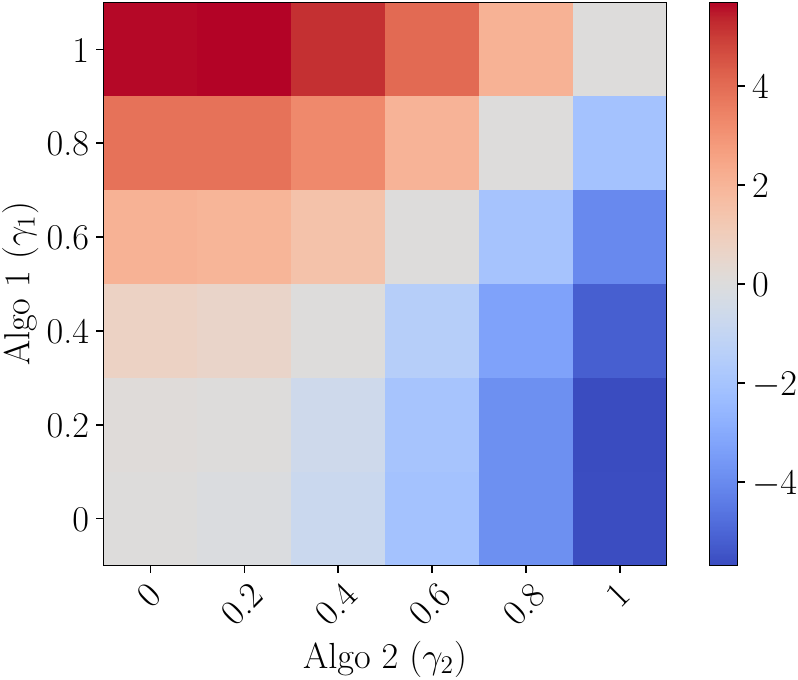}
    \begin{picture}(0,0)
        \put(-195,122){\makebox(0,0)[lt]{\fontsize{8}{10}\selectfont Individual run}}
        \put(-114,122){\makebox(0,0)[lt]{\fontsize{8}{10}\selectfont Joint run with data sharing}}
        \put(-175,110){\makebox(0,0)[lt]{\fontsize{8}{10}\selectfont $\Expect[R]$}}
        \put(-102,110){\makebox(0,0)[lt]{\fontsize{8}{10}\selectfont $\Expect[R(\mathcal{A}_1)] - \Expect[R(\mathcal{A}_2)]$}}
    \end{picture}
}
\subfloat[Probability of comparison]{
    \includegraphics[height=0.21\linewidth, trim={0 0 50 0},clip]{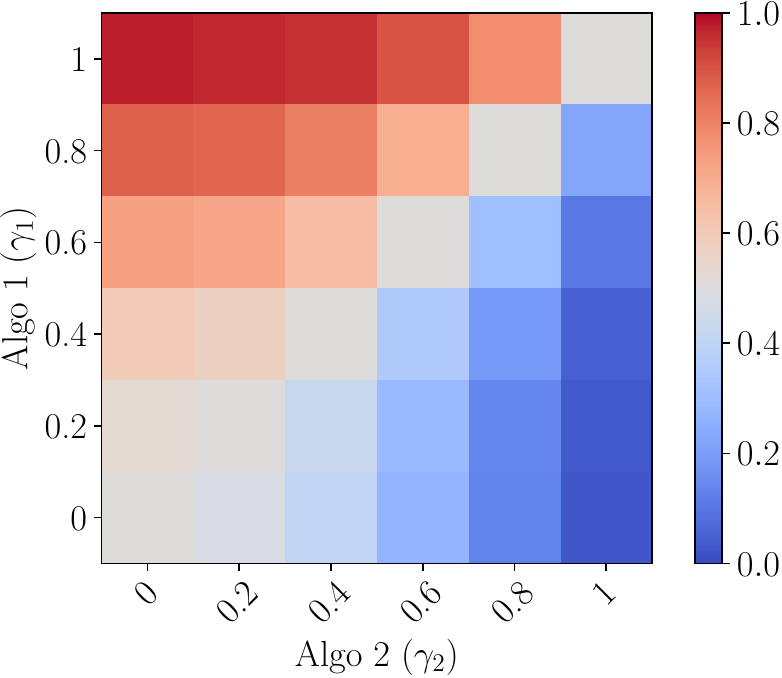}
    \includegraphics[height=0.21\linewidth]{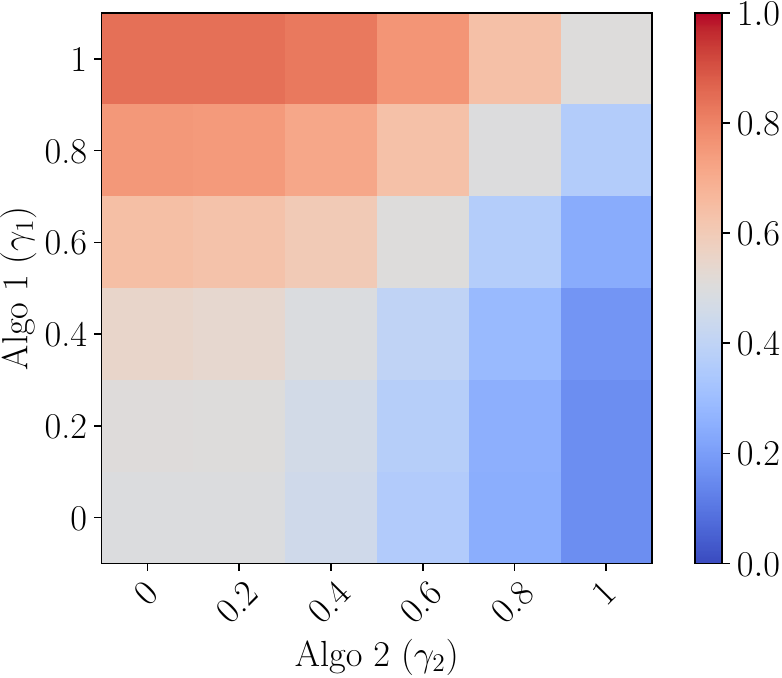}
    \begin{picture}(0,0)
        \put(-195,122){\makebox(0,0)[lt]{\fontsize{8}{10}\selectfont Individual run }}
        \put(-114,122){\makebox(0,0)[lt]{\fontsize{8}{10}\selectfont Joint run with data sharing}}
        \put(-205,110){\makebox(0,0)[lt]{\fontsize{8}{10}\selectfont $\Prob(\regret(\algo_1) > \regret(\algo_2))$}}
        \put(-125,110){\makebox(0,0)[lt]{\fontsize{8}{10}\selectfont $\Prob(\regret(\algo_1\given (\algo_1, \algo_2)) > \regret(\algo_2 \given (\algo_1, \algo_2)))$}}
    \end{picture}
}

    \caption{\ts. Red color means that algorithm 1 is worse than algorithm 2; blue color means algorithm 1 is better than algorithm 2. (Best viewed in color.)
    \label{fig:ts}}
    
\end{figure}

Figure~\ref{fig:ts} shows the regret of a single algorithm in an individual run, and the difference between the regrets of two algorithms in a joint run, in terms of both the expected regret and the probability of correct comparison. We again observe that the better algorithm when running individually remains better when running jointly in terms of both metrics; in other words, the comparison is preserved under data sharing. Interestingly, contrary to Figure~\ref{fig:probability}\subref{float:stochastic_ucb} from the previous section, we empirically observe that the comparison has a lower accuracy when the two algorithms run jointly than individually. We posit that the shared data helps the algorithm with the worse prior correct its misspecification more quickly, and hence decreases the performance gap between the two algorithms when running jointly.

\subsection{Detecting sign violation with ramp-up experiments}
We now simulate ramp-up experiments for the regimes of sign preservation and sign violation. We vary the treatment probability $\probassign\in [0, 1]$ following a Bernoulli assignment in Algorithm~\ref{algo:joint_bernoulli}. We take examples from the sign violation regime (running the greedy algorithm and the UCB algorithm jointly) and the sign preservation regime (running a pair of UCB algorithms), and examine whether our detection procedure from \Cref{sec:guidelines} behaves correctly.
\begin{figure}
    \centering
\subfloat[Greedy vs. $\algoucb_0$]{
        \begin{tabular}{c}
            \includegraphics[width=0.45\textwidth]{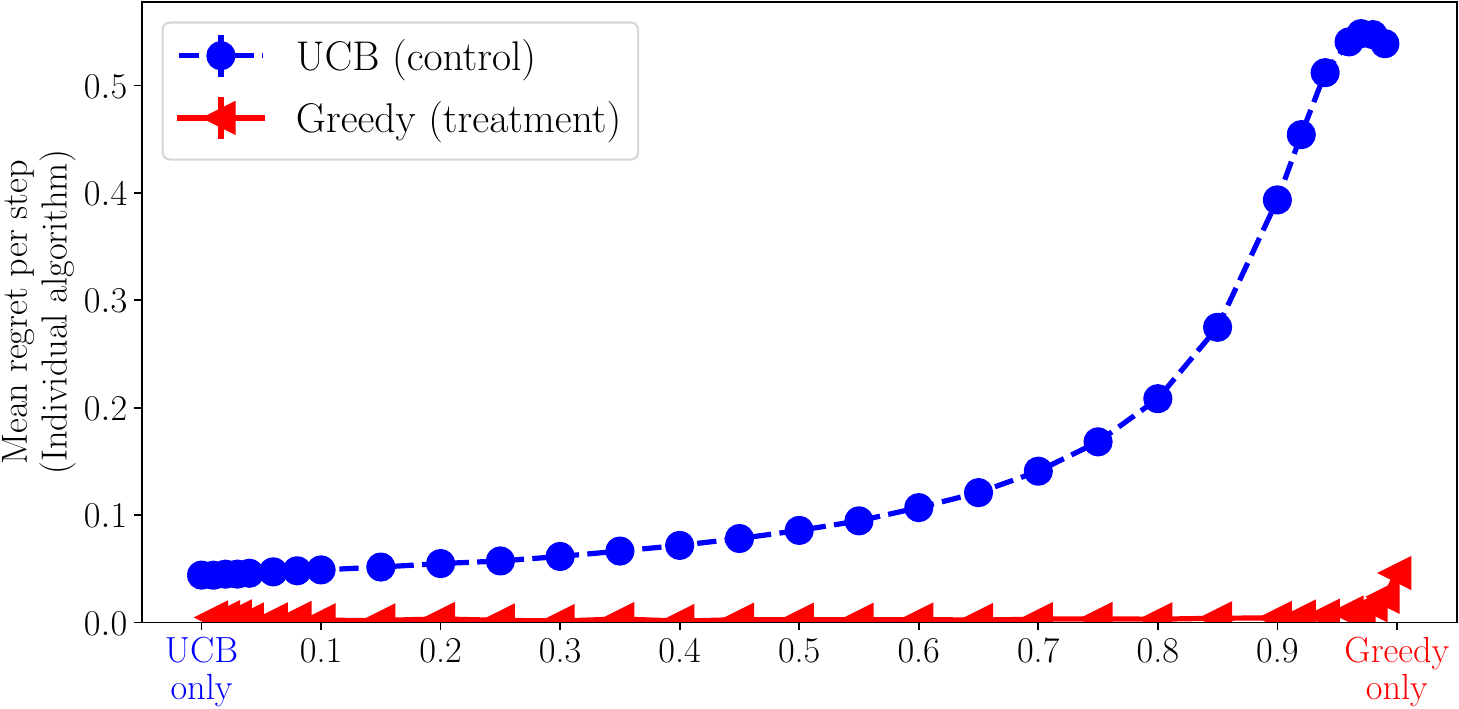}\\
            \includegraphics[width=0.45\textwidth]{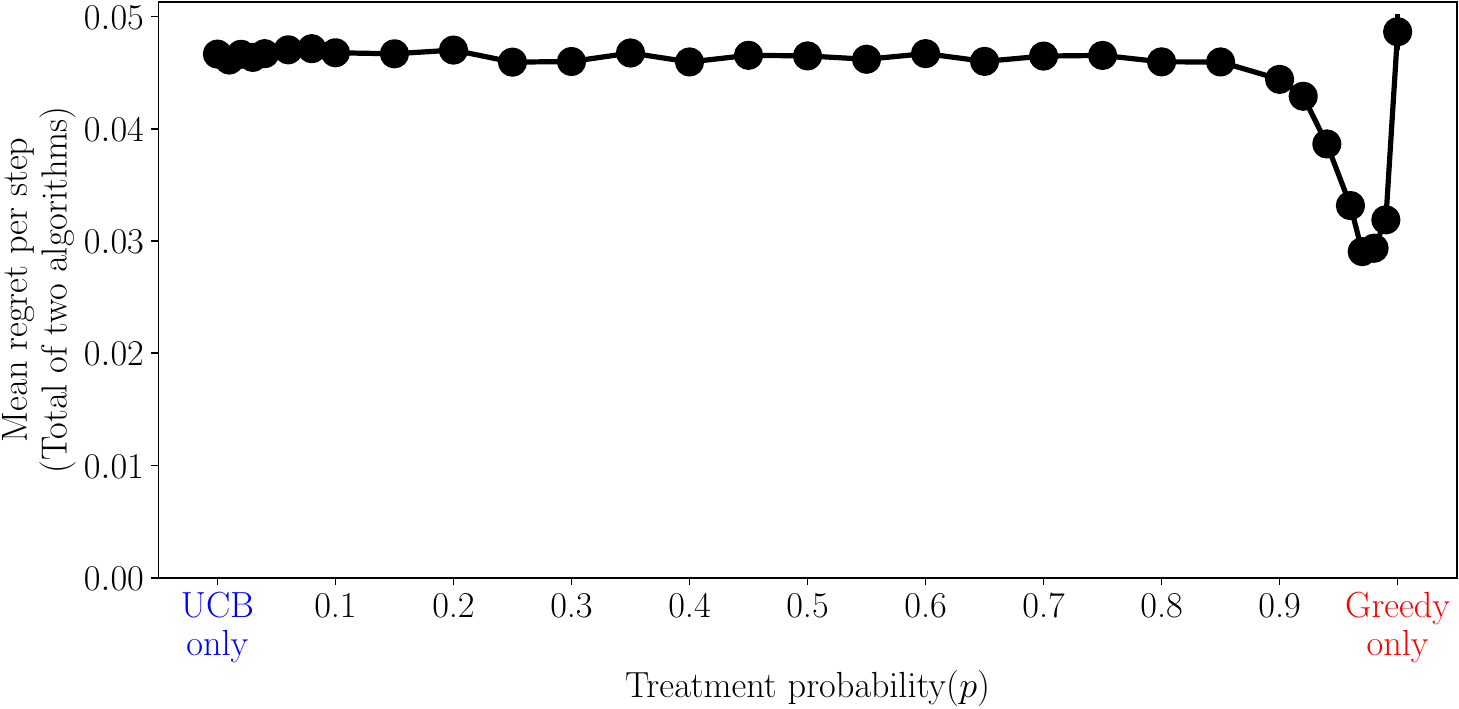}
        \end{tabular}
        \label{fig:rampup_greedy_ucb}
    }
    \subfloat[$\algoucb_0$ vs.$\algoucb_{0.5}$]{
        \begin{tabular}{c}
            \includegraphics[width=0.45\textwidth]{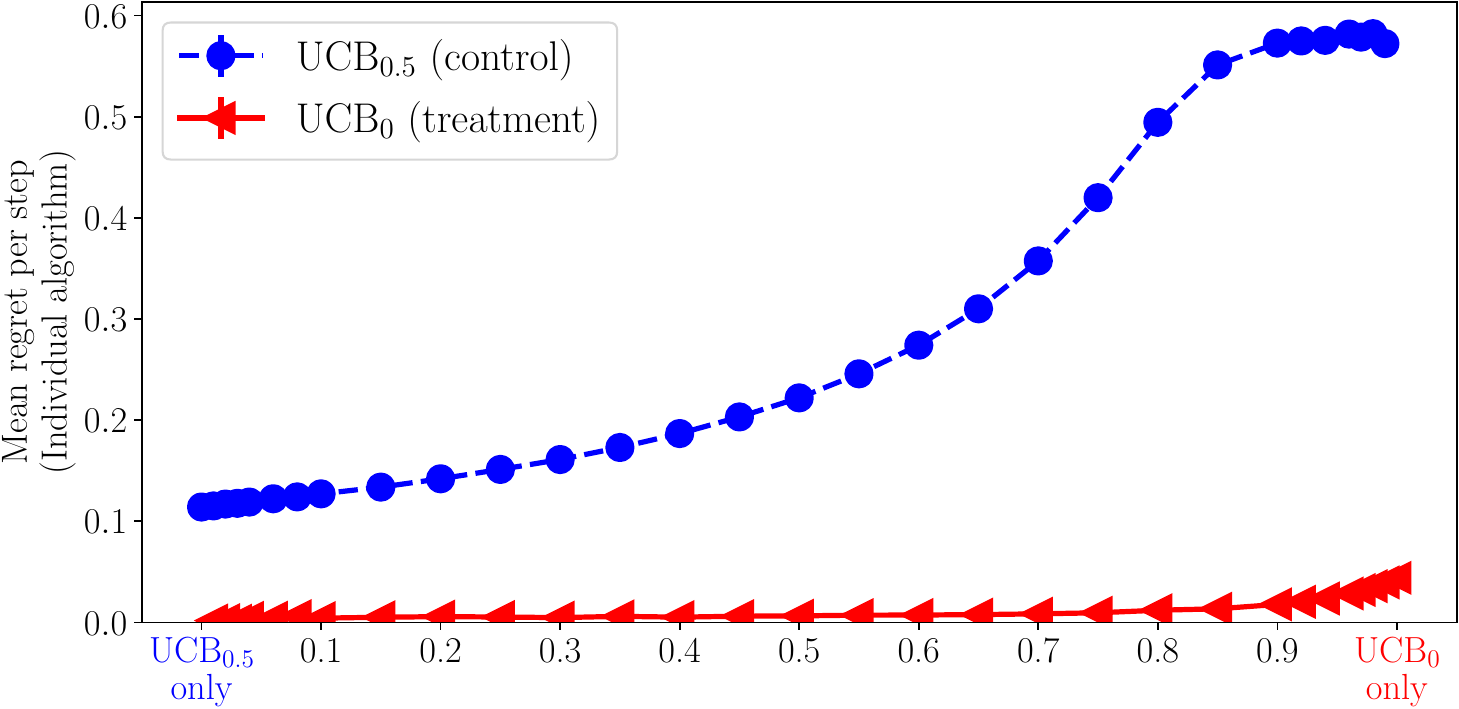}\\
            \includegraphics[width=0.45\textwidth]{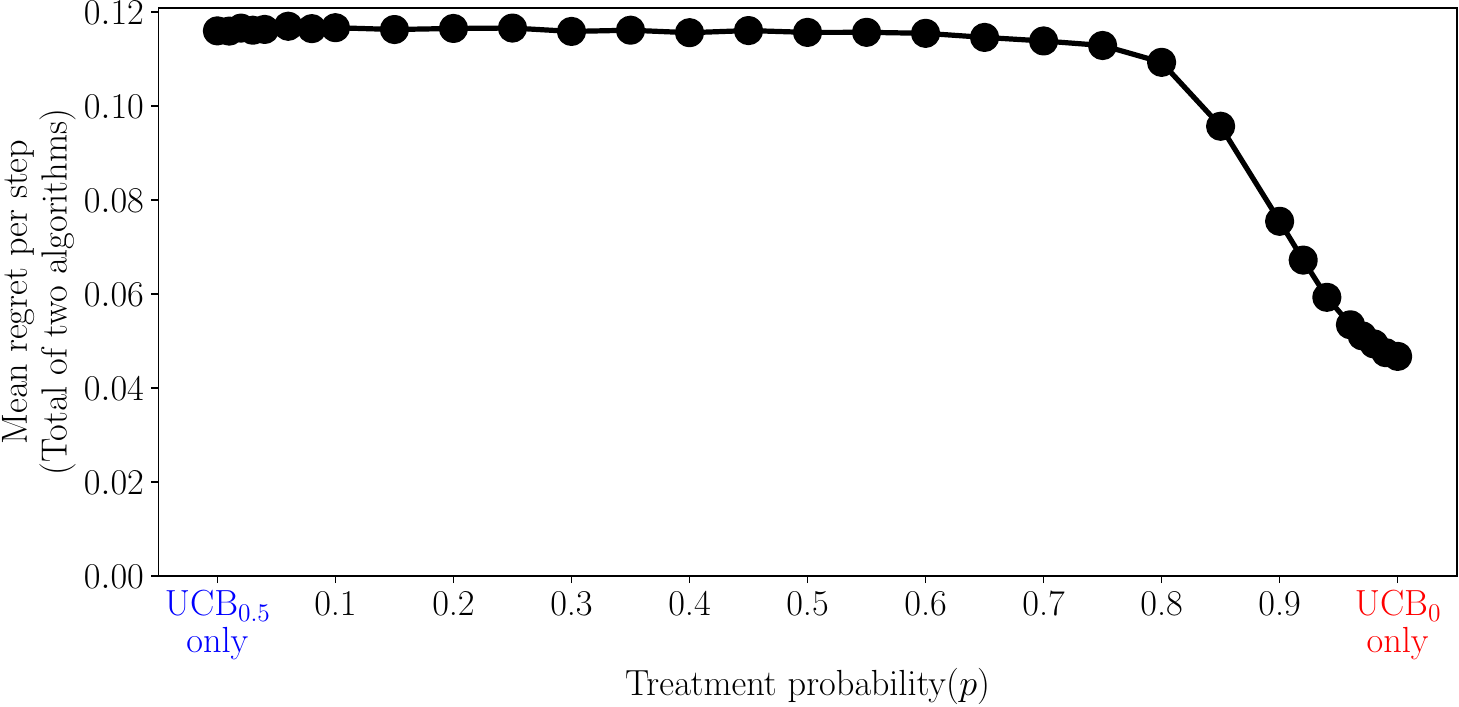}
        \end{tabular}
        \label{fig:rampup_ucb_ucb}
    }
    \caption{Ramp-up experiments with data sharing. 
    Varying the treatment probability $p$, we plot the average per-step regret of each algorithm under data sharing (top panel) and the total regret, summed over both algorithms and divided by $2T$ (bottom panel).}
    \label{fig:rampup_total}
\end{figure}

In Figure~\ref{fig:rampup_total}, we plot in the top panel the regret of each algorithm, normalized by the number of steps allocated to this algorithm, with control in blue and treatment in red. We also plot the total regret in the bottom panel, normalized by the total number of timesteps $2\horizon$. We gradually increase the treatment probability $\probassign$, where $\probassign=0$ (left end) and $\probassign=1$ (right end) correspond to running only the control or only the treatment algorithm, respectively. In both examples, the treatment algorithm appears to perform better in the experiment under data sharing. In Figure~\ref{fig:rampup_greedy_ucb}, we use the greedy algorithm and the UCB algorithm (with $\paramexplore=0$) to illustrate the regime of sign violation. We observe that while the greedy algorithm always incurs a lower regret throughout the ramp-up experiment, the total regret is not monotonic. Specifically, when the treatment probability becomes large (around $\probassign=0.96$), the total regret increases drastically, triggering a warning according to the detection criterion~\eqref{eq:detection_rule_bandit}. In contrast, Figure~\ref{fig:rampup_ucb_ucb} illustrates a ramp-up experiment with $\algoucb_{0.5}$ (control) and $\algoucb_{0}$ (treatment) in the sign preservation regime. We observe a monotonically decreasing total regret where no warning is triggered: Increasing the probability of the treatment algorithm leads to a lower total regret. All these results are consistent with our theory from Section~\ref{sec:guidelines}.

\section{Discussion}\label{sec:discuss}

We connect causal inference with bandit algorithms through analyzing A/B experiments on \ras. We identify different cases where the sign of the expected GTE estimate is correct or incorrect under symbiosis bias, and provide intuition for these results in terms of the tradeoff between exploration and exploitation.

Our causal inference perspective allows us to identify new bandit formulations. The A/B experiment setup motivates us to focus on comparing standard algorithms under \datafeedback that are optimized to run individually, contrary to extensive prior work that designs new algorithms tailored to the cooperative setting. With the goal of estimating the GTE in mind, data sharing is no longer a ``feature'' (providing flexibility to algorithm design) but a ``bug'' (introducing bias for GTE estimation), and we hope that our work will inspire future research in overcoming different types of interference when comparing the performance of online algorithms.
The bandit formulation, at the same time, provides us new views for causal inference. Evaluating adaptive learning algorithms rather than fixed policies draws attention to dynamic settings in causal inference, and as we have seen in this paper, such settings often require different tools from those used in classical A/B experiments.

While we study this problem in the context of \rss, our results serve as an important initial step towards evaluating more sophisticated algorithms and dynamic systems that interact with people. Some examples include comparing reinforcement learning algorithms for healthcare interventions, and language models that converse with people.

\paragraph{Limitations.}
We discuss several limitations of our model. First, we assume that user preferences are homogeneous, so that in the bandit formulation, users have the same reward distribution for any arm, and the algorithm does not need to account for any personalization. Imagine an extreme case where a separate personalized model is trained for each user, then data from other users will have minimal impact on this personalized model focused on a single user, eliminating symbiosis bias. Hence, the level of personalization may be another dimension that captures the impact of symbiosis bias. Second, changes in the reward distribution may also arise if users repeatedly interact with the \rs. Third, our model does not account for other types of interference such as network effects or inventory constraints in two-sided marketplaces. Our results suggest that symbiosis bias does not impact the sign of the expected GTE estimate in many cases, yet practitioners still need to carefully examine if our assumptions hold for their particular use cases before drawing such conclusions.

\paragraph{Open problems.}
Our results inspire a number of open questions. On the theoretical front, one question is to extend our results on sign preservation to general sublinear-regret algorithms beyond certain parameterized classes. One step towards this goal may involve identifying a measure to capture the balance of exploration versus exploitation for general algorithms. Another question is to extend our bias analysis using the expected regret to the probability of correctness for the stochastic regret, along the line of the simulation conducted in Section~\ref{sec:sim_prob}. 
More broadly, our work inspires future work on designing better experimental procedures. It would be useful to consider variants of Algorithm~\ref{algo:joint}, for example, by syncing data between the two algorithms only at certain timesteps, or using an adaptive meta-algorithm to assign users to the two groups as opposed to assigning uniformly at random.
So far we assume that the \ra chooses a single item (arm) to present to the user. Following prior work on interleaving items recommended by the treatment and control algorithms to users, a second question is to jointly design this interleaving ranking along with the corresponding GTE estimator under \datafeedback. 
Furthermore, it would be valuable to establish guidelines for companies to design A/B experiments, for example, in terms of how to balance the number of repetitions of an experiment versus the horizon $\horizon$ within each repetition.
Finally, our intuition from Figure~\ref{fig:conjecture} makes us conjecture that there is no experimental procedure that preserves the sign for all possible algorithms, and it is an open question to formalize (or disprove) this impossibility claim.

\subsection*{Acknowledgments}
We thank Ramesh Johari, Yusha Liu, Ashwin Pananjady, Aaditya Ramdas, and Chao Qin for helpful comments and discussions.

{\small
    \bibliographystyle{plain}
    \bibliography{references}
}

\normalsize
\appendix
\section{Additional results on the EXP3 algorithm}\label{app:bandit_algorithms}

The EXP3 algorithm, presented in Algorithm \ref{algo:exp3} and denoted by $\algoexp$, is originally designed for adversarial bandit problems~\cite{auer2002nonstochastic}, but also recently attracts attention in the stochastic bandit setting~\cite{seldin2012exp3,seldin2012pac}. By~\cite[Theorem 4]{seldin2012pac}, running the EXP3 algorithm individually yields the following regret bound.

\begin{proposition}
For every $\horizon\ge 1$, the regret of the EXP3 algorithm satisfies
\begin{align*}
    \EE{R_\horizon(\algoexp)}=O(\horizon^{2/3} (\log \horizon)^{1/2}).
\end{align*}
\end{proposition}
While careful tuning of the parameter $\varepsilon_t$ in Algorithm~\ref{algo:exp3} may further improve the regret rate, we stick with the setting $\varepsilon_t = \min\cb{1/K, K^{-2/3}t^{-1/3}}$ for simplicity.

We consider a different setting of data sharing, where the greedy algorithm uses the data generated by the EXP3 algorithm, but the EXP3 algorithm does not use data from the greedy algorithm. This one-way data sharing setting is for technical convenience. Specifically, in the definition of the EXP3 algorithm, it requires not only the observed reward $Y_t$ but also the sampling probability $\widetilde{\rho}_t(\idxarm)$ associated with the selected arm $\idxarm$ in Line~\ref{exp3-update} of Algorithm~\ref{algo:exp3}. However, this sampling probability is not well-defined for the samples shared by the greedy algorithm. Hence, we assume for simplicity that the EXP3 algorithm only utilizes data generated by itself. 

Under this one-way data sharing, we denote the regret of the greedy algorithm by $\Expect[\regret_\horizon(\algogreedy\given \algoexp\rightarrow \algogreedy)]$ to emphasize that only the EXP3 algorithm supplies data to the greedy algorithm. Since the greedy algorithm does not supply data to the EXP3 algorithm, we have $\Expect[\regret_\horizon(\algoexp\given \algoexp\rightarrow \algogreedy)] = \Expect[\regret_\horizon(\algoexp)]$.
We now provide regret guarantees for the EXP3 algorithm and the greedy algorithm under one-way data sharing.

\begin{theorem}\label{thm:greedy+exp3}
Consider jointly running the greedy algorithm $\algogreedy$ and the EXP3 algorithm $\algoexp$ under one-way data sharing, where only the EXP algorithm supplies data to the greedy algorithm. The expected regrets of the two algorithms under one-way data sharing satisfy
\begin{subequations}
\begin{align}
   & \EE{R_T(\algogreedy \mid \algoexp\rightarrow \algogreedy )} = O(1), \label{eq:thm_violate_grd_exp_ub}\\
   & \EE{\regret_\horizon(\algoexp \mid \algoexp\rightarrow \algogreedy)}= \Omega(T^{2/3}).\label{eq:thm_violate_grd_exp_lb}
\end{align} 
\end{subequations}
\end{theorem}
The proof of this theorem is provided in \Cref{sec:proof_thm_greedy+exp3}.
As in the case of jointly running the greedy algorithm with the \egreedy algorithm (Theorem~\ref{thm:greedy+e-greedy}) or the UCB algorithm (Theorem~\ref{thm:greedy+UCB}), this result again demonstrates that under (one-way) data sharing, the regret of the greedy algorithm reduces to a constant, while the regret of the EXP3 algorithm remains unchanged up to logarithmic factors, resulting in a sign violation. The proof is similar to those of Theorems~\ref{thm:greedy+e-greedy} and~\ref{thm:greedy+UCB}, by establishing and invoking a general result (see Proposition~\ref{prop:greedy+general_one_way} in \Cref{sec:proof_thm_greedy+exp3}) for one-way data sharing analogous to \Cref{thm:greedy+general}.

\begin{algorithm}[tb]
    \DontPrintSemicolon
    
    Initialize $\hat{Y}_0(\idxarm)=0$ for each $\idxarm\in [\numarms]$\;
    \For{\timestep=1, \ldots, \horizon}{
        Compute $\varepsilon_t =\min\{1/K, K^{-2/3}t^{-1/3}\}$\;
        Compute sampling probability $\widetilde{\rho}_t(\idxarm)=(1-K\varepsilon_t)\frac{e^{\varepsilon_{t-1}\hat{Y}_{t-1}(\idxarm)}}{\sum_{\idxarm'}e^{\varepsilon_{t-1}\hat{Y}_{t-1}(\idxarm')}}+\varepsilon_t$ for each $\idxarm\in [\numarms]$\;
        Select arm $\armselect_t$ according to the sampling distribution $\widetilde{\rho}_t$ and observe reward $Y_t$\;
        Update $\hat{Y}_t (\idxarm)=\hat{Y}_{t-1}(\idxarm)+\frac{Y_t}{ \tilde{\rho}_t(\idxarm)}\cdot \indicator\{\idxarm=\armselect_\timestep\}$ for each $\idxarm\in [\numarms]$\label{exp3-update}\;
    }
    \caption{The EXP3 algorithm~\cite{seldin2012exp3,seldin2012pac}.
    \label{algo:exp3}
}
\end{algorithm}

\section{Description of \ts algorithm}
\label{appendix:TS}
Algorithm~\ref{algo:ts} describes \ts algorithm with Bernoulli rewards~\cite[Algorithm1]{agrawal2012thompson} used in our simulation in \Cref{sec:sim_ts}.
\begin{algorithm}[tb]
    \DontPrintSemicolon
    Initialize $S_\idxarm=F_\idxarm=0$ for each arm $\idxarm\in [\numarms]$\;
    \For{\timestep=1, \ldots, \horizon}{
        Sample $\theta_\idxarm(\timestep)\sample \text{Beta}(S_\idxarm + 1, F_\idxarm + 1)$ for each arm $\idxarm\in [\numarms]$\;
        Select arm $\armselect_\timestep = \argmax_\idxarm \theta_\idxarm(t)$ and observe reward $r_t$

        \uIf{$r_t = 1$}{Update $S_{\armselect_\timestep} \leftarrow S_{\armselect_\timestep} + 1$}
        \uElse{Update $F_{\armselect_\timestep}\leftarrow F_{\armselect_\timestep} + 1$}
    }
    \caption{\ts with Bernoulli rewards~\cite[Algorithm1]{agrawal2012thompson}.
    \label{algo:ts}
}
\end{algorithm}

\section{Proofs}\label{app:proofs}
In this section, we present the proofs of our theoretical results.
\subsection{Preliminaries}

We set up notation followed by preliminary properties under data sharing that are used in subsequent proofs. Recall from Section~\ref{sec:classic_algorithms} that we define all algorithms such that whether an algorithm runs individually or two algorithms run jointly, each arm is pulled at least once after $\numarms$ timesteps. The regret incurred in the first $\numarms$ timesteps is a constant, and we often omit the first $\numarms$ timesteps in our regret analysis.

\subsubsection{Notation}

To describe how the stochastic rewards for each arm are generated, we follow the \emph{reward tape} construction~\cite{slivkins2019introduction}. Each arm $\idxarm\in [\numarms]$ is associated with a reward tape of length $2\horizon$ (because each arm is pulled at most $2\horizon$ times when two algorithms run jointly) from left to right, where each cell $\idxsample\in [2\horizon]$ of the tape contains a random sample $X_{\idxarm, \idxsample}\sample \dist_\idxarm$. Whenever arm $\idxarm$ is pulled, we move one cell to the right on the respective reward tape, and obtain a reward from the new cell. When two algorithms run jointly and they pull the same arm at a timestep, we assume without loss of generality that algorithm $1$ receives the reward from the new cell first followed by algorithm $2$ receiving the reward from the next cell on its right. In our proofs, we always consider the empirical mean of an arm using data collected by both algorithms, so the order that the two algorithms collect rewards within the same timestep does not matter. In Algorithm~\ref{algo:joint}, we used $Y_{\timestep}$ to denote the reward observed at timestep $t$. We clarify the difference between notation $X$ and $Y$: we use $Y_{1,t}$ or $Y_{2,t}$ to denote the reward received at timestep $\timestep$ by algorithm $\algo_1$ or $\algo_2$, respectively. On the other hand, $X_{k,s}$ denotes the $s$-th reward drawn from the tape of arm $k$, corresponding to the $s$-th pull of that arm. Let $\overline{X}_{k,s} \defn \frac{1}{s}\sum_{i=1}^s \overline{X}_{\idxarm, s}$ be the empirical mean of the rewards from the first $s$ pulls of arm $k$. 

Recall that the random variable $N_{k,t}(\algo\given (\algo, \algo'))$ denotes the number of pulls on arm $k\in [\numarms]$ before timestep $\timestep$ by algorithm $\algo$, when it runs jointly with another algorithm $\algo'$. Denote by random variable $\armselect_\timestep(\algo\given (\algo, \algoalt))$ the arm pulled by algorithm $\algo$ at timestep $t$ when it runs jointly with algorithm $\algoalt$.
Then we have
\begin{align*}
    N_{\idxarm,t}(\algo\given (\algo, \algoalt)) = \sum_{i=1}^{\timestep-1}\indicator\{\armselect_i(\algo\given (\algo, \algoalt)) = \idxarm\}.
\end{align*}
Define the number of pulls combined by the two algorithms by
\begin{align*}
    N_{k,t}(\algotot\given (\algo, \algo')) \defn N_{k,t}(\algo\given (\algo, \algo')) + N_{k,t}(\algo'\given (\algo, \algo')).
\end{align*}
We use the shorthand notation $N_\idxarm(\algo\given (\algo, \algo')) \defn N_{\idxarm, \horizon+1}(\algo\given (\algo, \algo'))$ for the total number of pulls by algorithm $\algo$, and likewise $N_\idxarm(\algotot\given (\algo, \algo')) \defn N_{\idxarm, \horizon+1}(\algotot\given (\algo, \algo'))$ for the two algorithms in total.
For the \egreedy algorithm, we call a timestep an ``exploration step'', if the \egreedy algorithm samples an arm uniformly at random at this timestep (which happens with probability $\probexplore_\timestep$ after the first $\numarms$ timesteps). We denote by $N_{1,t}^R(\algoegreedy)$ the number of pulls by exploration steps of the \egreedy algorithm, or $N_{1,t}^R(\algoegreedy \given (\algoegreedy, \algo))$ if the \egreedy algorithm runs jointly with another algorithm $\algo$. We have the deterministic relation that for any arm $\idxarm$, timestep $\timestep$, and algorithm $\algo$,\begin{align*}
N_{\idxarm,t}^R(\algoegreedy) & \le N_{\idxarm,t}(\algoegreedy)\\
N_{\idxarm,t}^R(\algoegreedy \given (\algoegreedy, \algo)) & \le N_{\idxarm,t}(\algoegreedy \given (\algoegreedy, \algo)).
\end{align*}
We use the shorthand notation $\armselect_\timestep(\algo), \varnumpulls_{\idxarm,\timestep}(\algo)$, and $N_{k,t}(\algotot)$ respectively for $\armselect_\timestep(\algo\given (\algo, \algoalt)), N_{\idxarm}(\algo\given (\algo, \algoalt))$, and $N_{k,t}(\algotot\given (\algo, \algo'))$ when it is clear from the context that algorithm $\algo$ runs jointly with algorithm $\algo'$.

Throughout the proofs, we assume without loss of generality that arm $1$ is the best arm. Then $\gap_k = \mu_1 - \mu_k$ for each suboptimal arm $\idxarm\ne 1$.
For the UCB algorithm, we denote by $c_{\timestep, \numsamples}$ the length of the confident interval at timestep $t$ for an arm with $s$ samples as defined in~\eqref{eq:ucb_def_conf_interval}. Since that $\abs{\history_{\timestep}} = 2(\timestep-1)$ because there are $2(\timestep-1)$ pulls in total by both algorithms before timestep $\timestep$, we have
\begin{align}
    c_{t, s}^\paramexplore=
    \begin{cases}
        \sqrt{\frac{2 \log (2t-2)}{s}} & \text{if }\alpha = 0,\\
        \sqrt{\frac{2((2t-2)^\alpha - 1)}{\alpha s}} & \text{if } 0< \alpha \le 1.
    \end{cases} \label{eq:confidence_define}
\end{align}

The following result shows that a higher value of $\paramexplore$ corresponds to a wider confidence interval for the UCB algorithm. This monotonic property aligns with our previous intuition that the value of $\paramexplore$ captures the level of exploration.

\begin{fact}\label{fact:confidence_increase_in_alpha}
    For any $\timestep > 2$ and any $s \ge 1$, the length of the confidence interval $c_{\timestep, \numsamples}^\paramexplore$ is strictly increasing in $\paramexplore$ on $[0, 1]$.
\end{fact}

\subsubsection{Basic concentration inequalities}
We use the following version of the Bernstein's inequality in our proofs.

\begin{fact}[Bernstein's inequality; Fact 2 in~\cite{auer2002finite}]\label{fact:bernstein}
    Let $X_1, \ldots, X_n$ be random variables bounded in $[0, 1]$ and
    \begin{align*}
        \sum_{i=1}^n \var[X_i\given X_{i-1}, \ldots, X_1] = \sigma^2.
    \end{align*}
    Define the sum $S_n \defn X_1 + \ldots+ X_n$. Then for every $a > 0$, we have
    \begin{subequations}
    \begin{align}\label{eq:bernstein_gt}
        \PP{S_n \ge \Expect[S_n] + a} \le \exp\p{-\frac{a^2/2}{\sigma^2 + a/2}}.
    \end{align}
    and
    \begin{align}\label{eq:bernstein_le}
        \PP{S_n \le \Expect[S_n] - a} \le \exp\p{-\frac{a^2/2}{\sigma^2 + a/2}}.
    \end{align}
    \end{subequations}
\end{fact}

\subsubsection{Properties under \datafeedback}\label{sec:properties_data_sharing}

We now present a few regret bounds for running two algorithms under \datafeedback. These results are used in subsequent proofs. 
The first result provides an upper bound for the UCB algorithm when it runs jointly with any other algorithm.

\begin{proposition}\label{prop:UCB_with_other}
Consider any $\paramexplore\in [0, 1]$. When we jointly run the UCB algorithm $\algoucb_\paramexplore$ with any other algorithm $\mathcal{A}$, the expected regret of $\algoucb_\paramexplore$ satisfies
\begin{align*}
    &\EE{R_T(\algoucb_\paramexplore \mid (\mathcal{A}, \algoucb_\paramexplore))} 
= \begin{cases}
    O( \log T)  & \text{if } \alpha = 0, \\
    O( T^\alpha) & \text{if } 0 < \alpha \le 1.
    \end{cases}
\end{align*}
\end{proposition}
The proof of this proposition is provided in \Cref{sec:proof_prop_UCB_with_other}. Intuitively, this result states that any additional samples do not hurt the regret rate of the UCB algorithm.

We then establish a lower bound for the UCB algorithm under \datafeedback as follows. 

\begin{proposition}\label{prop:UCB+other_lb}
Consider any $\paramexplore\in [0, 1]$. When we jointly run the UCB algorithm $\algoucb_\paramexplore$ with any other algorithm $\algo$, the sum of the expected regrets of the two algorithms satisfies
\begin{align*}
&\EE{R_T(\algoucb_\paramexplore \mid (\algo, \algoucb_\paramexplore))} + \EE{R_T(\algo \mid (\algo, \algoucb_\paramexplore))} 
= \begin{cases}
    \Omega( \log T) & \text{if } \alpha = 0, \\
    \Omega( T^\alpha) & \text{if } 0 < \alpha \le 1.
\end{cases}
\end{align*}
\end{proposition}
The proof of this proposition is provided in \Cref{sec:proof_prop_UCB+other_lb}.

We also establish upper and lower bounds for the \egreedy algorithm: 
\begin{proposition}
    \label{prop:egreedy+other}
Consider any $\paramexplore\in [0, 1]$ and any $C \ge \max\{120K,\, 32K/\gapmin^2\}$. When we jointly run the \egreedy algorithm $\algoegreedy_{\paramexplore, C}$ with any other algorithm $\mathcal{A}$, the expected regret of $\algoegreedy_\paramexplore$ satisfies
\begin{align*}
    &\EE{R_T(\algoegreedy_\paramexplore \mid (\mathcal{A}, \algoegreedy_\paramexplore))} 
= \begin{cases}
    O( \log T)  & \text{if } \alpha = 0, \\
    O( T^\alpha) & \text{if } 0 < \alpha \le 1.
    \end{cases}
\end{align*}
\end{proposition}
The proof of this proposition is provided in \Cref{sec:proof_prop_egreedy+other}.

\begin{proposition}\label{prop:egreedy+other_lb}
Consider any $\paramexplore\in [0, 1]$ and any $C > 0$. When we jointly run the \egreedy algorithm $\algoegreedy_{\paramexplore, C}$ with any other algorithm $\mathcal{A}$, then the expected regret of $\algoegreedy_{\paramexplore, C}$ satisfies
    \begin{align*}
        \Expect[\regret_\horizon(\algoegreedy_\paramexplore)\given (\algo, \algoegreedy_\paramexplore)] = \begin{cases}
            \Omega(\log\horizon) & \text{if }\paramexplore=0,\\
            \Omega(\horizon^\paramexplore) & \text{if } 0 < \paramexplore \le 1.
        \end{cases}
    \end{align*}
\end{proposition}
The proof of this proposition is provided in \Cref{sec:proof_prop_egreedy+other_lb}. We lower bound the regret incurred in the exploration steps, which are not affected by data sharing with another algorithm.

\subsection{Proof of Proposition~\ref{prop:run_individually}}\label{sec:proof_prop_run_individually}

We outline the proof for the UCB and \egreedy algorithms separately.

\paragraph{Upper bound for UCB.}
The proof is similar to our more general results that provide an upper bound (Proposition~\ref{prop:UCB_with_other}) and lower bound (Proposition~\ref{prop:UCB+other_lb} for the UCB algorithm under data sharing. We provide a brief description on how to modify these proofs for the setting when a single UCB algorithm runs individually.

For the upper bound, we follow the proof of Proposition~\ref{prop:UCB_with_other} in \Cref{sec:proof_prop_UCB_with_other}. At any timestep $\timestep$, instead of considering $2\timestep$ samples from the two algorithms, we only need to consider $\timestep$ samples from the single UCB algorithm. The number of pulls on any suboptimal arm becomes (cf. Equation~\eqref{eq:ucb+general_ub_num_pulls_expression}):
\begin{align*}
    N_{k,T+1}(\algoucb_\paramexplore) &\leq \ell+\numarms + \sum_{t=\numarms+1}^{\infty} \sum_{s=1}^{t-1} \sum_{s_k=\ell}^{t-1}\indicator\left\{\overline{X}_{1,s}+c_{t, s}^\paramexplore \leq \overline{X}_{k, s_k}+c_{t, s_k}^\paramexplore\right\},
\end{align*}
The rest of the proof follows the same step as the proof of Proposition~\ref{prop:UCB_with_other} in \Cref{sec:proof_prop_UCB_with_other}.

\paragraph{Lower bound for UCB.}
The lower bound follows the proof of Proposition~\ref{prop:UCB+other_lb} in \Cref{sec:proof_prop_UCB+other_lb}, where we simply replace $N_{1,\timestep}(\algotot\given (\algo, \algoucb_\paramexplore))$ and $N_{2,\timestep}(\algotot\given (\algo, \algoucb_\paramexplore))$ by $N_{1,\timestep}(\algoucb_\paramexplore)$ and $N_{2,\timestep}(\algoucb_\paramexplore)$, because all arm pulls are from the single UCB algorithm when it runs individually.

\paragraph{Upper bound for \egreedy.}
The upper bound of $O(\log\horizon)$ for the \egreedy algorithm when $\paramexplore= 0$ is proved in~\cite[Theorem 3]{auer2002finite}. The proof of the upper bound $O(\horizon^\paramexplore)$ when $0 < \paramexplore \le 1$ follows the same steps.

\paragraph{Lower bound for \egreedy.} 

The bound follows the proof of \Cref{prop:egreedy+other_lb} in \Cref{sec:proof_prop_egreedy+other_lb}, with the exploration probability $\probexplore_\timestep= \min \left\{1, \frac{\Const}{(2t-2)^{1-\paramexplore}}\right\}$ under data sharing replaced by $\probexplore_\timestep= \min \left\{1, \frac{\Const}{(t-1)^{1-\paramexplore}}\right\}$ for a single algorithm. In the proof, we only analyze the exploration steps, which are unaffected with or without data sharing.

\subsection{Proof of Theorem \ref{thm:greedy+e-greedy}}\label{sec:proof_thm_greedy+e-greedy}

We prove the lower bound for the \egreedy algorithm and the upper bound for the greedy algorithm separately.

\paragraph{Lower bound~\eqref{eq:thm_violate_egreedy_greedy_lb} for \egreedy.}
Applying Proposition~\ref{prop:egreedy+other_lb} with the other algorithm $\algo$ being the greedy algorithm yields the desired lower bound~\eqref{eq:thm_violate_egreedy_greedy_lb} for the \egreedy algorithm.

\paragraph{Upper bound~\eqref{eq:thm_violate_grd_egreedy_ub} for greedy.}
We apply Theorem~\ref{thm:greedy+general} to establish the desired upper bound by verifying condition~\eqref{eq:assume_high_probability_bound} for the \egreedy algorithm. Specifically, it suffices to prove condition~\eqref{eq:assume_high_probability_bound} for every $\horizon \ge (C+1)^2$. The derivation of condition~\eqref{eq:assume_high_probability_bound} largely follows~\cite[Theorem 3]{auer2002finite}. To lower bound the number of times that the optimal arm $1$ is pulled, we provide a lower bound to the number of times $N_{1,\horizon}^R(\algoegreedy_{\alpha, C})$ that arm $1$ is pulled in an exploration step.
We define $\cumprob(\horizon)$ to be the expected value of $N_{\idxarm,\horizon}^R(\algoegreedy_{\alpha, C})$:
\begin{align}
    \cumprob(\horizon) \defn \EE{N_{1,\horizon}^R(\algoegreedy_{\alpha, C})}=\frac{1}{\numarms} \sum_{\timestep=K+1}^{\horizon-1} \epsilon_\timestep.\label{eq:greedy+egreedy_expected_pulls_exploration}
\end{align}
The variance of $N_{\idxarm,\horizon}^R(\algoegreedy_{\alpha, C})$ is:
\begin{align*}
    \var\sqb{N_{1,\horizon}^R(\algoegreedy_{\alpha, C})}=\sum_{\timestep=K+1}^{\horizon-1} \frac{\epsilon_\timestep}{K}\left(1-\frac{\epsilon_\timestep}{K}\right) \leq \frac{1}{K} \sum_{\timestep=\numarms+1}^{\horizon-1} \epsilon_\timestep = \cumprob (\horizon).
\end{align*}
By Bernstein's inequality~\eqref{eq:bernstein_le} in \Cref{fact:bernstein}, we have
\begin{equation}
    \PP{N_{k,\horizon}^R(\algoegreedy_{\alpha, C}) \leq \frac{\cumprob(\horizon)}{2}} \leq e^{-\cumprob(\horizon) / 10}. \label{eq:grd_egreedy_ub_bernstein}
\end{equation}
We now provide a lower bound for $\cumprob(\horizon)$. 
Plugging in $\epsilon_t=\min \left\{1, C/(2t-2)^{1-\alpha}\right\} \geq \min \left\{1, C/(2t)\right\}$ to~\eqref{eq:greedy+egreedy_expected_pulls_exploration}, we have
\begin{align*}
    \cumprob(\timestep)=\frac{1}{K} \sum_{\timestep=K+1}^{\horizon-1} \epsilon_\timestep 
    \geq \frac{C}{2K} \sum_{\timestep=\lfloor C \rfloor+1}^{\horizon-1} \frac{1}{\timestep} 
    \geq \frac{C}{2K}\p{\log\horizon-\log(\lfloor C \rfloor+1)}.
\end{align*}
For any $\horizon\ge (C+1)^2$, we have $\frac{\log\horizon}{2}\ge \log(\floor{C} + 1)$ and hence
\begin{align}
    \cumprob(\horizon)\ge \frac{C\log\horizon}{4\numarms}.\label{eq:grd_egreedy_cum_prob}
\end{align}
Combining~\eqref{eq:grd_egreedy_cum_prob} with~\eqref{eq:grd_egreedy_ub_bernstein}, we have
\begin{align*}
    \PP{N_{k,\horizon}(\algoegreedy_{\alpha, C}) \le \frac{4}{\gapmin^2}\log\horizon}& \le \PP{N_{k,\horizon}^R(\algoegreedy_{\alpha, C}) \le \frac{4}{\gapmin^2}\log\horizon} \\
    & \stackrel{\stepone}{\le} \PP{N_{k,\horizon}^R(\algoegreedy_{\alpha, C})\le \frac{\cumprob(\horizon)}{2}}\\
    & \le e^{-\cumprob(\horizon) / 10}
    \le \horizon^{-C/40K} \stackrel{\steptwo}{\le} \frac{1}{\horizon^2},
\end{align*}
where steps~\stepone and~\steptwo are due to the assumption that $C\ge\max\left\{80K, \frac{32\numarms}{\gapmin^2}\right\}$, proving condition~\eqref{eq:assume_high_probability_bound}. Applying Theorem~\ref{thm:greedy+general} yields the upper bound~\eqref{eq:thm_violate_grd_egreedy_ub} for the greedy algorithm.

\subsection{Proof of Theorem~\ref{thm:greedy+UCB}}\label{sec:proof_thm_greedy+UCB}

We prove the upper bound for the UCB algorithm and the lower bound for the greedy algorithm separately.

\paragraph{Lower bound~\eqref{eq:thm_violate_ucb_greedy_lb} for UCB.}
By \Cref{prop:UCB+other_lb}, we have
\begin{align}
\EE{R_T(\algoucb_\paramexplore \given(\algogreedy,\algoucb_\paramexplore))} + \EE{R_T(\algogreedy\given(\algogreedy,\algoucb_\paramexplore))} = \begin{cases}
    \Omega( \log T) & \text{if } \alpha = 0,\\
    \Omega(\horizon^{\alpha}) & \text{if } 0 < \paramexplore\le 1.
\end{cases}\label{eq:UCB+grd_lb_sum}
\end{align}
We first assume that the upper bound~\eqref{eq:thm_violate_ucb_greedy_ub} for the greedy algorithm holds, which is proved subsequently. Combining~\eqref{eq:thm_violate_ucb_greedy_ub} with~\eqref{eq:UCB+grd_lb_sum} yields the desired lower bound~\eqref{eq:thm_violate_ucb_greedy_lb} for the UCB algorithm.

\paragraph{Upper bound~\eqref{eq:thm_violate_ucb_greedy_ub} for greedy.}
We apply Theorem~\ref{thm:greedy+general} to establish the desired upper bound by verifying condition~\eqref{eq:assume_high_probability_bound} for the UCB algorithm. Specifically, it suffices to prove condition~\eqref{eq:assume_high_probability_bound} for every $\horizon \ge \max\{2(\numarms+1), \const_1\}$ for some constant $\const_1 > 0$ that is specified later.

The proof largely follows~\cite[Appendix C]{guo2022demonstrator}, with the difference that we consider an anytime UCB algorithm, whereas the UCB algorithm considered in~\cite{guo2022demonstrator} assumes the knowledge of the horizon $T$ and uses a factor of $\sqrt{\log\horizon}$ instead of $\sqrt{\log\timestep}$ for the confidence bound. In the anytime algorithm, the empirical mean of an arm no longer concentrates around the true mean when the timestep $\timestep$ is small. However, our goal is weaker as we only need to show that the optimal arm is pulled $\Omega(\horizon)$ times (as opposed to $\horizon - o(\horizon)$ times). Therefore, it suffices to show that the optimal arm is pulled $\Omega(\horizon)$ times during the time interval $[\horizon/2, \horizon]$.
We show that the empirical mean of each arm always concentrates around its true mean during this time interval $[\horizon/2, \horizon]$ with high probability. Conditional on this event, each suboptimal arm can only be pulled a limited number of times in this interval. As a result, the optimal arm must be pulled $\Omega(T)$ times with high probability.

Recall from~\eqref{eq:confidence_define} the length $\conf_{\timestep, \numsamples}^\paramexplore$ of the confidence interval is defined as:
\begin{align}\label{eq:confidence_define_recall_greedy+ucb}
    \conf^{\alpha}_{t, s}=
    \begin{cases}
        \sqrt{\frac{2 \log (2t-2)}{s}} & \text{if }\alpha = 0,\\
        \sqrt{\frac{2((2t-2)^\alpha - 1)}{\alpha s}} & \text{if } 0 < \alpha \le 1.
    \end{cases}
\end{align}
For each arm $k\in [K]$, we define the event
\begin{align*}
    \mathcal{E}_k = \cb{\text{there exists a $t \in [T/2, T]$ such that } \abs{\overline{X}_{k,N_{k,t}(\algotot)} - \mu_k} > \conf^{\alpha}_{t, N_{k,t}(\algotot)} }.
\end{align*}
In words, event $\event_\idxarm$ is a ``bad'' event that the empirical mean of arm $\idxarm$ does not concentrate around its true mean at some timestep between $\horizon/2$ and $\horizon$. 
We bound the probability of each event $\event_\idxarm$. In particular, for $T \geq 2(K+1)$ (so that there is at least one sample from each arm at time $\ceil{\horizon/2}$):
\begin{align}
    \PP{\event_k}
    &\leq \sum_{t = \lceil T/2 \rceil}^T \PP{\abs{\overline{X}_{k,N_{k,t}(\algotot)} - \mu_k} > \conf^{\alpha}_{t, N_{k,t}(\algotot)}}\nonumber
    \\
    & \stackrel{\stepone}{\leq} \sum_{t = \lceil T/2 \rceil}^T \sum_{s = 1}^{2(t-1)} \PP{\abs{\overline{X}_{k,s} - \mu_k} > \conf^{0}_{t, s}}\nonumber\\
    & \stackrel{\steptwo}{\leq} \sum_{t = \lceil T/2 \rceil}^T \sum_{s = 1}^{2(t-1)}\frac{1}{(2t-2)^4}
    \leq \sum_{t = \lceil T/2 \rceil}^T \frac{1}{(2t-2)^3}
    \le \frac{\horizon}{2}\cdot \frac{1}{(\horizon-2)^3}
    \stackrel{\stepthree}{\leq} \frac{4}{T^2},\label{eq:ucb+greedy_ucb_ub_prob_event}
\end{align}
where step~\stepone is true by a union bound with \Cref{fact:confidence_increase_in_alpha} that the value of $\conf_{\timestep, \numsamples}^\paramexplore$ increases in $\paramexplore\in [0, 1]$; step~\steptwo follows from Hoeffding's inequality; and step~\stepthree is true because $\horizon \ge 2(\numarms+1) > 4$.
Define the event $\mathcal{E}_0 \defn (\mathcal{E}_1 \cup \mathcal{E}_2\union \cdots \cup \mathcal{E}_K)^c$, which is a ``good'' event that the empirical mean of all arms concentrate around their true means during the time interval $[\horizon/2, \horizon]$. Applying a union bound to~\eqref{eq:ucb+greedy_ucb_ub_prob_event}, we have that for $T \geq 2(K+1)$,
\begin{align*}
    \PP{\event_0} \geq 1 - \frac{4K}{T^2}.
\end{align*}
Now we bound the number of pulls on any suboptimal arm $k \neq 1$ conditional on the event $\event_0$. We consider the additional arm pulls when arm $\idxarm$ has already been pulled $\ell^{\alpha}_{k}$ times, where
\begin{align*}
    \ell^{\alpha}_{k} = \begin{cases}
        \frac{8 \log (2T - 2)}{\Delta_k^2} & \text{if }\alpha = 0\\
        \frac{8((2T - 2)^\alpha - 1}{\Delta_k^2} & \text{if } 0 < \alpha \le 1.
    \end{cases}
\end{align*}
For any timestep $\timestep \in [T/2, T]$, if $N_{k,t}(\algotot) > \ell^{\alpha}_{k}$, then using the definition of the confidence interval $c_{\timestep, \numsamples}^\paramexplore$ from~\eqref{eq:confidence_define_recall_greedy+ucb}, it can be verified that $2\conf^{\alpha}_{N_{k,t}(\algotot)} < \Delta_k$. Therefore, conditional on event $\mathcal{E}_0$, for any timestep $t \in [T/2, T]$, if $N_{k,t}(\algotot) > \ell^{\alpha}_{k}$, then arm $k$ cannot be pulled at timestep $t$, because
\begin{align*}
    \overline{X}_{k,N_{k,t}(\algotot)} + \conf^{\alpha}_{t, N_{k,t}(\algotot)}
    \stackrel{\stepone}{\leq} \mu_k + 2\conf^{\alpha}_{t, N_{k,t}(\algotot)}
    < \mu_k + \Delta_k
    = \mu_1
    \stackrel{\steptwo}{\leq} \overline{X}_{1,N_{1,t}(\algotot)} + \conf^{\alpha}_{t, N_{1,t}(\algotot)},
\end{align*}
where step~\stepone is true conditional on $\event_\idxarm^c$ and step~\steptwo is true conditional on $\event_1^c$.
Therefore, conditional on event $\mathcal{E}_0$, during the time interval $[T/2, T]$, each suboptimal arm $k$ can be pulled at most $\ell^{\alpha}_{k}$ times. Therefore, conditional on event $\mathcal{E}_0$, arm 1 must be pulled at least $T/2 - \sum_k \ell^{\alpha}_{k}$ times, and there exists a constant $c_1 > 0$ (dependent only on $\{\gap_\idxarm\}_{\idxarm\in[\numarms]}$ and $\paramexplore$), such that for any $T >c_1$, 
\begin{align*}
    T/2 - \sum_{\idxarm\ne 1} \ell^{\alpha}_{k} > \frac{4}{\gapmin^2}\log T.
\end{align*}
Therefore, for $\horizon > \max\cb{2(K+1), c_1}$,
\begin{align*}
    \Prob\left(\varnumpulls_{1,T}(\algoucb_\paramexplore \given (\algoucb_\paramexplore, \algogreedy)) < \frac{4}{\gapmin^2}\log T\right) 
    \leq \PP{\mathcal{E}_0^c}
    \leq 
    \frac{4K}{T^2},
\end{align*}
proving condition~\eqref{eq:assume_high_probability_bound}. Applying Theorem~\ref{thm:greedy+general} yields the upper bound~\eqref{eq:thm_violate_ucb_greedy_ub} for the greedy algorithm.

\subsection{Proof of Theorem~\ref{thm:greedy+general}}\label{sec:proof_thm_greedy+general}

In this proof, we decompose a pull on any suboptimal arm $\idxarm$ by considering two causes: the empirical mean of arm $\idxarm$ does not concentrate, or the empirical mean of the optimal arm $1$ does not concentrate. In the latter case, condition~\eqref{eq:assume_high_probability_bound} indicates that the other algorithm $\algo$ supplies $\Omega(\log\horizon)$ samples for the optimal arm $1$, and hence the probability that the empirical mean of the optimal arm $1$ does not concentrate can be bounded. In the former case, the algorithm pulls arm $\idxarm$ because the empirical mean of arm $\idxarm$ does not concentrate. We note that each pull contributes towards the concentration of this arm, and apply a union bound over the number of samples on arm $\idxarm$ (as opposed to over the timesteps) to obtain a constant regret.

\paragraph{Decomposing the cause of pulling a suboptimal arm.}
Under data sharing, the number of pulls for any suboptimal arm $\idxarm\ne 1$ by the greedy algorithm is
\begin{align}
    N_{k}(\algogreedy) &= \sum_{\timestep=1}^\horizon \indicator\{I_t(\algogreedy) = \idxarm\}\nonumber\\
    & = \sum_{\timestep=1}^\horizon \indicator\{I_t(\algogreedy)=\idxarm, \; \overline{X}_{k,N_{k,t}(\algotot)} > \overline{X}_{1,N_{1,t}(\algotot)}\}\nonumber\\
    & \le \underbrace{\sum_{\timestep=1}^\horizon \indicator\left\{I_t(\algogreedy)=\idxarm, \; \overline{X}_{k,N_{k,t}(\algotot)} > \mu_\idxarm + \frac{\gap_\idxarm}{2}\right\}}_{\term_1} 
    + \underbrace{\sum_{\timestep=1}^\horizon \indicator\left\{I_t(\algogreedy)=\idxarm, \;\overline{X}_{1,N_{1,t}(\algotot)} < \mu_1 - \frac{\gap_\idxarm}{2}\right\}}_{\term_2},\label{eq:general_decompose_two_terms}
\end{align}
where the last inequality is true by a union bound. In~\eqref{eq:general_decompose_two_terms}, term $\term_1$ corresponds to the event that arm $\idxarm$ does not concentrate, and term $\term_2$ corresponds to the event that the optimal arm $1$ does not concentrate. We now bound these two terms separately.

\paragraph{Bounding term $\term_1$.}
We apply a similar argument as in~\cite[Thm 4.1]{yang2024abstention}. We decompose term $\term_1$ as:
\begin{align}
    \term_{1} = \sum_{\timestep=1}^\horizon \sum_{s=0}^{2(\timestep-1)} \indicator\left\{I_t(\algogreedy) = \idxarm,\; \overline{X}_{\idxarm, s} > \meantrue_\idxarm + \frac{\gap_\idxarm}{2},\; N_{k,t}(\algotot) = s\right\},\label{eq:general_term_one_expression}
\end{align}
where in the summation $s$ goes up to $2(t-1)$, because there are two samples collected per timestep under data sharing.
For any fixed $s\in \cb{0, \dots, 2(\horizon-1)}$, we have the deterministic relation
\begin{align}
    \sum_{\timestep=1}^\horizon \indicator\left\{I_t(\algogreedy) = \idxarm,\; N_{k,t}(\algotot) = s\right\} \le 1,\label{eq:general_moving_at_most_one}
\end{align}
because whenever the event $\indicator\{I_t(\algogreedy) = \idxarm,\; N_{k,t}(\algotot) = s\}$ happens for the first time, the greedy algorithm pulls arm $\idxarm$, and hence the total number of samples on arm $\idxarm$ is at least $(s+1)$ for all later timesteps.
Rearranging~\eqref{eq:general_term_one_expression}, we have
\begin{align}
    \term_{1} &\le \sum_{s=0}^{2(\horizon-1)}\sum_{\timestep=1}^\horizon \indicator\left\{I_t(\algogreedy) = \idxarm,\; N_{k,t}(\algotot) = s\right\}\cdot \indicator\left\{\overline{X}_{\idxarm, s} > \meantrue_\idxarm + \frac{\gap_\idxarm}{2}\right\}\nonumber\\
    &\le \sum_{s=0}^{2(\horizon-1)} \indicator\left\{\overline{X}_{\idxarm, s} > \meantrue_\idxarm + \frac{\gap_\idxarm}{2}\right\},\label{eq:general_term_one_indicator_sum}
\end{align}
where the last inequality is due to~\eqref{eq:general_moving_at_most_one}.
Taking an expectation over~\eqref{eq:general_term_one_indicator_sum} and applying Hoeffding's inequality, we have
\begin{align}
    \Expect[\term_{1}] \le \sum_{s=0}^{2(\horizon-1)} e^{-\frac{\gap_\idxarm^2 s}{2}} \le 1+\frac{2}{\gap_\idxarm^2},\label{eq:general_term_one_expect}
\end{align}
where the last inequality is true due to the fact that $\sum_{s=0}^{\infty} e^{-\kappa s} = \frac{1}{1 - e^{-\kappa}} \leq 1+\frac{1}{\kappa}$ for any $\kappa > 0$.

\paragraph{Bounding term $\term_2$.}

Condition~\eqref{eq:assume_high_probability_bound} ensures that the optimal arm $1$ is pulled sufficiently often with high probability. Once this happens, the empirical mean of the optimal arm $1$ becomes close to its true mean.
Specifically, for every $\timestep\in [\horizon]$, we have
\begin{align}
    \Prob\left(\overline{X}_{1, N_{1,t}(\algotot)} < \meanarm_1 - \frac{\gap_\idxarm}{2}\right) 
    & = \sum_{s=1}^{2\horizon} \PP{\overline{X}_{1, s} < \meanarm_1 - \frac{\gap_\idxarm}{2},\; N_{1,t}(\algotot) = s} \nonumber\\
    & \le \sum_{s=1}^{\ell} \PP{N_{1,t}(\algotot) =s} + \sum_{s=\ell+1}^{2\horizon} \PP{\overline{X}_{1, s} < \meanarm_1 - \frac{\gap_\idxarm}{2}} \nonumber\\
    & \stackrel{\stepone}{\le} \ell \cdot \PP{N_{1,t}(\algotot) \le \ell} + \sum_{s=\ell+1}^{2\horizon} e^{-\frac{\gap_\idxarm^2\cdot s}{2} } \nonumber\\
    & \stackrel{\steptwo}{\le} \ell \cdot \PP{N_{1,t}(\algo\given (\algo, \algogreedy)) \le \ell} + \frac{2}{\gap_\idxarm^2} e^{-\frac{\gap_\idxarm^2 \cdot \ell }{2}},\label{eq:general_term_two_expression}
\end{align}
where step~\stepone is true due to Hoeffding's inequality, and step~\steptwo is due to the fact that $\sum_{s=i+1}^{\infty} e^{-\kappa s} \leq \frac{1}{\kappa} e^{-\kappa i}$ for any $\kappa > 0$ and integer $i$. Setting $\ell = \frac{4}{\gap_\idxarm^2}\log\timestep$,
we have
\begin{align}
    \PP{N_{1,t}(\algo\given (\algo, \algogreedy)) \le \ell} & = \PP{N_{1,t}(\algo\given (\algo, \algogreedy)) \le\frac{4}{\gap_\idxarm^2} \log\timestep} \nonumber\\
    &\le \PP{N_{1,t}(\algo\given (\algo, \algogreedy)) \le \frac{4}{\gapmin^2} \log\timestep} 
    < \frac{\const}{\timestep^2},\label{eq:assume_high_probability_bound_anytime}
\end{align}
where the last inequality is true by setting $\horizon = \timestep$ in condition~\eqref{eq:assume_high_probability_bound}.
Plugging~\eqref{eq:assume_high_probability_bound_anytime} to~\eqref{eq:general_term_two_expression}, we have
\begin{align*}
     \Prob\left(\overline{X}_{1, N_{1,t}(\algotot)} < \meanarm_1 - \frac{\gap_\idxarm}{2}\right) \le \frac{4}{\gap_\idxarm^2}\log\timestep \cdot\frac{\const}{\timestep^2} + \frac{2}{\gap_\idxarm^2}\frac{1}{\timestep^2},
\end{align*}
and hence
\begin{align}
    \Expect[\term_2] &\le \sum_{\timestep=1}^\horizon \Prob\left(\overline{X}_{1, N_{1,t}(\algotot)} < \meanarm_1 - \frac{\gap_\idxarm}{2}\right) \nonumber\\
    & \le \frac{2}{\gap_\idxarm^2}\sum_{\timestep=1}^\horizon \p{2\const \cdot \frac{\log\timestep}{\timestep^2} + \frac{1}{\timestep^2}}
    \le \frac{\const_1}{\gap_\idxarm^2}\label{eq:general_term_two_expect}
\end{align}
for some constant $\const_1 > 0$ that depends only on $\const$.

\paragraph{Combining terms $\term_1$ and $\term_2$.} Plugging the two terms from~\eqref{eq:general_term_one_expect} and~\eqref{eq:general_term_two_expect} back to~\eqref{eq:general_decompose_two_terms}, we have
\begin{align}
    \varnumpulls_{\idxarm}(\algogreedy) \le 1+ \frac{2+\const_1}{\gap_\idxarm^2}.\label{eq:general_two_terms_combine}
\end{align}
Summing~\eqref{eq:general_two_terms_combine} over $\idxarm \ne 1$ completes the proof.

\subsection{Proof of Theorem~\ref{thm:sign-preserve-combined}}\label{sec:proof_thm_sign_preservev_combined}

\paragraph{Upper bound for UCB.}

Applying Proposition \ref{prop:UCB_with_other} to $\algoucb_{\paramexplore_1}$ with the other algorithm being $\algoucb_{\paramexplore_2}$ yields
\begin{align}
    &\EE{R_T(\algoucb_{\paramexplore_1} \mid (\algoucb_{\paramexplore_1}, \algoucb_{\paramexplore_2}))} 
= \begin{cases}
    O( \log T)  & \text{if } \alpha = 0, \\
    O( T^{\alpha_1}) & \text{if } 0 < \alpha \le 1,
    \end{cases}\label{eq:UCB+UCB_ub}
\end{align}
proving the desired upper bound for $\algoucb_{\paramexplore_1}$. A symmetric argument yields the upper bound for $\algoucb_{\paramexplore_2}$.

\paragraph{Lower bound for UCB.}
Applying Proposition~\ref{prop:UCB+other_lb}, we have
\begin{align}
\EE{R_T(\algoucb_{\paramexplore_1}\given (\algoucb_{\paramexplore_1},\algoucb_{\paramexplore_2}))} + \EE{R_T(\algoucb_{\paramexplore_2}\given (\algoucb_{\paramexplore_1},\algoucb_{\paramexplore_2}))}= \Omega(T^{\alpha_2}).\label{eq:UCB+UCB_lb_sum}
\end{align}
Combining~\eqref{eq:UCB+UCB_ub} and~\eqref{eq:UCB+UCB_lb_sum} along with the fact that $\paramexplore_1 < \paramexplore_2$ yields
\begin{align*}
    \EE{R_T(\algoucb_{\paramexplore_2}\given (\algoucb_{\paramexplore_1},\algoucb_{\paramexplore_2}))} \ge \Omega(\horizon^{\alpha_2}),
\end{align*}
proving the desired lower bound.

\paragraph{Upper bound for \egreedy.}
Applying Proposition~\ref{prop:egreedy+other} to $\algoegreedy_{\paramexplore_1,C}$ with the other algorithm $\algo$ being $\algoegreedy_{\paramexplore_2,C}$ yields the desired upper bound for $\algoegreedy_{\paramexplore_1,C}$. A symmetric argument yields the upper bound for $\algoegreedy_{\paramexplore_2,C}$.

\paragraph{Lower bound for \egreedy.}
Applying Proposition~\ref{prop:egreedy+other_lb} to $\algoegreedy_{\paramexplore_2,C}$ with the other algorithm $\algo$ being $\algoegreedy_{\paramexplore_1,C}$ yields the desired lower bound for $\algoegreedy_{\paramexplore_2,C}$.

\subsection{Proofs of the detection procedure}
In this section, we present the proofs of the theoretical results from \Cref{sec:guidelines} for the detection procedure.

\subsubsection{Proof of Theorem~\ref{thm:e-greedy_ramp_up}}\label{sec:proof_thm_egreedy_ramp_up}

\paragraph{Lower bound.}
The $\Omega (T^{\max\{\paramexplore_1, \paramexplore_2 + \beta-1\}})$ lower bound for the cases without $\alpha_1=1$ or $1-\beta\ge \alpha_2$. Denote by $\epsilon_{1, t}$ and $\epsilon_{2, t}$ the exploration probabilities of the two algorithms. The total number of pulls on any suboptimal arm $k$ is:
    \begin{align*}
        \Expect[N_{k,T}(\algotot)] &\ge \frac{1}{\numarms}\sum_{t=1}^{2T} \big[(1-p)\cdot\epsilon_{1,t}+p\cdot \epsilon_{2,t}\big]\\
        &= \frac{1}{\numarms}\sum_{t=1}^{2T} \left((1-p)\cdot \min\left\{1, \frac{C}{t^{1-\alpha_1}}\right\}+p\cdot \min\left\{1, \frac{C}{t^{1-\alpha_2}}\right\}\right).
    \end{align*}
With a similar analysis to the proof of Proposition~\ref{prop:egreedy+other_lb} in \Cref{sec:proof_prop_egreedy+other_lb}, we have that for $0 < \paramexplore_1 \le 1$,
    \begin{align*}
        \Expect[N_{k,T}(\algotot)] &\ge (1-\probassign) \cdot \Omega(\horizon^{\paramexplore_1}) + \probassign\cdot \Omega(\horizon^{\paramexplore_2}),
    \end{align*}
    where the term $\Omega(\horizon^{\paramexplore_1})$ is replaced by $\Omega(\log\horizon)$ when $\paramexplore_1 = 0$.
    Setting $\probassign = \horizon^{\beta - 1}$ and $T \ge 2^{\frac{1}{1-\beta}}$, we have $1-\probassign = 1 - \horizon^{\beta - 1} > \frac{1}{2}$. Therefore, when $0 < \paramexplore_1 \le 1$ or $\paramexplore_2 + \beta > 1$,
    \begin{align*}
        \Expect[N_{k,T}(\algotot)]
        &= \Omega(T^{\alpha_1})+\Omega(T^{\alpha_2+\beta-1}),
    \end{align*}
    and otherwise
    \begin{align*}
        \Expect[N_{k,T}(\algotot)]
        &= \Omega(\log\horizon),
    \end{align*}
    completing the proof of the desired lower bound.

\paragraph{Upper bound.} The proof is similar to the proof of \Cref{prop:egreedy+other} in \Cref{sec:proof_prop_egreedy+other}.
The probability that any suboptimal arm $k$ is pulled at any timestep $\timestep \ge \numarms+1$ can be decomposed into the exploration step and the greedy step as (cf. Eq.~\eqref{eqn:egreedy_eqn}): 
\begin{align*}
&\PP{\armselect_\timestep = \idxarm} \nonumber\\
& \qquad \le \frac{(1-p)\cdot\epsilon_\timestep(\alpha_1, C)+p\cdot\epsilon_\timestep(\alpha_2, C)}{K} + \PP{ \overline{X}_{k, N_{k,t}(\algotot)} \ge \overline{X}_{1, N_{1,t}(\algotot)} } \nonumber\\
&\qquad \leq \frac{C}{K}\left(\frac{1}{t^{1-\alpha_1}}+\frac{T^{1-\beta}}{t^{1-\alpha_2}}\right) + \underbrace{
    \PP{\overline{X}_{k, N_{k,t}(\algotot)} \geq \mu_k+\frac{\Delta_k}{2}}
}_{\term_\timestep} + \underbrace{
    \PP{\overline{X}_{1, N_{1,t}(\algotot)} \leq \mu_1 -\frac{\gap_k}{2}}
}_{\term'_\timestep}, 
\end{align*}
We bound term $\term_\timestep$ following the steps in the proof of \Cref{prop:egreedy+other}, outlined below for completeness; a symmetric argument applies to term $\term_\timestep'$. We define $\widetilde{p}_t\defn{(1-p)\cdot\epsilon_\timestep(\alpha_1, C)+p\cdot\epsilon_\timestep(\alpha_2, C)}$. Define by $N_{k,t}^R$ the number of times in which arm $k$ is chosen in exploration steps. We define its expectation by:
\begin{align*}
    \cumprob(\timestep) \defn \EE{N_{k,t}^R}=\frac{1}{\numarms} \sum_{i=K+1}^{\timestep-1} \widetilde{p}_i.
\end{align*}
The variance of $N_{k,t}^R$ is:
\begin{align*}
    \var\sqb{N_{k,t}^R}=\sum_{\idxtimestep=K+1}^{\timestep-1}\frac{\widetilde{p}_t}{K}\left(1-\frac{\widetilde{p}_t}{K}\right) \leq \frac{1}{K} \sum_{\idxtimestep=K+1}^{\timestep-1} \widetilde{p}_i = \cumprob(\timestep).
\end{align*}
By Bernstein's inequality~\eqref{eq:bernstein_le} in \Cref{fact:bernstein}, we have
\begin{align*}
\PP{N_{k,t}^R \leq \frac{\cumprob(\timestep)}{2}} \leq e^{-\cumprob(\timestep) / 10}.
\end{align*}
Following the same steps as in \Cref{prop:egreedy+other} bounds term $\term_\timestep$ as:
\begin{align}
    \term_\timestep &\le \frac{\cumprob(\timestep)}{2} \cdot e^{-\cumprob(\timestep) / 10} + \frac{2}{\Delta_k^2} e^{-\Delta_k^2\left\lfloor \cumprob(\timestep)/4\right\rfloor}.\label{eq:rampup_egreedy_term_bound}
\end{align}
We now provide upper and lower bounds for $\cumprob(\timestep)$. 
Recall that \begin{align*}
    \widetilde{p}_t= (1-p)\cdot \min \left\{1, C/t^{1-\alpha_1}\right\}+p\cdot \min \left\{1, C/t^{1-\alpha_2}\right\} \geq \min \left\{1, C/t\right\}. 
\end{align*}
In the proof of Theorem~\ref{thm:greedy+e-greedy}, we have established the lower bound~\eqref{eq:grd_egreedy_cum_prob} that for any $\timestep\ge (C+1)^2$, 
\begin{subequations}
\begin{align*}
    \cumprob(\timestep)\ge \frac{C\log(\timestep)}{4 \numarms}.
\end{align*}
On the other hand, an upper bound for $\cumprob(\timestep)$ is
\begin{align*}
    \cumprob(\timestep) \le \frac{1}{\numarms} \sum_{\idxtimestep=K+1}^{\timestep-1} \left[\probexplore_\idxtimestep(\paramexplore_1, C)+ p\cdot \probexplore_\idxtimestep(\paramexplore_2, C)\right]
    \le \frac{C t^{\alpha_1}}{\paramexplore_1 \numarms}+\frac{C T^{\beta-1}\cdot t^{\alpha_2}}{\paramexplore_2 \numarms}
    \le \frac{C t^{\max\{\alpha_1,\alpha_2+\beta-1\}}}{\min\{\paramexplore_1,\paramexplore_2\} \numarms}.
\end{align*} 
\end{subequations}
Plugging in the lower and upper bounds for $\cumprob(\timestep)$ back to~\eqref{eq:rampup_egreedy_term_bound}, we have that for any $\timestep \ge (C+1)^2$,
\begin{align*}
    \term_\timestep & \le \frac{C t^{\max\{\alpha_1,\alpha_2+\beta-1\}}}{\min\{\paramexplore_1,\paramexplore_2\} \numarms} \timestep^{-\frac{C}{40 K}} + \frac{2}{\gap_k^2} e^{-\Delta_k^2 (\frac{ C\log\timestep}{16 K}-1)}\\
    & \le \frac{C}{2\min\{\paramexplore_1,\paramexplore_2\}\numarms} \cdot \frac{1}{\timestep^{3-\max\{\alpha_1,\alpha_2+\beta-1\}}} + \frac{2e^{\gap_\idxarm^2}}{\gap_\idxarm^2} \cdot \frac{1}{t^2},
\end{align*}
where the last inequality is true by the assumption that $C\ge\max\left\{120K,\frac{32\numarms}{\gapmin^2}\right\}$. Following the same steps in the proof of \Cref{prop:egreedy+other}, we have (cf. Eq.~\eqref{eq:egreedy_egreedy_ub_alpha_greater_than_zero})
\begin{align*}
     \Expect[N_{k,T}(\algotot)] = O(\horizon^{\max\{\paramexplore_1,\paramexplore_2+\beta-1\}})
\end{align*}
when $\paramexplore_1 > 0$ or $\paramexplore_2 + \paramrampup - 1 > 1$, and otherwise $O(\log\horizon)$.

\subsubsection{Proof of Theorem~\ref{thm:ucb_ramp_up}}\label{sec:proof_thm_ucb_ramp_up}

\textbf{Upper bound.}
We first generalize the upper bound of the UCB algorithm under data sharing in \Cref{prop:UCB_with_other} to a Bernoulli assignment.
\begin{proposition}\label{prop:UCB_with_other_bernoulli}
    Consider any $\paramexplore\in [0, 1]$ and any $\probassign\in (0, 1)$. Under a Bernoulli assignment, we jointly run the UCB algorithm $\algoucb_\paramexplore$ with any other algorithm $\mathcal{A}$, where the probability of the UCB algorithm is $\probassign$. The expected regret of $\algoucb_\paramexplore$ satisfies
\begin{align*}
    &\EE{R_T(\algoucb_\paramexplore \mid (\mathcal{A}, \algoucb_\paramexplore))} 
= \begin{cases}
    O( \log T)  & \text{if } \alpha = 0, \\
    O( T^\alpha) & \text{if } 0 < \alpha \le 1.
    \end{cases}
\end{align*}
\end{proposition}
The proof of this proposition is provided in \Cref{sec:proof_prop_ucb_with_other_bernoulli}. Applying \Cref{prop:UCB_with_other_bernoulli} to both UCB algorithms yields
\begin{subequations}\label{eq:ucb_rampup_ub}
\begin{align}
    \EE{R_T(\algoucbone \mid (\algoucbone, \algoucbtwo))} 
&= \begin{cases}
    O( \log T)  & \text{if } \alpha_1 = 0, \\
    O( T^{\alpha_1}) & \text{if } 0 < \alpha_1 \le 1.
    \end{cases}\\
    \EE{R_T(\algoucbtwo \mid (\algoucbone, \algoucbtwo))} &= O( T^{\alpha_2}).
\end{align}
\end{subequations}

Moreover, for algorithm $\algoucb_{\paramexplore_2}$, since its assignment probability is $\probassign=\horizon^{\paramrampup-1}$, it is sampled $2\horizon\probassign$ times in expectation. Therefore, its regret is also upper bounded by
\begin{align}\label{eq:ucb_rampup_naive_ub}
    \Expect[\regret_\horizon(\algoucb_{\paramexplore_2} \given (\algoucbone, \algoucbtwo))] = O(\horizon\probassign) = O(\horizon^{\paramrampup}).
\end{align}
Combining~\eqref{eq:ucb_rampup_ub} and~\eqref{eq:ucb_rampup_naive_ub} completes the proof of the upper bound.

\noindent\textbf{Lower bound.}
The proof is similar to the proof of \Cref{prop:UCB+other_lb} in \Cref{sec:proof_prop_UCB+other_lb}. At a high level, the proof of \Cref{prop:UCB+other_lb} considers the case where the best arm is pulled a large number of times (or otherwise the regret is high and gives the lower bound directly). At the last time the best arm is pulled, the confidence interval of the best arm is sufficiently small. The fact that the best arm is chosen implies that any suboptimal arm has to have a small confidence interval, yielding a lower bound on the number of pulls for any suboptimal arm.

Under a Bernoulli assignment, we apply this argument separately to the treatment and control algorithms. A challenge is that if the treatment probability is very small, then the treatment algorithm is sparsely sampled and the above argument does not hold directly: The last time the \emph{treatment algorithm} pulls the best arm can be quite early in the experiment, at which point the total number of pulls on the optimal arm is still small. A minor modification to the previous argument introduces an additional factor of $\Omega(\horizon^{\paramrampup})$; we outline the proof below for completeness.

We define events
    \begin{align*}
        \event_1 & = \{\algoucb_{\paramexplore_1} \text{ is pulled at least $\frac{\horizon}{4}$ times in the last $\horizon$ rounds}\}\\
        \event_2 &= \{\algoucb_{\paramexplore_2}\text{ is pulled at least $\frac{\horizon\probassign}{2}$ times in the last $\horizon$ rounds}\}.
    \end{align*}
    For any $T >2^{\frac{1}{1-\beta}}$, 
    the probability of sampling the control algorithm is $1-\probassign = 1 - {\horizon}^{\paramrampup-1} > \frac{1}{2}$. To bound the probability of event $\event_1$, we apply Hoeffding's inequality:
    \begin{align*}
        \Prob(\setcomplement{\event_1}) =
        \Prob\left(\binomial(\horizon, 1-p) < \frac{\horizon}{4} \right) & < \Prob\left(\binomial\left(\horizon, \frac{1}{2}\right) < \frac{\horizon}{4} \right)\\
        & < e^{-\frac{\horizon}{8}} < 0.01,
    \end{align*}
    where the last inequality holds for any $T\ge 40$. To bound event $\event_2$, we apply Bernstein's inequality from~\Cref{fact:bernstein}:
    \begin{align*}
        \Prob(\setcomplement{\event_2}) = \Prob\left(\binomial(\horizon, \probassign) < \frac{\horizon\probassign}{2}\right) & < \exp\left(-\frac{\horizon^2\probassign^2 / 8}{\horizon\probassign(1-\probassign) + \horizon\probassign / 4}\right)\\
        & = \exp\left(-\frac{\horizon\probassign / 8}{1-\probassign + 1 / 4}\right) \\
        & < \exp\left(-\frac{\horizon\probassign}{10}\right)
        = \exp\left(-\frac{\horizon^{\paramrampup}}{ 10}\right) < \exp\left(-\frac{1}{10}\right) < 0.95.
    \end{align*}
    By a union bound, we have
    \begin{align*}
        \Prob(\event_1 \intersect \event_2)  > 0.04.
    \end{align*}
    Since the Bernoulli assignment is independent from other sources of randomness, it suffices to prove the lower bound conditional on any fixed assignment under $\event_1\intersect\event_2$. We further condition on the event that the optimal arm is pulled at least $\horizon/2$ times in the first $\horizon$ timesteps (jointly by both algorithms), because otherwise the regret is linear and trivially satisfies the desired lower bound.

    For the control algorithm $\algoucb_{\paramexplore_1}$, consider the last time that $\algoucb_{\paramexplore_1}$ pulls the best arm. If the last time is within the first $\horizon$ timesteps, then conditional on $\event_1$, the regret incurred by the control algorithm in the last $\horizon$ timesteps is linear, proving the desired lower bound. If the last pull of the optimal arm happens in the last $\horizon$ timesteps, we denote this timestep by the random variable $\horizon_0$. Similar to the proof of \Cref{prop:UCB+other_lb}, we define $L_1$ and $L_2$ as the number of times that arm $1$ or arm $2$ is pulled  before timestep $\horizon_0$ (jointly by both algorithms before). 
    We define the length of the confidence interval for an arm with $s$ samples by (cf. Eq.~\eqref{eq:confidence_define}):
    \begin{align}\label{eq:confidence_define_bernoulli}
        \conf^\paramexplore_{\timestep, s} = \begin{cases}
            \sqrt{\frac{2\log(\timestep -1)}{s}} & \text{if } \paramexplore=0,\\
            \sqrt{\frac{2(\timestep-1)^\paramexplore-1}{\paramexplore s}} & \text{if } 0 < \paramexplore\le 1.
        \end{cases}
    \end{align}
    Then we have
    \begin{align*}
        \overline{X}_{2, L_2} + c_{T_0, L_2}^{\paramexplore_1} & \le \overline{X}_{1, L_1} + c_{T_0, L_1}^{\paramexplore_1}.
    \end{align*}
    Applying the rest of the steps from the proof of \Cref{prop:UCB+other_lb} yields that the total regret of the two algorithms is bounded by
    \begin{align}\label{eq:rampup_ucb_lb_one}
        L_2 = \begin{cases}
            \Omega(\log\horizon) &\text{if }\paramexplore_1=0\\
            \Omega(\horizon^{\paramexplore_1}) & \text{if } 0< \paramexplore_1\le 1.
        \end{cases}
    \end{align}
    Now consider the treatment algorithm $\algoucb_{\paramexplore_2}$, we again consider the last time $\algoucb_2$ pulls the best arm.  If the last time is within the first $\horizon$ timesteps, then conditional on event $\event_2$, the number of pulls on the suboptimal arm in the last $\horizon$ timesteps is at least $\frac{\horizon\probassign}{2} = \frac{\horizon^{\paramrampup}}{2}$. Now consider the case when the last pull of the optimal arm happens in the last $\horizon$ timesteps. Using a similar argument to that for the control algorithm, the total regret of the two algorithms is at least $\Omega(\horizon^{\paramexplore_2})$. Combining these two cases for $\algoucb_{\paramexplore_2}$ yields
    \begin{align}
        \Expect[\regret_\horizon(\algoucb_{\paramexplore_2} \given (\algoucb_{\paramexplore_1}, \algoucb_{\paramexplore_2}))] = \Omega(\horizon^{\min\{\paramexplore_2, \paramrampup\}}).\label{eq:rampup_ucb_lb_two}
    \end{align}
    Finally, combining~\eqref{eq:rampup_ucb_lb_one} and~\eqref{eq:rampup_ucb_lb_two} yields the desired lower bound $\Omega(\horizon^{\max\{\paramexplore_1, \min\{\paramexplore_2, \paramrampup\}\}})$ for the total regret.

\subsubsection{Proof of Proposition~\ref{prop:UCB_with_other_bernoulli}}\label{sec:proof_prop_ucb_with_other_bernoulli}

The proof is similar to the proof of \Cref{prop:UCB_with_other}. In the proof of \Cref{prop:UCB_with_other}, we show that when any suboptimal arm $\idxarm$ has already been pulled $\ell$ times, the UCB algorithm only pulls it an additional constant number of times, where $\ell$ is of order $\horizon^\paramexplore$ (or order $\log \horizon$ when $\paramexplore=0$). The same analysis follows under a Bernoulli assignment. For completeness, we outline the difference to the proof of \Cref{prop:UCB_with_other} below.

Algorithm~\ref{algo:joint_bernoulli} is equivalent to the following procedure. At each timestep $\timestep$, Algorithm 1 computes an arm $\algo_1(\history_\timestep)$, and likewise Algorithm 2 computes an arm $\algo_2(\history_\timestep)$. Then one of these two arms is selected based on the Bernoulli assignment $W_\timestep$. We denote by $\armselect_\timestep(\algo)$ the arm computed by algorithm $\algo$. Then at each timestep $\timestep$, only one of $\{\armselect_\timestep(\algo_1), \armselect_\timestep(\algo_2)\}$ is selected to pull the arm.

Without loss of generality, we term the UCB algorithm as the treatment algorithm (that is, it is selected when $\assignment_\timestep = 1$). The number of times that the UCB algorithm pulls any suboptimal arm $\idxarm$ is (cf.~\eqref{eq:ucb+general_ub_num_pulls_expression} in \Cref{prop:UCB_with_other}):
\begin{align*}
    N_\idxarm(\algoucb_\paramexplore) &= \sum_{\timestep=1}^{2\horizon} \indicator\{I_{\timestep}(\algoucb_\paramexplore) = \idxarm, \assignment_\timestep = 1\}\\
    & \le \sum_{\timestep=1}^{2\horizon} \indicator\{\armselect_{\timestep}(\algoucb_\paramexplore) = \idxarm\}\\
    & \le \ell + \sum_{\timestep=1}^{2\horizon} \indicator\{\armselect_{\timestep}(\algoucb_\paramexplore) = \idxarm, N_{\idxarm, \timestep}(\algotot)\ge \ell\}\\
    & \le \ell + \sum_{\timestep=1}^{2\horizon} \indicator\left\{\min_{1\le s \le \timestep-1} \overline{X}_{1, s} + \conf^\paramexplore_{\timestep, s} \le \max_{\ell \le s_\idxarm \le \timestep-1} \overline{X}_{\idxarm, s_\idxarm} + \conf^\paramexplore_{t, s_\idxarm}\right\}\\
    & \le \ell + \numarms + \sum_{\timestep=\numarms+1}^{\infty}\sum_{s=1}^{\timestep-1}\sum_{s_\idxarm=1}^{\timestep-1} \indicator\{\overline{X}_{1, s} + \conf^\paramexplore_{\timestep, s} \le \overline{X}_{\idxarm, s_\idxarm} + \conf^\paramexplore_{\timestep, s_\idxarm}\}.
\end{align*}
The rest of the proof follows the steps in the proof of \Cref{prop:UCB_with_other} in \Cref{sec:proof_prop_UCB_with_other}. Setting
\begin{align*}
    \ell = \begin{cases}
        \ceil*{(8\log \horizon) / \gap_\idxarm^2} & \text{if }\paramexplore=0\\
        \ceil*{8 (\horizon^\paramexplore-1) / (\paramexplore\gap_\idxarm^2)} & \text{if } 0< \paramexplore\le 1,
        \end{cases}
\end{align*}
yieldsthe desired upper bound
\begin{align*}
   \Expect[N_\idxarm(\algoucb_\paramexplore)]  \le \ell + \numarms + \const,
\end{align*}
for some universal constant $\const > 0$.

\subsection{Proof of Theorem~\ref{thm:greedy+exp3}}\label{sec:proof_thm_greedy+exp3}

We prove the upper bound for the greedy algorithm and the lower bound for the EXP3 algorithm separately.

\paragraph{Lower bound~\eqref{eq:thm_violate_grd_exp_ub} for EXP3.} 
Note that in Algorithm~\ref{algo:exp3}, the probability of pulling any arm $\idxarm$ at any timestep $\timestep\in [\horizon]$ is at least $\varepsilon_t$, where $\varepsilon_\timestep = \numarms^{-2/3} \timestep^{-1/3}$ for any $\timestep > \numarms$. Hence,
\begin{align}\label{eq:exp3_prob_lb}
    \Expect[\varnumpulls_{\idxarm}(\algoexp)] & \ge \sum_{\timestep=1}^\horizon \varepsilon_\timestep 
    \ge \frac{1}{\numarms^{2/3}}\sum_{\timestep=\numarms+1}^\horizon \frac{1}{\timestep^{1/3}} =\Omega(\horizon^{2/3}),
\end{align}
and
\begin{align*}
    \EE{R_T(\algoexp \mid \algoexp\rightarrow \algogreedy )} 
    = \EE{R_T(\algoexp)}
    = \sum_{\idxarm\ne 1} \gap_\idxarm\cdot \Expect[\varnumpulls_\idxarm(\algoexp)] = \Omega(\horizon^{2/3}),
\end{align*} 
completing the proof of the lower bound~\eqref{eq:thm_violate_grd_exp_lb} for the EXP3 algorithm.

\paragraph{Upper bound~\eqref{eq:thm_violate_grd_exp_ub} for greedy.}
It can be verified that Theorem~\ref{thm:greedy+general} holds when the algorithm $\algo$ does not utilize data from greedy. In this case, condition~\eqref{eq:assume_high_probability_bound} concerns the number of pulls by algorithm $\algo$ (note that algorithm $\algo$ does not access data from the greedy algorithm, so that we have $\varnumpulls_{\idxarm,t}(\algo\given (\algo\rightarrow \algogreedy)) = \varnumpulls_{\idxarm,t}(\algo)$ for any arm $\idxarm$ and timestep $\timestep$). This result is stated formally as follows. 
\begin{proposition}\label{prop:greedy+general_one_way}
    Let $\const > 0$ be a constant that depends only on $\numarms$ and $\{\gap_\idxarm\}_{\idxarm\in [\numarms]}$. Consider jointly running the greedy algorithm with any anytime algorithm $\algo$ under one-way data sharing, where the greedy algorithm has access to data from algorithm $\algo$, but algorithm $\algo$ does not have access to data from the greedy algorithm. Suppose that for any $\horizon \ge 1$, algorithm $\algo$ satisfies
\begin{align}
    \Prob\left(\varnumpulls_{\idxarmbest,T}(\algo) \le \frac{4}{\gapmin^2}\log\horizon\right) < \frac{\const}{\horizon^2}.\label{eq:assume_high_probability_bound_one_way}
\end{align}
Then under one-way data sharing, the regret incurred by the greedy algorithm satisfies
\begin{align*}
    \Expect\left[\regret_\horizon(\algogreedy\given (\algo\rightarrow \algogreedy))\right] 
    &\le \const',
\end{align*}
where $\const' > 0$ is a constant that only depends on the $\const$ and the arm gaps $\{\gap_\idxarm\}_{\idxarm\in [\numarms]}$.
\end{proposition}

The proof follows the same steps as the proof of Theorem~\ref{thm:greedy+general} in \Cref{sec:proof_thm_greedy+general} and is therefore omitted. It now remains to verify condition~\eqref{eq:assume_high_probability_bound_one_way} for the EXP3 algorithm. We denote by $\varnumpulls_{\idxarm,\horizon}^R(\algoexp)$ the number of pulls for any arm $\idxarm$ contributed from the probability $\varepsilon_t$; the rest of the pulls are contributed from the probability $(1-K\varepsilon_t)\frac{e^{\varepsilon_{t-1}\hat{Y}_{t-1}(\idxarm)}}{\sum_{\idxarm'}e^{\varepsilon_{t-1}\hat{Y}_{t-1}(\idxarm')}}$.
We thus have the deterministic relation $\varnumpulls_{\idxarm, \horizon}(\algoexp) \ge \varnumpulls_{\idxarm, \horizon}^R(\algoexp)$. We define the expected value of $\varnumpulls_{\idxarm, \horizon}^R(\algoexp)$ as:
\begin{align*}
    \cumprob(\horizon) = \Expect[\varnumpulls_{\idxarm, \horizon}^R(\algoexp)] = \sum_{\timestep=1}^{\horizon-1}\varepsilon_\timestep.
\end{align*}
From~\eqref{eq:exp3_prob_lb}, we have $\cumprob(\horizon) \ge \const_1 \horizon^{2/3}$ for some positive constant $\const_1 > 0$ independent from $\horizon$. Hence,
\begin{align}
    \PP{\varnumpulls_{\idxarm,\horizon}(\algoexp) \le \frac{4}{\Delta^2_{\min}}\log T} &\le \Prob\left(\varnumpulls_{\idxarm,\horizon}^R(\algoexp) \le \frac{4}{\Delta^2_{\min}}\log T\right) \nonumber\\
    &= \Prob\left(\varnumpulls_{\idxarm,\horizon}^R(\algoexp)-\cumprob(\horizon) \le \frac{4}{\Delta^2_{\min}}\log T-\cumprob(\horizon)\right) \nonumber\\
    & \stackrel{\stepone}{\le} \exp\left(-\frac{2}{\horizon-1}
    \left( \frac{4}{\Delta^2_{\min}}\log T-\const_{1}T^{2/3}\right)^2\right) \nonumber\\
    & \le \exp(-c_{2}T^{1/3}) \nonumber\\
    & \le \frac{1}{T^{2}},\label{eq:violate_grd_exp_condition}
\end{align}
where $c_{2} > 0$ is a positive constant independent of $T$, and step~\stepone is by Hoeffding's inequality. 
Setting $\idxarm = \idxarmbest$ in~\eqref{eq:violate_grd_exp_condition} proves condition \eqref{eq:assume_high_probability_bound_one_way}. Invoking Proposition~\ref{prop:greedy+general_one_way} completes the proof for the constant regret of the greedy algorithm. 

\subsection{Proofs of properties under \datafeedback}
In this section, we present the proofs of the properties under data sharing from \Cref{sec:properties_data_sharing}.
\subsubsection{Proof of Proposition~\ref{prop:UCB_with_other}}\label{sec:proof_prop_UCB_with_other}

The proof largely follows~\cite{auer2002finite}, with the difference being the additional data from the other algorithm. Intuitively, the UCB algorithm performs as well under data sharing, as the additional data provided by the other algorithm does not hurt. We include the proof below for completeness. 

For the UCB algorithm, recall from~\eqref{eq:confidence_define} that $c_{\timestep, \numsamples}^\paramexplore$ is the confidence bound associated with the UCB algorithm when an arm has $\numsamples$ samples at timestep $\timestep$.
Let $\ell$ be a positive integer whose value is specified later. The number of times that the UCB algorithm pulls any suboptimal arm $k$ is:
\begin{align}
N_{k}(\algoucb_\paramexplore) &= \sum_{t=1}^T \indicator
\left\{I_t(\algoucb_\paramexplore)=k\right\} \nonumber\\
&\leq \ell+\sum_{t=1}^T\indicator\left\{I_t(\algoucb_\paramexplore)=k, N_{k,t}(\algoucb_\alpha) \geq \ell\right\} \nonumber\\
&\leq \ell+\sum_{t=1}^T\indicator\left\{I_t(\algoucb_\paramexplore)=k, N_{k,t}(\algotot) \geq \ell\right\} \nonumber\\
&\stackrel{\stepone}{\leq} \ell+\sum_{t=1}^T\indicator\left\{\overline{X}_{1, N_{1,t}(\algotot)} +c_{t, N_{1,t}(\algotot)}^\paramexplore \leq \overline{X}_{k, N_{k,t}(\algotot)} +c_{t, N_{k,t}(\algotot)}^\paramexplore,\; N_{k,t}(\algotot) \geq \ell\right\} \nonumber\\
&\leq \ell+\sum_{t=1}^T\indicator\left\{\min _{1\le s\le 2t-2} \overline{X}_{1,s} +c_{t, s}^\paramexplore \leq \max _{\ell \leq s_k\le 2t-2} \overline{X}_{k, s_k}+c_{t, s_k}^\paramexplore\right\} \nonumber\\
&\leq \ell + \numarms + \sum_{t=\numarms+1}^{\infty} \sum_{s=1}^{2t-2} \sum_{s_k=\ell}^{2t-2}\indicator\left\{\overline{X}_{1,s}+c_{t, s}^\paramexplore \leq \overline{X}_{k, s_k}+c_{t, s_k}^\paramexplore\right\},\label{eq:ucb+general_ub_num_pulls_expression}
\end{align}
where step~\stepone is true by the definition of the UCB algorithm pulling arm $\idxarm$.
Now observe that $\overline{X}_{1,s}+c_{t, s}^\paramexplore \leq \overline{X}_{k, s_k}+c_{t, s_k}^\paramexplore$ implies that at least one of the following must hold
\begin{subequations}
\begin{align}
    \overline{X}_{1,s} & \leq \mu_1-c_{t, s}^\paramexplore\label{eq:ucb_ub_event_one} \\
    \overline{X}_{k, s_k} & \geq \mu_k+c_{t, s_k}^\paramexplore \label{eq:ucb_ub_event_two}\\
    \mu_1 & <\mu_k + 2 c_{t, s_k}^\paramexplore.\label{eq:ucb_ub_event_three}
\end{align}
\end{subequations}
Recall from Fact~\ref{fact:confidence_increase_in_alpha} that the value of $c_{t,s}^\paramexplore$ is increasing in $\paramexplore$ on $[0, 1]$. We bound the probability of event~\eqref{eq:ucb_ub_event_one} by Hoeffding's inequality: 
\begin{align*}
    \PP{\overline{X}_{1,s} \leq \mu_1-c_{t, s}^\paramexplore}\le \PP{\overline{X}_{1,s} \leq \mu_1-c_{t, s}^0} 
    \leq e^{-4 \log (2t-2)}=\frac{1}{(2t-2)^{4}}
\end{align*}
and similarly for event~\eqref{eq:ucb_ub_event_two}:
\begin{align*}
    \PP{\overline{X}_{k, s_k} \geq \mu_k+c_{t, s_k}^\paramexplore} \leq \PP{\overline{X}_{k, s_k} \geq \mu_k+c_{t, s_k}^0} \le e^{-4 \log (2t-2)}=\frac{1}{(2t-2)^{4}}.
\end{align*}
We set \begin{align}
    \ell=\begin{cases}
        \left\lceil(8 \log 2T) / \Delta_k^2\right\rceil & \text{ if } \alpha = 0\\
        \left\lceil8 ((2T)^\alpha - 1) / (\alpha\Delta_k^2)\right\rceil & \text{ if } 0 < \alpha \le 1.
    \end{cases}\label{eq:ucb+general_ub_define_l}
\end{align}
Then event~\eqref{eq:ucb_ub_event_three} is false for $s_k \geq \ell$, because it can be verified that when $\paramexplore=0$, we have
\begin{align*}
    \mu_1-\mu_k-2 c_{t, s_k}^0 =\mu_1-\mu_k-2 \sqrt{2(\log (2t-2)) / s_k} \geq \mu_1-\mu_k-\Delta_k=0
\end{align*}
and similarly when $0 < \paramexplore \le 1$, we have
\begin{align*}
    \mu_1-\mu_k-2 c_{t, s_k}^\paramexplore = \mu_1-\mu_k-2 \sqrt{2((2t-2)^\alpha - 1) / (\alpha s_k)} \geq \mu_1-\mu_k-\Delta_k=0.
\end{align*}
Hence, we have
\begin{align}
    \mathbb{E}\left[N_{k}(\algoucb_{\alpha})\right]
    &\leq \ell + \numarms + \sum_{t=\numarms+1}^{\infty} \sum_{s=1}^{2t-2} \sum_{s_k=\ell}^{2t-2} \PP{\overline{X}_{1,s} \leq \mu_1-c_{t, s}}+\PP{\overline{X}_{k, s_k} \geq \mu_k+c_{t, s_k}}\nonumber\\
    &\leq \ell + \numarms + 2\sum_{t=\numarms+1}^{\infty} \sum_{s=1}^{2t-2} \sum_{s_k=1}^{2t-2} \frac{1}{(2t-2)^4} \nonumber\\
    &\le \ell + \numarms + \frac{1}{2}\sum_{\timestep=1}^\infty\frac{1}{t^2}\nonumber\\
    &= \ell +\numarms + \frac{\pi^2}{12}\label{eq:ucb+general_ub_num_pulls}.
\end{align}
Plugging in the definition of $\ell$ from~\eqref{eq:ucb+general_ub_define_l} to~\eqref{eq:ucb+general_ub_num_pulls} yields
\begin{align*}
    \mathbb{E}\left[N_{k}(\algoucb_{\alpha})\right] \le \begin{cases}
     \frac{8 \log 2T}{\Delta_k^2} +\numarms + \frac{\pi^2}{12}+1 & \text{ if } \paramexplore = 0,\\
     \frac{8 ((2T)^\alpha - 1)} {\alpha\Delta_k^2} +\numarms + \frac{\pi^2}{12}+1 & \text{ if } 0 < \paramexplore \le 1.
     \end{cases}
\end{align*}
Finally, noting that $\EE{R_T(\algoucb_\paramexplore \mid (\mathcal{A}, \algoucb_\paramexplore))} \leq \sum_{k\ne 1}\gap_\idxarm\cdot \EE{N_{k}(\algoucb_\paramexplore)}$ completes the proof. 

\subsubsection{Proof of Proposition~\ref{prop:UCB+other_lb}}\label{sec:proof_prop_UCB+other_lb}

We consider the last timestep at which the optimal arm $1$ is pulled by the UCB algorithm. In order for the sum of the regrets of the two algorithms to be sublinear, the optimal arm $1$ must have $\Omega(\horizon)$ pulls. Hence, the upper confidence bound for the optimal arm $1$ concentrates around its true mean. Whenever arm $1$ is selected over arm $2$ by the UCB algorithm, the confidence bound for arm $2$ cannot be too large, and thus the number of pulls on arm $2$ cannot be too small, yielding the desired lower bound.

In the proof, it suffices to consider the case $N_{1}(\algoucb_\paramexplore \mid (\algo, \algoucb_\paramexplore))\ge \frac{T}{2}+1$, because otherwise the UCB algorithm must have $T/2$ pulls on suboptimal arms, incurring a linear regret and thus proving the $\Omega(T^\paramexplore)$ lower bound.

We now assume $N_{1}(\algoucb_\paramexplore \mid (\algo, \algoucb_\paramexplore))\ge \frac{T}{2}+1$.
Denote by random variable $T_0$ the last time that the optimal arm $1$ is pulled by $\algoucb_\paramexplore$. Then we have $T_0 \ge \frac{T}{2} + 1$. Define the number of times that arm $1$ or arm $2$ is pulled before timestep $T_0$ by
\begin{subequations}\label{eq:define_number_of_pulls_lb}
\begin{align}
    L_1 & \defn N_{1,T_0}(\algotot\given (\algo, \algoucb_\paramexplore)),\\
    L_2 & \defn N_{2,T_0}(\algotot \given (\algo, \algoucb_\paramexplore)).
\end{align}
\end{subequations}
We have $L_1 \ge \frac{T}{2}$, because the UCB algorithm must have pulled arm $1$ for $T/2$ times before timestep $T_0$. Recall the definition of the confidence bound $c_{t, s}^\paramexplore$ from~\eqref{eq:confidence_define}. Let us first consider $0 < \paramexplore_1 \le 1$.
Since the UCB algorithm selects arm $1$ over arm $2$ at timestep $T_0$, we have
\begin{align}
    \overline{X}_{2, L_2} + \conf_{T_0, L_2}^\paramexplore & \le \overline{X}_{1, L_1} + \conf_{T_0, L_1}^\paramexplore\nonumber\\
    & \stackrel{\stepone}{\le} 1 + \sqrt{\frac{2((2T_0-2)^{\paramexplore}-1)
    }{\paramexplore L_1}} \label{eq:ucb_general_lb_replace}\\
    & \stackrel{\steptwo}{\le} 1 + \sqrt{\frac{4((2T_0-2)^{\paramexplore}-1)
    }{\paramexplore T}} < c,\nonumber
\end{align}
for some positive constant $\const > 0$,
where step~\stepone is true because all rewards are bounded in $[0, 1]$, and step~\steptwo is true because $L_1 \ge \frac{T}{2}$. Hence, we have
\begin{align}
    c > \overline{X}_{2, L_2} + \conf_{T_0, L_2}^\paramexplore & \stackrel{\stepone}{\ge} 0 + \sqrt{\frac{2((2T_0-2)^{\paramexplore} - 1)}{\paramexplore L_2}}\nonumber\\
    & \stackrel{\steptwo}{\ge} \sqrt{\frac{2(\horizon^{\paramexplore} - 1)}{\paramexplore L_2}},\label{eq:ucb+general_lb_pull_bound}
\end{align}
where step~\stepone is true because all rewards are again bounded in $[0, 1]$, and step~\steptwo is true because $T_0 \ge \frac{T}{2} + 1$. Rearranging~\eqref{eq:ucb+general_lb_pull_bound}, we have the deterministic relation
\begin{align*}
    L_2 > \frac{2(\horizon^\paramexplore-1)}{\const^2\paramexplore},
\end{align*}
and therefore,
\begin{align*}
    \EE{R_T(\algoucb_\paramexplore \mid (\algo, \algoucb_\paramexplore))} + \EE{R_T(\algo \mid (\algo, \algoucb_\paramexplore))} &\ge \gap_2 \cdot \Expect[N_{2}(\algotot \given (\algo, \algoucb_\paramexplore))]\\
    & \ge \Delta_2 \cdot \Expect[L_2]=\Omega(T^\paramexplore),
\end{align*}
completing the proof for $0 < \paramexplore \le 1$.

Finally, when $\paramexplore= 0$, we use the fact that $\lim_{\paramexplore\rightarrow 0} \frac{\kappa^\paramexplore-1}{\paramexplore} = \log \kappa$ for any real-valued $\kappa > 0$. Following the same steps with the term $\frac{2((2T_0-2)^\alpha-1)}{\alpha L_1}$ in~\eqref{eq:ucb_general_lb_replace} replaced by $\frac{2\log(2T_0-2)}{ L_1}$ yields the desired $\Omega(\log T)$ lower bound when $\alpha=0$.

\subsubsection{Proof of Proposition~\ref{prop:egreedy+other}}\label{sec:proof_prop_egreedy+other}

The proof largely follows~\cite{auer2002finite}.
For an \egreedy algorithm, a suboptimal arm can be pulled in two cases:
(1) an exploration step selects this arm when sampling uniformly at random; or
(2) a greedy step selects this arm because its empirical mean is the highest among all arms.
For the exploration steps, the number of pulls on any suboptimal arm is bounded by summing the exploration probabilities where $\probexplore_\timestep = \Theta\left(\frac{1}{t^{1-\paramexplore}}\right)$, incurring a regret on the order of $\sum_{\timestep=1}^\horizon \frac{1}{t^{1-\paramexplore}} = \horizon^\paramexplore$. 
For the greedy steps, we show that the exploration steps allow the algorithm to collect sufficient samples for each suboptimal arm, and the number of pulls from the greedy steps is thus bounded by a constant.

We use the notation $\epsilon_t(\alpha, C)$ for the exploration probability of $\algoegreedy_{\alpha, C}$ to emphasize the dependence on the parameters $\paramexplore$ and $C$ explicitly:
\begin{align*}
    \epsilon_t(\alpha, C)=\min \left\{1, C/(2t-2)^{1-\alpha}\right\}.
\end{align*}
The probability that arm $k$ is pulled by $\algoegreedy_{\alpha, C}$ at any timestep $\timestep \ge \numarms+1$ can be decomposed into the exploration step and the greedy step as: 
\begin{align}
&\PP{\armselect_\timestep(\algoegreedy_{\alpha, C}) = \idxarm} \nonumber\\
& \qquad \le \frac{\epsilon_\timestep(\alpha, C)}{K} + \PP{ \overline{X}_{k, N_{k,t}(\algotot)} \ge \overline{X}_{1, N_{1,t}(\algotot)} } \nonumber\\
&\qquad \leq \frac{C}{K (2t-2)^{1-\alpha}} + \underbrace{
    \PP{\overline{X}_{k, N_{k,t}(\algotot)} \geq \mu_k+\frac{\Delta_k}{2}}
}_{\term_\timestep} + \underbrace{
    \PP{\overline{X}_{1, N_{1,t}(\algotot)} \leq \mu_1 -\frac{\gap_k}{2}}
}_{\term'_\timestep}, 
\label{eqn:egreedy_eqn}
\end{align}
where the first inequality holds because arm $k$ is selected by $\algoegreedy_{\alpha, C}$ in an exploration step with probability $\epsilon_\timestep(\alpha, C)/K$, and in a greedy step if its empirical mean is at least as large as that of the best arm.

Now we bound term $\term_\timestep$; a symmetric argument applies to term $\term_\timestep'$. Consider any suboptimal arm $k\ne 1$. Recall that $N_{k,t}^R(\algoegreedy_{\alpha, C})$ denotes the number of times in which arm $k$ is chosen in exploration steps. We define its expectation by:
\begin{align*}
    \cumprob(\timestep) \defn \EE{N_{k,t}^R(\algoegreedy_{\alpha, C})}=\frac{1}{\numarms} \sum_{i=K+1}^{\timestep-1} \epsilon_i(\alpha, C).
\end{align*}
The variance of $N_{k,t}^R(\algoegreedy_{\alpha, C})$ is:
\begin{align*}
    \var\sqb{N_{k,t}^R(\algoegreedy_{\alpha, C})}=\sum_{\idxtimestep=K+1}^{\timestep-1}\frac{\epsilon_\idxtimestep(\alpha, C)}{K}\left(1-\frac{\epsilon_i(\alpha, C)}{K}\right) \leq \frac{1}{K} \sum_{\idxtimestep=K+1}^{\timestep-1} \epsilon_\idxtimestep(\alpha, C) = \cumprob(\timestep).
\end{align*}
By Bernstein's inequality~\eqref{eq:bernstein_le} in \Cref{fact:bernstein}, we have
\begin{equation}
\PP{N_{k,t}^R(\algoegreedy_{\alpha, C}) \leq \frac{\cumprob(\timestep)}{2}} \leq e^{-\cumprob(\timestep) / 10}. \label{eq:egreedy_egreedy_ub_bernstein}
\end{equation}
We bound the term $\term_\timestep$ as:
\begin{align}
\term_\timestep &=\PP{\overline{X}_{k, N_{k,t}(\algotot)} \geq \mu_k+\frac{\Delta_k}{2}} \nonumber\\
&=\sum_{s=1}^{2\horizon-2} \PP{N_{k,t}(\algotot)=s,\; \overline{X}_{\idxarm, s} \geq \mu_k+\frac{\Delta_k}{2}} \nonumber\\
&=\sum_{s=1}^{2\horizon-2} \Prob\left(N_{k,t}(\algotot)=s \,\middle|\, \overline{X}_{\idxarm, s} \geq \mu_k+\frac{\gap_k}{2}\right) \cdot \PP{\overline{X}_{\idxarm, s} \geq \mu_k+\frac{\Delta_k}{2}} \nonumber\\
&\stackrel{\stepone}{\leq} \sum_{s=1}^{2\horizon-2} \Prob\left(N_{k,t}(\algotot)=s \,\middle|\, \overline{X}_{\idxarm, s} \geq \mu_k+\frac{\Delta_k}{2}\right) \cdot e^{-\gap_\idxarm^2 s / 2} \nonumber\\
& \stackrel{\steptwo}{\leq} \sum_{s=1}^{\left\lfloor \cumprob(\timestep) /2\right\rfloor} \Prob\left(N_{k,t}(\algotot)=s \,\middle|\, \overline{X}_{\idxarm, s} \geq \mu_k+\frac{\Delta_k}{2}\right) + \frac{2}{\Delta_k^2} e^{-\Delta_k^2\left\lfloor \cumprob(\timestep)/ 4\right\rfloor } \nonumber\\
& \leq \sum_{s=1}^{\left\lfloor \cumprob(\timestep)/2\right\rfloor} \left(N_{k,t}^R(\algoegreedy_{\alpha, C}) \leq s \,\middle| \, \overline{X}_{\idxarm, s} \geq \mu_k+\frac{\Delta_k}{2}\right) + \frac{2}{\Delta_k^2} e^{-\Delta_k^2\left\lfloor \cumprob(\timestep)/ 4\right\rfloor } \nonumber\\
& \leq \frac{\cumprob(\timestep)}{2} \cdot \PP{N_{k,t}^R(\algoegreedy_{\alpha, C}) \leq \frac{\cumprob(\timestep)}{2}} + \frac{2}{\Delta_k^2} e^{-\Delta_k^2\left\lfloor \cumprob(\timestep)/ 4\right\rfloor},\label{eq:grd_egreedy_ub_term_one}
\end{align}
where step~\stepone is due to Hoeffding's inequality, and step~\steptwo is based on the fact that $\sum_{s=i+1}^{\infty} e^{-\kappa i} \leq \frac{1}{\kappa} e^{-\kappa x}$ for any $\kappa > 0$ and integer $i$.
Plugging in Bernstein's inequality~\eqref{eq:egreedy_egreedy_ub_bernstein} to~\eqref{eq:grd_egreedy_ub_term_one}, we have 
\begin{align}
    \term_\timestep &\le \frac{\cumprob(\timestep)}{2} \cdot e^{-\cumprob(\timestep) / 10} + \frac{2}{\Delta_k^2} e^{-\Delta_k^2\left\lfloor \cumprob(\timestep)/4\right\rfloor}.\label{eq:grd_egreedy_ub_term_one_bernstein}
\end{align}
We now provide upper and lower bounds for $\cumprob(\timestep)$. 
Recall that $\epsilon_t=\min \left\{1, C/(2t-2)^{1-\alpha}\right\} \geq \min \left\{1, C/(2t)\right\}$. 
In the proof of Theorem~\ref{thm:greedy+e-greedy}, we have established the lower bound~\eqref{eq:grd_egreedy_cum_prob} that for any $\timestep\ge (C+1)^2$, 
\begin{subequations}\label{eq:grd_egreedy_ub_cumprob_bounds}
\begin{align}
    \cumprob(\timestep)\ge \frac{C\log(\timestep)}{4 \numarms}.\label{eq:grd_egreedy_ub_cumprob_lb}
\end{align}
On the other hand, an upper bound for $\cumprob(\timestep)$ is
\begin{align}
    \cumprob(\timestep) \le \frac{1}{\numarms} \sum_{\idxtimestep=K+1}^{\timestep-1} \probexplore_\idxtimestep(\paramexplore, C)
    \le\frac{1}{\numarms} \sum_{\idxtimestep=K+1}^{\timestep-1} \frac{C}{(2\idxtimestep-2)^{1-\alpha}}
    \le \frac{C t^{\alpha}}{\paramexplore \numarms}.\label{eq:grd_egreedy_ub_cumprob_ub}
\end{align} 
\end{subequations}
Plugging in the lower and upper bounds~\eqref{eq:grd_egreedy_ub_cumprob_bounds} for $\cumprob(\timestep)$ back to~\eqref{eq:grd_egreedy_ub_term_one_bernstein}, we have that for any $\timestep \ge (C+1)^2$,
\begin{align}
    \term_\timestep & \le \frac{C t^{\alpha}}{2\alpha K} \timestep^{-\frac{C}{40 K}} + \frac{2}{\gap_k^2} e^{-\Delta_k^2 (\frac{ C\log\timestep}{16 K}-1)} \nonumber\\
    & \le \frac{C}{2\alpha\numarms} \cdot \frac{1}{\timestep^{3-\alpha}} + \frac{2e^{\gap_\idxarm^2}}{\gap_\idxarm^2} \cdot \frac{1}{t^2},\label{eq:egreedy+egreedy_term_bound}
\end{align}
where the last inequality is true by the assumption that $C\ge\max\left\{120K,\frac{32\numarms}{\gapmin^2}\right\}$.
A similar argument yields the same upper bound for $\term_\timestep'$.

Let us now first consider $0 < \paramexplore \le 1$. Plugging~\eqref{eq:egreedy+egreedy_term_bound} back to~\eqref{eqn:egreedy_eqn}, we have
\begin{align}
    & \Expect[N_{k}(\algoegreedy_{\alpha, C})] \nonumber\\
    & \qquad = \sum_{\timestep=1}^\horizon \PP{I_t(\algoegreedy_{\alpha, C}) = \idxarm} \nonumber\\
    & \qquad\le (C+1)^2 + \sum_{t=(C+1)^2}^\horizon \PP{I_t(\algoegreedy_{\alpha, C}) = \idxarm}\nonumber\\
    & \qquad\stackrel{\stepone}{\leq} (C+1)^2 +\sum_{\timestep=(C+1)^2}^T \frac{C}{K (2t-2)^{1-\alpha}} + \sum_{t = (C+1)^2}^T \p{\term_\timestep + \term_\timestep'}\nonumber\\
    & \qquad \stackrel{\steptwo}{\le} (C+1)^2 + \frac{C}{K} \sum_{\timestep=(C+1)^2}^T \frac{1}{(2t-2)^{1-\alpha}} + \frac{C}{\numarms\paramexplore}\sum_{\timestep=(C+1)^2}^\horizon \frac{1}{\timestep^{3-\paramexplore}} + \frac{4e}{\gap_\idxarm^2} \sum_{\timestep=(C+1)^2}^\horizon\frac{1}{t^2}\label{eq:egreedy_egreedy_ub_alpha_greater_than_zero}\\
    & \qquad \leq (C+1)^2 + \frac{C T^{\alpha}}{K \alpha} 
    + \frac{C}{K\alpha}\cdot \frac{1}{\left(2-\alpha\right)} \frac{1}{\left((C+1)^2-1\right)^{2-\alpha}} +\frac{4e}{\Delta_k^2} \cdot \frac{1}{ ((C+1)^2-1)} \nonumber\\
    & \qquad = O(\horizon^{\paramexplore}),\nonumber
\end{align}
where step~\stepone follows from~\eqref{eqn:egreedy_eqn}, and step~\steptwo is true by plugging in the expression of $\term_\timestep$ from~\eqref{eq:egreedy+egreedy_term_bound} and likewise for the term $\term'_\timestep$.
Summing over $\idxarm \ne 1$, the expected regret of the \egreedy algorithm $\algoegreedy_{\paramexplore,C}$ when jointly running with any other algorithm $\algo$ is upper bounded by
\begin{align*}
    \EE{R_T(\algoegreedy_{\alpha, C}\given (\algoegreedy_{\alpha, C},\algo))}=O(\horizon^{\alpha}),
\end{align*}
for $0 < \paramexplore \le 1$.
Finally, when $\paramexplore=0$, we follow the same derivation with the difference that in~\eqref{eq:egreedy_egreedy_ub_alpha_greater_than_zero}, the second term becomes \begin{align*}
    \frac{C}{\numarms} \sum_{\timestep=(C+1)^2}^\horizon\frac{1}{(2\timestep-2)} = O(\log\horizon),
\end{align*}
completing the proof of the upper bound for the \egreedy algorithm.

\subsubsection{Proof of Proposition~\ref{prop:egreedy+other_lb}}\label{sec:proof_prop_egreedy+other_lb}

Note that the \egreedy algorithm incurs a constant instantaneous regret at every exploration step. The probability of an exploration step is $\probexplore_\timestep = \Theta\left(\frac{1}{\timestep^{1-\paramexplore}}\right)$. We lower bound the regret by computing the expected number of exploration steps as \begin{align*}
    \sum_{\timestep=1}^\horizon \frac{1}{\timestep^{1-\paramexplore}}=\begin{cases}
    \Theta(\log\horizon) & \text{if } \paramexplore=0,\\
    \Theta(\horizon^\paramexplore) & \text{if } 0 < \paramexplore\le 1.
    \end{cases}.
\end{align*}
We now formalize this argument separately for the case of $\paramexplore=0$ and $0 < \paramexplore \le 1$.

\paragraph{Case $\paramexplore=0$.}
For any suboptimal arm $\idxarm\ne 1$, we have
\begin{align*}
    \Expect[\varnumpulls_{\idxarm}(\algoegreedy_{0, C})] &\ge \frac{1}{K}\sum_{t=1}^T \epsilon_t \\
    & = \frac{1}{\numarms} \sum_{\timestep=1}^\horizon \min\left\{1, \frac{C}{2t-2}\right\}\\
    &\ge \frac{C}{2K}+ \frac{C}{2\numarms} \sum_{t=\lfloor C/2 \rfloor+2}^T \frac{1}{t}\\
    &\ge \frac{C}{2K}+\frac{C}{2\numarms}\left(\log(\horizon+1)-\log(C/2+2)\right).
\end{align*}
If $C$ satisfies $\horizon+1 \ge (C/2+2)^2$, then
\begin{align*}
    \Expect[\varnumpulls_{\idxarm}(\algoegreedy_{0, C})] \ge \frac{C}{2\numarms} \cdot \frac{\log(\horizon+1)}{2} \ge \frac{C}{4\numarms} \cdot \log\horizon.
\end{align*}
Otherwise, if $\horizon+1 < (C/2+2)^2$, then $C > 2(\sqrt{\horizon+1} - 2)$ and hence
\begin{align*}
    \Expect[\varnumpulls_{\idxarm}(\algoegreedy_{0, C})] \ge \frac{\sqrt{\horizon+1}-2}{\numarms}.
\end{align*}
Summing over $\idxarm\ne 1$, the expected regret of the \egreedy algorithm is lower bounded by
\begin{align*}
     \EE{R_T(\algoegreedy_{0, C}\given (\algo,\algoegreedy_{0, C}))}= \sum_{\idxarm\ne 1} \gap_\idxarm \cdot \Expect[\varnumpulls_{\idxarm}(\algoegreedy_{0, C})] = \Omega(\log\horizon),
\end{align*}
completing the case of $\paramexplore=0$.

\paragraph{Case $0 < \paramexplore\le 1$.}
A similar argument follows as in the case of $\paramexplore=0$. We have
\begin{align*}
    \Expect[\varnumpulls_{\idxarm}(\algoegreedy_{\paramexplore, C})] &\ge \frac{1}{K}\sum_{t=1}^T \epsilon_t\\
    &\ge \frac{C}{2K}+ \frac{C}{2K}\sum_{t=\lfloor C/2 \rfloor+2}^T \frac{1}{t^{1-\paramexplore}}\\
    &\ge \frac{C}{2K}+\frac{C}{2K}\cdot\frac{1}{\paramexplore}\left((T+1)^\paramexplore-(C/2 +2)^\paramexplore\right).
\end{align*}
If $C$ satisfies $\horizon+1 \ge 2(C/2+2)$, then
\begin{align*}
    \Expect[\varnumpulls_{\idxarm}(\algoegreedy_{\paramexplore, C})] \ge \frac{C}{2\numarms} \cdot \frac{1}{\paramexplore}\cdot \frac{3}{4}\horizon^\paramexplore = \Omega(\horizon^\paramexplore).
\end{align*}
Otherwise, if $\horizon+1 < 2(C/2+2)$, then $C > \horizon-3$ and hence
\begin{align*}
    \Expect[\varnumpulls_{\idxarm}(\algoegreedy_{\paramexplore, C})] \ge \frac{\horizon-3}{2\numarms}.
\end{align*}
Summing over $\idxarm$, the expected regret of the \egreedy algorithm is lower bounded by
\begin{align*}
     \EE{R_T(\algoegreedy_{\paramexplore, C} \given(\algo,\algoegreedy_{\paramexplore, C}))}= \sum_{\idxarm\ne 1} \gap_\idxarm \cdot \Expect[\varnumpulls_{\idxarm}(\algoegreedy_{\paramexplore, C})] = \Omega(\horizon^\paramexplore),
\end{align*}
completing the case of $0 < \paramexplore\le 1$.

\end{document}